%% file: thesis.tex
\documentclass
[
    oneside,
    openany,
    cleardoublepage = empty,
    fontsize = 12 pt,
    american,
    captions = tableheading,
    numbers = noenddot,
    footheight = 35 pt,
]
{scrbook}

\input{core/preamble/document_settings}

\extraBorder{3 mm}
\bindingCorrection{6 mm}
\fancyTheoremsfalse

\myTitle{Prompt-agnostic Language Models}
\myName{Frederic Sadrieh}
\myProgram{Computer Science}
\dateOfHandingIn{July 1, 2026}
\nameOfMySupervisor{Prof.\,Dr. Gerard de Melo}
\additionalExaminers{Dr. Michal Štefánik\newline Konstantin Dobler}
\mySubject{Gradient merging for prompt-robust instruction tuning of large language models}
\myKeywords{prompt robustness | gradient merging | instruction tuning}

\input{core/preamble/format}
\input{core/bibliography/bibliography_format}
\input{packages_and_commands/standard_packages}
\input{packages_and_commands/general_commands}

\newcommand{\QwenI}{Qwen3-0.6B}
\newcommand{\QwenB}{Qwen3-8B}
\newcommand{\LlamaI}{Llama3.2-1B}
\newcommand{\LlamaB}{Llama3.1-8B}
\newcommand{\clv}{conditional local variance}
\newcommand{\testall}{\textsc{test-all}}
\newcommand{\testmaj}{\textsc{test-majority}}
\newcommand{\valmaj}{\textsc{validation-majority}}

\begin{document}

\frontmatter
\include{core/title_page/title_page}

\pagestyle{plain}

\addchap{Abstract}

Despite their strong performance, large language models remain highly sensitive to prompt formulation. This issue is typically addressed at inference-time through manual prompt optimization, which introduces overhead and lacks scalability. Conversely, we focus on train-time approaches that build prompt robustness into the model during instruction tuning. A common strategy is to mix multiple paraphrases of each prompt into training. We demonstrate that across trainable parameters, the per-paraphrasing gradients interfere, with $57\%$ exhibiting sign conflicts and the naive gradient average discarding approximately a fifth of the magnitude. Therefore, we develop two gradient-merging methods, \textsc{Prompt-TIES} and \textsc{Prompt-DARE}, that address this interference by leveraging interference-resolution operators from model and gradient merging and applying them to prompt gradients. In this study, we compare the efficacy of our methods to increase model robustness to train-time robustness methods spanning all three established classes: data construction, consistency regularization, and contrastive alignment. The evaluation is conducted across four models from two families at two scales.

We observe the interaction between prompt phrasing and the specific examples accounts for $54\%$ of per-example variability on average, and exceeds $70\%$ on tasks such as ANLI and WiC. The gradient-merging methods directly improve this local robustness and nearly halve per-example variance on fully fine-tuned models at little cost to overall accuracy. The other train-time methods enhance raw performance, but fail to improve per-example variance reliably.\footnote{The project's code can be found at \url{https://github.com/FSadrieh/prompt-agnostic-llms}, and the model checkpoints can be found at \url{https://huggingface.co/prompt-agnostic-language-models}.}

\setuptoc{toc}{totoc}
\tableofcontents

\pagestyle{headings}
\mainmatter

\chapter{Introduction}
\label{intro}
Large language models (LLMs) are increasingly deployed in real-world decision-making pipelines and agentic systems. Yet two semantically equivalent prompts routinely produce different outputs \cite{inger2025forget, cao2024worst}. This variance can undermine users' trust in the system \cite{Li2025WhenMS}. Model scale and standard instruction tuning raise mean performance but leave prompt sensitivity largely intact \cite{chatterjee-etal-2024-posix, wei2025paft, cao2024worst}. We analyze prompt robustness across four models from two families and two scales and find that a single template is rarely best or worst across all examples. The primary source of variability is the interaction between a template and the specific example to which it is applied. This local, example-level variance is the focal point of this thesis. We decrease it by resolving the gradient interference that arises when a model is trained on many phrasings at once.

Prompt brittleness is often addressed at inference-time, through manual prompt engineering or search. However, these remedies are labor-intensive, must be redone for every task, and inherently lack scalability \cite{schulhoff2025promptreportsystematicsurvey}. Conversely, we focus on methods that train the model to be more prompt robust \cite{wei2025paft, yan2024contrastive, qiang-etal-2024-ppcl}. The simplest concept underlying most methods is fine-tuning on multiple phrasings of each instruction rather than one. This approach improves upon single-prompt instruction tuning \cite{wei2025paft}. More elaborate methods add auxiliary loss objectives \cite{yan2024contrastive, qiang-etal-2024-ppcl}. Consistency regularization aligns the output distributions of the phrasings \cite{qiang-etal-2024-ppcl} and contrastive approaches align their hidden representations \cite{yan2024contrastive}. All current train-time methods treat the optimizer as a black box. They decide which phrasings the model sees and which loss it optimizes but leave the gradients untouched \cite{wei2025paft, yan2024contrastive, qiang-etal-2024-ppcl}.

This thesis opens the black box. We find that per-phrasing gradients interfere. Across trainable parameters, $57\%$ see at least two phrasings pushing in opposite directions. The naive gradient average of standard training discards approximately one-fifth of the gradient magnitude through cancellation. Model and gradient merging encounter conflicting update directions when combining independently fine-tuned models in weight space \cite{YadavTies2023, chen2020just}. They resolve direction or sign conflicts using operators such as \textsc{TIES} \cite{YadavTies2023} and \textsc{DARE} \cite{YuDare2024}. We translate the methods applied traditionally to per-task updates \cite{YadavTies2023, YuDare2024, chen2020just, lu2024online} to per-prompt gradients developing \textsc{Prompt-TIES} and \textsc{Prompt-DARE}. These methods resolve the sign conflict of per-prompt gradients and reduce the model's sensitivity to prompt phrasing.

We compare \textsc{Prompt-TIES} and \textsc{Prompt-DARE} in a controlled comparison of train-time prompt robustness approaches spanning all three established classes: data construction \cite{wei2025paft}, consistency regularization (\textsc{PPCL}) \cite{qiang-etal-2024-ppcl}, and contrastive alignment (\textsc{CoIN}) \cite{yan2024contrastive}. By scoring every method on a performance--robustness trade-off, we find a clear distinction: data construction methods improve mean accuracy the most, while gradient merging provide the best local robustness on fully fine-tuned models. \textsc{Prompt-TIES} and \textsc{Prompt-DARE} lower per-example variance at minimal cost to accuracy by directly reducing the example--template interaction that drives the brittleness. We analyze the consistency regularization and contrastive approaches in and out of domain and find that their gains do not transfer between training and testing distributions.

For practitioners, this yields a simple rule of thumb. If the goal is raw performance, then data construction suffices. However, if local per-example robustness matters and the model is fully fine-tuned, then adding \textsc{Prompt-TIES} or \textsc{Prompt-DARE} improves robustness significantly.

Our main contributions are:
\begin{enumerate}
\item We show why prompt brittleness is best measured as \emph{local} example-level variance rather than as global worst-template performance. The local variance is dominated by the example--template interaction.
\item We characterize gradient interference in multi-prompt instruction fine-tuning, showing it is dominated by per-parameter sign conflict and concentrated in the early layers of the network.
\item We propose \textsc{Prompt-TIES} and \textsc{Prompt-DARE}, which resolve this interference online by bringing model-merging operators to per-prompt gradients. We show they deliver the best local prompt robustness on fully fine-tuned models.
\item We provide a controlled comparison of \textsc{Prompt-TIES} and \textsc{Prompt-DARE} against train-time prompt-robustness methods across three classes: data construction, consistency regularization, and contrastive alignment.
\item We diagnose why consistency regularization and contrastive objectives underperform simple data construction and gradient merging, tracing each to the lack of transfer between train and test distributions.
\end{enumerate}

\chapter{Related work}

\section{Prompt sensitivity}

Prompt sensitivity is a well-documented phenomenon in large language models. First, we review work that characterizes and quantifies the problem. Then, we turn to approaches designed to increase robustness. There are two types of approaches: inference-time and training-time. Inference-time approaches do not change the model weights and are applied after training. Train-time approaches modify the weights during training.

\subsection{The problem of prompt sensitivity}
Recent studies have shown that the performance of LLMs is highly sensitive to prompt formulation \cite{mizrahi-etal-2024-state, seleznyovWhenPunctuationMatters2025, Habba_2025, zhu2024promptbench}. This sensitivity has been measured along complementary axes. \textsc{FormatSpread} \cite{sclar2024quantifying} isolates purely spurious formatting choices, such as separator tokens, casing, and spacing, and reports the resulting performance spread. \textcite{sclar2024quantifying} find a gap of up to $76$ accuracy points between semantically equivalent formats. \textsc{PromptBench} \cite{zhu2024promptbench} simulates user errors through adversarial perturbations at the character, word, and sentence level, measuring the performance drop. \textcite{mizrahi-etal-2024-state} take a paraphrase-centric approach, evaluating over $6.5$ million instances across $20$ models and $39$ tasks. They demonstrate that different paraphrases not only alter absolute scores but also alter the relative ranking of models. \textsc{DOVE} \cite{Habba_2025} generalizes this to a holistic, multidimensional setting by jointly perturbing dimensions such as delimiters, answer enumerators, instruction wording, and few-shot examples. Brittleness emerges across all these benchmarks under a wide range of perturbations and tasks \cite{chatterjee-etal-2024-posix, he2024doespromptformattingimpact, loya-etal-2023-exploring, Zhuo2024ProSAAA, salinasButterflyEffectAltering2024}. 

Crucially, this instability does not vanish with scale \cite{salinasButterflyEffectAltering2024, cao2024worst, sclar2024quantifying}. \textcite{cao2024worst} report that the standard deviation across prompts for the Llama-2-chat family slightly increases from the $7$B to the $70$B model ($0.133 \rightarrow 0.156$) even as mean performance rises. The gap between the best and worst prompt for Llama-2-70B-chat reaches $45.48$ points, with its worst prompt scoring as low as $9.38\%$ \cite{cao2024worst}. Similarly, standard instruction fine-tuning is insufficient. It raises mean performance but leaves prompt sensitivity largely intact \cite{chatterjee-etal-2024-posix, wei2025paft}. Instruction-tuned models still degrade consistently when given valid but unobserved instruction phrasings \cite{sun2024evaluating}.

\subsection{Inference-time approaches}
Inference-time approaches address prompt brittleness through interventions applied at deployment, leaving the model weights untouched. They fall into three broad groups.

\paragraph{Calibration and output debiasing.}
Calibration methods adjust the model's output distribution to counteract biases that make predictions sensitive to the surface features of prompts. Contextual calibration \cite{zhao2021calibrate} estimates the model's bias toward each answer. It does so by querying the model with a content-free input (e.g., "N/A") and fitting an affine transformation that equalizes these predictions. This mitigates the majority-label, recency, and common-token biases that drive variance across prompt formats and example orderings. Batch Calibration \cite{zhou2024batch} unifies several estimators and infers them from the marginal predictions over a batch of unlabeled inputs. \textsc{PriDe} \cite{zheng2024large} limits the option-position bias of multiple-choice prompts by removing biases associated with specific option IDs estimated from a few permuted samples.

\paragraph{Prompt optimization.}
A second group searches the prompt itself for formulations the model handles well. \textsc{ZOPO} \cite{hu2024localized} argues that well-performing local optima are prevalent and uses a Neural-Tangent-Kernel-based Gaussian process within zeroth-order optimization to locate them with few model queries. \textsc{COPLE} \cite{Zhan2024UnveilingTL} demonstrates that LLMs are overly sensitive to imperceptible lexical changes in their instructions. They cast prompt repair as a combinatorial optimization that substitutes words iteratively according to their influence on a batch of proxy tasks. Closest to training-time methods,  \textsc{BATprompt} \cite{shiRobustnessawareAutomaticPrompt2025} perturbs the input and then uses the LLM's self-reflection capabilities to refine the prompt so that it can withstand such perturbations without access to gradients or parameters.

\paragraph{Aggregation.}
Rather than committing to a single prompt, a third group queries the model several phrasings of the same prompt and aggregates the predictions to reduce the variance attributable to any one formulation. \textsc{Ask Me Anything} \cite{arora2022ask} reformats each input into multiple question-answering prompts and combines their noisy predictions through weak supervision. This allows an ensemble of imperfect prompts to outperform a single hand-tuned prompt. \textsc{Self-consistency} \cite{wang2022self} aggregates sampled reasoning paths for a fixed prompt and reduces decoding variance.

For every task, prompt, or input, the inference-time intervention must be paid. Calibration parameters and selection priors are re-estimated per setting. Prompt search is re-run whenever the task changes. Aggregation multiplies the number of forward passes. Importantly, none of these methods alters the underlying model. This causes the sensitivity to reemerge once the intervention is removed \cite{hu2024localized, shiRobustnessawareAutomaticPrompt2025}.

\subsection{Train-time approaches}
Train-time approaches instead build invariance directly into the model by increasing the diversity of prompt formulations seen during training. These can be separated into three subclasses:

\paragraph{Data construction approaches.}
Data approaches are the simplest subclass and form the basis for the others. They leave the objective untouched and only enlarge the set of prompt formulations on which the model trains \cite{wei2022Flan, wei2025paft}. The first studies source prompt formulations manually and demonstrate that instruction tuning on these formulations outperforms standard instruction tuning \cite{wei2022Flan, sanh2022t0}. Since manually authoring templates is not scalable, subsequent studies have delegated the generation process to a model \cite{wei2025paft, wang-etal-2023-self-instruct}. Here, two lineages diverge in purpose. The first keeps the task fixed and generates semantically equivalent paraphrases of the prompt, similar to our approach. \textsc{PAFT} \cite{wei2025paft} samples paraphrased prompts from a commercial LLM, \textsc{RIFF} \cite{najafi2024riff} trains a dedicated paraphraser, and \textsc{GANPrompt} \cite{li2024ganprompt} uses a GAN to generate diverse prompts for robustness. The second lineage has the model instead bootstrap new instructions and instances to widen task coverage and instruction complexity \cite{wang-etal-2023-self-instruct, honovich-etal-2023-unnatural, xu2024wizardlm}. None of these methods investigate how the chosen prompt formulations should be arranged once assembled. PAFT commits to a single schedule, cycling one formulation every $K$ steps, without testing alternatives. We therefore fix the formulations and study this batch construction (see \autoref{sec:data_construction}).

\paragraph{Contrastive alignment.}
\textcite{aissi-etal-2025-reinforcement} show that models are prompt-sensitive by default, clustering prompts in their hidden states by lexical similarity rather than semantic meaning. Contrastive approaches address this issue by adding a loss objective that aligns formulations according to their semantic meaning. \textsc{CoIN} \cite{yan2024contrastive} adds such a contrastive objective to the standard cross-entropy loss. Subsequent work adapts this idea to other training stages. \textcite{aissi-etal-2025-reinforcement} use the contrastive loss function to improve prompt robustness in reinforcement learning. \textcite{Liu2025ImprovingCL} first select the most sensitive layers by measuring the hidden states between prompt formulations, and then only train these layers with the contrastive loss function. We test our gradient-merging approaches against \textsc{CoIN} \cite{yan2024contrastive}, since it modifies the same training phase.

\paragraph{Consistency regularization.}
Consistency regularization approaches introduce additional loss objectives that penalize divergence in a model's output distribution for different formulations of the same example. \textsc{PPCL} \cite{qiang-etal-2024-ppcl} measures this divergence with the Jensen--Shannon divergence between the average output token probabilities. A similar objective is used in \textcite{sun2024evaluating}, which only trains soft prompt tokens \cite{lesterPowerScaleParameterEfficient2021a}. Additionally, unsupervised approaches use model predictions as labels between paraphrases \cite{hejabi2025flipflopconsistencyunsupervisedtraining, zhou-etal-2022-prompt}. Notably, \textcite{zhou-etal-2022-prompt} employ the Jensen--Shannon divergence loss from \cite{qiang-etal-2024-ppcl} as well as distillation across paraphrases. We select \textsc{PPCL} as a baseline for our gradient-merging approaches, since it most closely resembles our experimental setup.

\section{Model and gradient merging}
\label{sec:rw_merging}

Our gradient-merging methods (see \autoref{sec:gradient}) are based on previous work of model merging and gradient merging, both of which are typically applied to multi-task training \cite{zhou2025task, wang2020gradient}. Model merging combines weight updates from several models fine-tuned independently on the same base checkpoint \cite{ilharcoTA2023}, while gradient merging combines per-task gradients during training \cite{wang2020gradient}.

\subsection{Model merging}
\label{sec:rw_modelmerge}

\textcite{ilharcoTA2023} show that adding, negating, and averaging the weight updates (task vectors) of fine-tuned models alters their behavior in predictable ways. Recent work identifies interference as a flaw in this merging process, whereby the task vectors disagree and cancel each other out \cite{YadavTies2023, YuDare2024}.

\textsc{TIES} \cite{YadavTies2023} focuses on sign- and redundancy-based per-parameter interference. The merging process has three steps: (1) trimming the low-magnitude weight updates, which tend to decrease the average update step, (2) calculating the majority sign of the per-template updates, and (3) merging only over entries with the majority sign \cite{YadavTies2023}. The goal is to receive larger weight updates as redundancy and sign interference are removed so that they do not drag down the average. \textsc{DARE} \cite{YuDare2024} replaces magnitude-based trimming with random dropping. Then, it rescales the task vector so that the expectation of each task vector is preserved. Otherwise it applies the same \textsc{TIES} steps. \textsc{KnOTS} \cite{stoica2025knots} show that these approaches are not optimal for LoRA-generated task vectors \cite{hu2021loralowrankadaptationlarge}, since these vectors occupy incompatible subspaces. \textcite{stoica2025knots} therefore first project them into a shared subspace via singular value decomposition before merging with \textsc{TIES} or \textsc{DARE}.

\subsection{Gradient merging}
\label{sec:rw_gradsurgery}

Interference is also a problem for gradient-merging approaches. \textcite{yu2020gradient} find that averaging conflicting gradients yields a poor compromise update that can stall optimization. Two subclasses of methods attempt to eliminate this interference:

\paragraph{Cosine similarity approaches.}
\textcite{yu2020gradient} declare two gradients to be conflicting when their cosine similarity is negative and project each gradient onto the normal plane of the other gradient, thereby removing the conflicting component. \textcite{wang2020gradient} observe that \textcite{yu2020gradient} implicitly target a gradient similarity of zero for every task pair, whereas closely related tasks naturally share positive similarity. They therefore set adaptive, pair-specific similarity objectives and align gradients toward them. Since we train on multiple formulations rather than different tasks, our cosine similarities between gradients are consistently positive and relatively high, while still exhibiting substantial sign-based interference (see \autoref{sec:res_gradient}). Therefore, we opt for a sign-based interference reduction.

\paragraph{Sign-based approaches.}
\textsc{GradDrop} \cite{chen2020just} probabilistically elects a sign for each parameter and only averages over the sign-agreeing entries. Similarly, \textcite{lu2024online} create \textsc{OnTIES} and \textsc{OnDARE}, the gradient variants of the model-merging approaches. These sign-based approaches are most similar to ours, but still distinct: \textsc{GradDrop} trains on multiple tasks rather than prompt formulations \cite{chen2020just}, while \textsc{OnTIES} and \textsc{OnDARE} merge the current gradient during reinforcement learning with a task vector obtained from prior supervised fine-tuning \cite{lu2024online}.

\paragraph{Criticism of gradient merging.}
Gradient merging has its critics. Large-scale studies report that standard training matches specialized multi-task optimizers once task weights are tuned, and that reported gains stem from under-tuned baselines rather than the optimizer itself \cite{xin2022current, kurin2022defense}. We address this issue by training multiple competitive baselines and finding similar evidence.

\chapter{Methods}
\label{ch:methods}
\section{Problem definition}
\label{sec:problem_def}

\paragraph{Single-template instruction fine-tuning (SIFT).}
Given a collection of $n$ datasets $\{\mathcal{D}_1,$ $\dots, \mathcal{D}_n\}$, where each $\mathcal{D}_i = \bigl((X_{i,j},Y_{i,j})\bigr)_{1\le j\le m_i}$ contains $m_i$ samples of raw inputs $X$ and labels $Y$, a prompt template $\mathcal{T}_{i,1} : (X_{i,j}, Y_{i,j}) \mapsto (x_{i,j,1}, y_{i,j,1})$ renders a sample into a textual instruction $x$ and an expected response $y$. Single-template instruction fine-tuning (SIFT) applies one such template per dataset to obtain the fine-tuning set

\begin{equation}
\mathcal{D}_\text{SIFT} \;=\; \bigl( (x_{i,j,1},\, y_{i,j,1}) \bigr)_{\,1 \le i \le n,\; 1 \le j \le m_i}.
\end{equation}
The fine-tuning set is used to train a model $\mathcal{M}$ to maximize the likelihood of each target response given its instruction. We iteratively update $\mathcal{M}$ on batches $\mathcal{B} \subseteq \mathcal{D}_\text{SIFT}$ by minimizing the token-level cross-entropy $\mathcal{L}_\text{CE}$:
\begin{equation}
\mathcal{L}_\text{SIFT} = \frac{1}{|\mathcal{B}|} \sum_{(x, y) \in \mathcal{B}} \mathcal{L}_\text{CE}(x, y).
\label{eq:ce}
\end{equation}

\paragraph{Multi-template instruction fine-tuning (MIFT).}
The methods we study exploit the fact that each dataset $\mathcal{D}_i$ is associated with not one but a set of templates $\{\mathcal{T}_{i,1}, \dots, \mathcal{T}_{i,k_i}\}$, where each $\mathcal{T}_{i,l} : (X_{i,j}, Y_{i,j}) \mapsto (x_{i,j,l}, y_{i,j,l})$ produces a distinct formulation of the same sample.

Given a per-dataset template selection $\mathcal{S}=\{\mathcal{S}_1, \dots, \mathcal{S}_n\}$ with $\mathcal{S}_i \subseteq \{1, \dots, k_i\}$, we generalize the fine-tuning set to
\begin{equation}
\mathcal{D}_\text{MIFT}(\mathcal{S}) \;=\;
\bigl( (x_{i,j,l},\, y_{i,j,l}) \bigr)_{\,1 \le i \le n,\; 1 \le j \le m_i,\; l \in \mathcal{S}_i}.
\end{equation}

Methods then differ along three orthogonal axes: (1) the template selection $\mathcal{S}$; (2) a batch-construction policy $\pi$ that induces a distribution over batches $\mathcal{B} \subseteq \mathcal{D}_\text{MIFT}(\mathcal{S})$ and, in particular, governs which formulations of the same example co-occur within a batch; (3) and an optional auxiliary objective added to the cross-entropy of Eq.~\autoref{eq:ce}, giving the general training loss

\begin{equation}
\mathcal{L}_\text{MIFT} = \frac{1}{|\mathcal{B}|} \sum_{(x, y) \in \mathcal{B}} \mathcal{L}_\text{CE}(x, y)
\;+\; \mathcal{L}_\text{aux}(\mathcal{B}; \mathcal{M}),
\label{eq:mift}
\end{equation}
recovering SIFT with $|\mathcal{S}_i| = 1$, $\pi$ random shuffling, and $\mathcal{L}_\text{aux}(\mathcal{B}; \mathcal{M}) = 0$.

\paragraph{Majority templates.}
The consistency and contrastive methods compare the model's outputs across formulations of the same example and therefore require those formulations to share a single target. We call two templates \emph{label-consistent} if they assign the same expected response to every sample
\begin{equation}
\mathcal{T}_{i,l} \sim \mathcal{T}_{i,l'}
\quad\Longleftrightarrow\quad
\forall\, j \in \{1, \dots, m_i\}:\; y_{i,j,l} = y_{i,j,l'} .
\end{equation}
This partitions the templates of $\mathcal{D}_i$ into label-consistent classes. We define the \emph{majority templates} $\mathcal{S}_i^\star$ as the class of greatest cardinality, breaking ties arbitrarily.\footnote{We implement the majority class calculation as a greedy majority class vote to reduce the number of example-template checks in preprocessing.} Setting $\mathcal{S}_i = \mathcal{S}_i^\star$ yields the largest selection for which every $(X_{i,j}, Y_{i,j}) \in \mathcal{D}_i$ has a single expected response $y_{i,j} \coloneqq y_{i,j,l}$ shared across all $l \in \mathcal{S}_i^\star$, making the shared-target assumption of \textsc{PPCL} and \textsc{CoIN} well-defined.

\section{Baseline training methods}
\label{sec:baselines}

All of our training methods use the same batch size $B = |\mathcal{B}|$. Each batch is drawn from a single dataset, $\mathcal{D}_i$, and an optimizer step aggregates several batches sampled from different datasets (for exact hyperparameters see \autoref{tab:shared_hyp}). Therefore, the methods act within a single dataset, where the number of templates $k_i$, is the same for each example in the batch. Each strategy is trained using the same underlying examples. Only the number of rendered instances differs based on the size of $\mathcal{S}_i$. In this section, we will describe the baselines against which we will compare our gradient merging approaches.

\subsection{Data construction approaches}
\label{sec:data_construction}
The data-construction strategies leave the objective unchanged ($\mathcal{L}_\text{aux} = 0$) and differ only in how the available formulations are scheduled across batches. Varying only the template and example diversity in a batch allows us to isolate the effect of that diversity on downstream robustness and test whether the schedule in prior work is optimal \cite{wei2025paft}. We will test four different construction approaches:

\paragraph{\textsc{Single}.}
\textsc{Single} sets $|\mathcal{S}_i| = 1$ for every dataset and $\pi$ to shuffle examples uniformly. This reproduces SIFT \cite{wei2022Flan} (\autoref{sec:problem_def}). Each batch contains $B$ distinct examples in a single formulation with maximal example diversity and no template diversity. We use this construction approach to test two factors: the extent to which MIFT improves performance and robustness, and the impact of the choice of single template.

\paragraph{\textsc{All shuffled}.}
\textsc{All shuffled} sets $\mathcal{S}_i = \{1, \dots, k_i\}$ and $\pi$ to shuffle examples uniformly. There are no further constraints on batch composition, so a batch may contain several formulations of the same example and mix arbitrary examples and formulations. This is the least structured use of the templates and serves as our multi-template reference with no constraints.

\paragraph{\textsc{One-at-a-Time}.}
\textsc{One-at-a-Time} sets $\mathcal{S}_i = \{1, \dots, k_i\}$ and constrains $\pi$ so that every batch is template-homogeneous. $\pi$ applies a single template $l \in \mathcal{S}_i$ to a set of $B$ examples $\mathcal{E}_\mathcal{B} \subseteq \{1, \dots, m_i\}$,
\begin{equation}
\mathcal{B}(i, l, \mathcal{E}_\mathcal{B}) = \bigl( (x_{i,j,l},\, y_{i,j,l}) \bigr)_{j \in \mathcal{E}}, \qquad |\mathcal{E}_\mathcal{B}| = B.
\label{eq:one_at_a_time}
\end{equation}
For a given set of examples $\mathcal{E}_\mathcal{B}$ $\pi$ emits the full set of single-template batches $\{\mathcal{B}(i, l, \mathcal{E}_\mathcal{B}) \mid l \in \mathcal{S}_i\}$, holding $\mathcal{E}_\mathcal{B}$ fixed while varying only $l$. Thus, the same examples recur under every formulation, and all of $\mathcal{S}_i$ is covered during training. This construction approach ensures there is no template diversity within each batch but full template diversity across the training set, and mirrors the schedule of \cite{wei2025paft}.

\paragraph{\textsc{All-in-One-Batch}.}
\textsc{All-in-One-Batch} sets $\mathcal{S}_i = \{1, \dots, k_i\}$ and constrains $\pi$ to the opposite extreme. $\pi$ draws a set of examples $\mathcal{E}_\mathcal{B} \subseteq \{1, \dots, m_i\}$ and renders each under every template,

\begin{equation}
\mathcal{B}(i, \mathcal{E}_\mathcal{B}) = \bigl( (x_{i,j,l},\, y_{i,j,l}) \bigr)_{j \in \mathcal{E}_\mathcal{B},\; l \in \mathcal{S}_i}, \qquad |\mathcal{E}_\mathcal{B}| = B / |\mathcal{S}_i|,
\label{eq:all_in_one_batch}
\end{equation}

holding only $B / |\mathcal{S}_i|$ distinct examples. We trade within-batch example diversity for within-batch template diversity and place all $|\mathcal{S}_i|$ renderings of an example in the same gradient step. \textsc{All-in-One-Batch} forms the basis for many train-time robustness approaches, like \cite{qiang-etal-2024-ppcl}, and our gradient merging methods.

\subsection{Consistency regularization (\textsc{PPCL})}
\label{sec:PPCL}
We choose \textsc{PPCL} \cite{qiang-etal-2024-ppcl} from the consistency regularization approaches as a baseline for our gradient merging methods. It adds an auxiliary loss term that penalizes divergence between the model's outputs on different formulations of the same example. It uses the \textsc{All-in-One-Batch} data construction, restricted to the majority templates ($\mathcal{S}_i = \mathcal{S}_i^\star$), so every base example occurs in a batch together with all of its label-consistent renderings.

For a rendered pair $(x_{i,j,l}, y_{i,j,l})$, let $\bar{p}_{i,j,l}$ be the model's output distribution averaged over the response tokens,
\begin{equation}
\bar{p}_{i,j,l} = \frac{1}{|y_{i,j}|}\sum_{t=1}^{|y_{i,j}|} p_{\mathcal{M}}\left(\,\cdot \mid y_{i,j,<t},\, x_{i,j,l}\,\right) \in \Delta^{|\mathcal{V}|}.
\end{equation}
The original paper contrasts one \emph{clean} to one \emph{perturbed} formulation and calculates the Jensen--Shannon divergence over $\bar{p}_{\text{clean}}$ and $\bar{p}_{\text{perturbed}}$ \cite{qiang-etal-2024-ppcl}. In our setup, we do not have clean renderings, so we treat all formulations symmetrically and average the Jensen--Shannon divergence over every unordered pair of templates:
\begin{equation}
\mathcal{L}_\text{aux} =
\frac{\lambda_3}{|\mathcal{E}_\mathcal{B}|} \sum_{(i,j)\in\mathcal{E}_\mathcal{B}}
\binom{|\mathcal{S}_i^\star|}{2}^{-1} \sum_{\{l,l'\}\subseteq \mathcal{S}_i^\star}
\mathrm{JS}\left(\bar{p}_{i,j,l} \,\big\|\, \bar{p}_{i,j,l'}\right),
\label{form:PPCL_loss}
\end{equation}
where $\mathcal{E}_\mathcal{B}$ is the set of base examples in the batch,
$\mathrm{JS}(p\|q) = \tfrac{1}{2}\mathrm{KL}(p\|m) + \tfrac{1}{2}\mathrm{KL}(q\|m)$ with
$m = \tfrac{1}{2}(p+q)$, and $\lambda_3$ controls the trade-off between instruction following and
robustness \cite{qiang-etal-2024-ppcl}. The original objective splits the cross-entropy into separate clean and perturbed terms, with individual weights $\lambda_1, \lambda_2$ \cite{qiang-etal-2024-ppcl}. As we have varying amounts of templates per dataset, we set all cross-entropy weights to $1$, so the supervised loss is the standard batch cross-entropy and the Jensen--Shannon term is its only addition.

\subsection{Contrastive alignment (\textsc{CoIN})}
\label{sec:coin}
Among the contrastive approaches, we select \textsc{CoIN} \cite{yan2024contrastive}, which adds a contrastive term calculated over the hidden representations. It pulls together formulations of the same example and pushes apart formulations of different examples. \textsc{CoIN} requires a structured batch. Each \emph{\textsc{CoIN} group} contains four renderings drawn from one dataset under its majority templates,
\begin{equation}
\mathcal{B}_\text{CoIN} = \bigl\{ (x_{i,j,l}, y_{i,j,l}),\; (x_{i,j,l^*}, y_{i,j,l^*}),\; (x_{i,j^*,l}, y_{i,j^*,l}),\; (x_{i,j^*,l^{**}}, y_{i,j^*,l^{**}}) \bigr\},
\end{equation}
with $j \neq j^*$ ($1 \le j, j^* \le m_i$), distinct $l, l^*, l^{**} \in \mathcal{S}_i^\star$ and $y_{i,j,l} = y_{i,j,l^*}$, $y_{i,j,l} \neq y_{i,j^*,l}$. The example selection is constrained by $y_{i,j,l} \neq y_{i,j^*,l^{**}}$. The first pair is the \emph{anchor} for which the group is constructed. The second pair is the \emph{positive} pair, with the same input and a different formulation. The last two are \emph{negatives}, with the third having the anchor's template applied to a different example so that the instruction matches but the instance differs \cite{yan2024contrastive}. To keep $B$ fixed, we pack several \textsc{CoIN} groups into one batch and average their losses, computing the negatives within each group only.

Let $h^\bullet_{i,j,l} = h_{\mathcal{M}}(x_{i,j,l}, y_{i,j,l})$ be the hidden state of the designated readout token for a rendered pair, and write the anchor, positive, and two negatives as $h, h^+, \{h^-\}$. The contrastive loss for one group is
\begin{equation}
\mathcal{L}_\text{ctr} = -\log
\frac{\exp\bigl(\mathrm{sim}(h, h^+)/\tau\bigr)}
{\exp\bigl(\mathrm{sim}(h, h^+)/\tau\bigr) + \sum_{h^-}\exp\bigl(\mathrm{sim}(h, h^-)/\tau\bigr)},
\end{equation}
where $\mathrm{sim}(h_1, h_2) = \frac{h_1^\top h_2}{\lVert h_1\rVert\,\lVert h_2\rVert}$ is the cosine similarity and $\tau$ the temperature \cite{yan2024contrastive}. The auxiliary objective $\mathcal{L}_\text{aux}$ is $\mathcal{L}_\text{ctr}$ averaged over the \textsc{CoIN} groups in the batch. It is weighted by the contrastive loss ratio $\lambda$ and capped at $\mathcal{L}_\text{CE}$ following \cite{yan2024contrastive}.

\textcite{yan2024contrastive} read $h^\bullet_{i,j,l}$ from the last target token during training, which we denote as $\text{CoIN}_\text{label}$. Additionally, we experiment with taking the hidden state from the last prompt token, denoted as $\text{CoIN}_\text{prompt}$.

\section{Gradient-space merging}
\label{sec:gradient}

Training on multiple prompt formulations can be viewed as a multitask learning problem. A large body of work has trained models on multiple tasks using gradient merging approaches and has shown that a naive sum or mean of these vectors underperforms due to interference \cite{yu2020gradient,wang2020gradient,chen2020just}. We predict similar failure modes for multi-prompt training, especially in the \textsc{All-in-One-Batch}-mode. We confirm this prediction by measuring interference in the per-template gradients (see \autoref{sec:gradient_interference}). Therefore, we use the interference-resolution tools developed for model and gradient merging to merge per-template gradients. 

We focus on per-parameter, sign-based merging methods like \cite{YadavTies2023, YuDare2024, lu2024online}. Since our per-template gradients have relatively high cosine similarity (see \autoref{sec:gradient_interference}), we do not use cosine methods like \cite{yu2020gradient,wang2020gradient}. Additionally, more complex subspace alignment methods, such as \textsc{KnOTS} \cite{stoica2025knots} are designed to run once after training and are too costly to directly port into the gradient space.

\paragraph{Setup.}
Consider a single-dataset batch drawn from $\mathcal{D}_i$ under the \textsc{All-in-One-Batch} construction, in which all formulations of its examples are present. Instead of averaging $\mathcal{L}_\text{CE}$ over the entire batch, we compute it per template. Using the backward pass, we obtain one gradient, $g_l \in \mathbb{R}^N$, over the trainable parameters $N$, per template. Stacking these gradients row-wise yields a gradient matrix $G = \bigl[\,g_l\,\bigr]_{l \in \mathcal{S}_i} \in \mathbb{R}^{T_i \times N}$. We define a gradient merging function $f : \mathbb{R}^{T_i \times N} \rightarrow \mathbb{R}^N$ that reduces $G$ to a single gradient. This gradient is then passed to the optimizer in place of the usual mean gradient.

This view recovers standard instruction fine-tuning as $f(G) = g_1$ and multi-template instruction fine-tuning, specifically \textsc{All-in-One-Batch} as the simple mean $f(G) = \tfrac{1}{T_i}\sum_{l} g_l$.

\paragraph{Memory and parallelization.}
Constructing $G$ raises the gradient memory requirements from $N$ to $T_i {\times} N$. To keep this tractable, we shard the gradients across devices along the parameter dimension $N$. With a world size of $W$, each device stores only $T_i\,(N / W)$ coordinates rather than the full matrix. We pair this with distributed data parallelism since each model replica fits on a single device. Gradient sharding is independent of other sharding methods and thus works with other distribution schemes. Increasing the effective batch size through data parallelism leaves the gradient dimensionality unchanged, ensuring effective GPU utilization for both gradient computation and merging. Moreover, sharding along $N$ is a natural fit for the merge itself, since we focus on per-parameter-based gradient merging.

\paragraph{\textsc{Prompt-TIES}.}
\textsc{Prompt-TIES} applies \textsc{TIES} \cite{YadavTies2023} to the rows of $G$ in three steps.
\begin{enumerate}
    \item \emph{Trim}: For each formulation, keep the top-$k\%$ gradients by magnitude and reset the rest to zero, $\hat g_l = \mathrm{trim}_k(g_l)$. Due to gradient sharding across $N$, we perform trimming per rank. This makes our trimming a communication-free approximation of the global top-$k\%$ selection.

    \item \emph{Elect sign}: Assign each trainable parameter the sign with the largest total gradient magnitude across formulations, 
    \begin{equation}
    \gamma = \mathrm{sgn}\Bigl(\textstyle\sum_{l \in \mathcal{S}_i} \hat g_l\Bigr) \in \{-1, 0, +1\}^N .
    \end{equation}
    \item \emph{Disjoint mean}: For each trainable parameter $p$, only average the formulations whose sign agrees with the elected sign, while ignoring the trimmed entries,
    \begin{equation}
    f(G)^{(p)} = \frac{1}{|A_p|} \sum_{l \in A_p} \hat g_l^{(p)},
    \qquad A_p = \bigl\{\, l \in \mathcal{S}_i \;\big|\; \mathrm{sgn}(\hat g_l^{(p)}) = \gamma^{(p)} \,\bigr\},
    \end{equation}
    with $f(G)^{(p)} = 0$ when the elected sign is zero (exact cancellation). 
\end{enumerate}

\textsc{TIES} additionally scales the merged vector by a factor of $\lambda$, which is tuned on a validation set \cite{YadavTies2023}. In the gradient space, this coincides with a rescaling of the learning rate, which should be considered when performing learning-rate sweeps. Additionally, tuning it on a validation set is too expensive. Therefore, we remove the scaling step, as in \cite{lu2024online}.

\paragraph{\textsc{Prompt-DARE}.}
\textsc{Prompt-DARE} replaces magnitude-based trimming with random trimming. Then, it merges with the other \textsc{TIES} steps \cite{YuDare2024}. According to \textcite{YuDare2024}, we independently drop each coordinate with probability $p$ and rescale the survivors to keep the gradient unbiased,

\begin{equation}
m_l \sim \mathrm{Bernoulli}(p)^N,
\qquad
\hat g_l = \frac{(\mathbf{1} - m_l) \odot g_l}{1 - p}.
\end{equation}

\section{Evaluation}
\subsection{Performance and robustness metrics}
\label{sec:perf_robust_metrics}

We evaluate each trained model on held-out datasets using two types of metrics: raw performance and robustness to the choice of prompt formulation. All metrics are based on per-example, per-template correctness scores $S_{i,j,l} = \mathrm{Score}\big(\mathcal{M}(x_{i,j,l}),\, y_{i,j,l}\big) \in [0,1]$. The scores are agnostic to the specific correctness function.

\paragraph{Performance.}
We report the macro-average score across datasets, giving each dataset an equal weight regardless of its size:

\begin{equation}
\mathrm{avg} = \frac{1}{n}\sum_{i=1}^{n} \frac{1}{k_i}\sum_{l=1}^{k_i} \frac{1}{m_i}\sum_{j=1}^{m_i} S_{i,j,l}.
\end{equation}

\paragraph{Robustness.} We measure how performance changes as formulations change, separating two sources of robustness. \textbf{Global variance} considers template averages across examples and determines which templates perform well or poorly. \textbf{Local variance} asks how much performance fluctuates for a single example across formulations. These two sources of robustness can diverge if certain prompt-example combinations are weak, but no single template is. Per-template averages can hide sensitivity exposed by a per-example view. We analyze the divergence on our datasets in \autoref{sec:globallocalresult}.

\paragraph{Global variance.}
We determine the best and worst templates for each dataset by averaging the performance of each example formatted with a specific template. These quantify how template phrasing affects model performance and have been used in previous work, such as \cite{cao2024worst}.

\paragraph{Local variance.}
We measure the variance in scores across templates for a fixed example by reversing the order of aggregation and averaging the variance across examples,

\begin{equation}
\mathrm{LV} = \frac{1}{n}\sum_{i=1}^{n} \frac{1}{m_i}\sum_{j=1}^{m_i} v_{i,j},
\qquad
v_{i,j} = \frac{1}{k_i}\sum_{l=1}^{k_i}\big(s_{i,j,l} - \bar s_{i,j}\big)^2,
\quad \bar s_{i,j} = \frac{1}{k_i}\sum_{l=1}^{k_i} s_{i,j,l}.
\end{equation}

The per-example variance is zero whenever the templates agree, including the degenerate case in which the model is incorrect for every template. These examples, which are always wrong, are not prompt-robust. They are uniformly unsolved, yet they contribute zero variance, which makes the model appear more stable than it is. Therefore, we call an example \textbf{solvable} if the model answers it correctly under at least one template. We create a set of solvable examples $\mathcal{J}_i^{\text{solv}}$ per dataset,

\begin{equation}
c_{i,j,l} = \mathbb{1}\big[s_{i,j,l} \ge \theta\big],
\qquad
\mathcal{J}_i^{\text{solv}} = \big\{\, j : \textstyle\sum_l c_{i,j,l} \ge 1 \,\big\},
\end{equation}
with $\theta$ being a threshold based on the concrete scoring function. Based on the \textbf{solvable} examples, we report the \textbf{conditional local variance}, restricting the average to solvable examples only,
\begin{equation}
\mathrm{LV}_\text{cond} = \frac{1}{n}\sum_{i=1}^{n}
\frac{1}{|\mathcal{J}_i^{\text{solv}}|}\sum_{j\in\mathcal{J}_i^{\text{solv}}} v_{i,j}.
\end{equation}
This isolates prompt sensitivity on the examples the model can demonstrably solve, under some template, instead of crediting it for failing consistently.

\paragraph{Relating the two.}
The span formally links global and local variance. The local span is defined as the per-example average gap between the best and worst templates: $\mathrm{span}^\text{loc}_i = \frac{1}{m_i}\sum_j (\max_l s_{i,j,l} - \min_l s_{i,j,l})$.The global span is defined as the gap between the best and worst templates at the dataset level: $\mathrm{span}^\text{glob}_i = \max_l\frac{1}{m_i}\sum_j s_{i,j,l} - \min_l \frac{1}{m_i}\sum_j s_{i,j,l}$. Under these definitions the global span is always bounded by the local span, $\mathrm{span}^\text{glob}_i \le \mathrm{span}^\text{loc}_i$, since the maximum of averages is at most the average of maxima. Therefore, the difference between the two is a direct measure of example-template interaction. It vanishes only when the best and worst templates are identical across all examples. We report the interaction fraction, defined as the share of per-example brittleness that cannot be attributed to a fixed template ranking: $\rho^{\text{int}}_i = 1 - \mathrm{span}^\text{glob}_i / \mathrm{span}^\text{loc}_i$.

\subsection{Gradient interference measures}
\label{sec:grad_interference_measure}
Unlike traditional training-time robustness approaches, our gradient-merging methods do not add an auxiliary objective that would cause the model to produce more consistent updates. Instead, we correct the interference found in the raw gradients. This renders an end-of-training diagnostic unsuitable for gradient analysis. First, our methods never reduce interference in the pre-merge-per-template gradients, so a converged model would not necessarily exhibit less interference. Second, the methods drift into different regions of parameter space during training, causing the gradients to become incomparable across runs. Therefore, we do not compare interference between models. Instead, we characterize the prevalence of gradient interference in the \textsc{All-in-One-Batch} setting on which our merging methods are based.

At each evaluation step, we compute the per-template gradient $g_l \in \mathbb{R}^N$, on the held-out batch without applying the resulting update and then stack it into the gradient matrix $G \in \mathbb{R}^{T_i \times N}$ of \autoref{sec:gradient}. We write $\bar g = \tfrac{1}{T_i}\sum_{l} g_l$ for the mean gradient update and $g_l^{(p)}$ for the $p$-th coordinate of $g_l$. We summarize the interference in $G$ with four geometry statistics, each averaged over the evaluation batches:\footnote{All four statistics are computed using the raw per-template gradients before learning-rate scaling or clipping. Since they are ratio statistics, they are invariant to overall scale.}
\begin{enumerate}
    \item \emph{Cosine similarity.} The pairwise mean cosine similarity is calculated over the $\binom{T_i}{2}$ unordered template pairs,
    \begin{equation}
    \mathrm{cos} = \binom{T_i}{2}^{-1} \sum_{l < l'} \frac{g_l^\top g_{l'}}{\lVert g_l\rVert\,\lVert g_{l'}\rVert} \in [-1, 1].
    \end{equation}
    It measures whether the formulations agree on a common descent direction. Cosine similarity is equal to $1$ when all gradients are collinear and the mean is a faithful summary, it is $0$ when the gradients are orthogonal, and it is negative when the gradients oppose one another. This has been widely used in previous gradient merging literature, such as \cite{yu2020gradient, wang2020gradient}.
    
    \item \emph{Sign disagreement.} The fraction of parameters for which at least two formulations push in opposite directions,
    \begin{equation}
    \mathrm{sign\text{-}conflict} = \frac{1}{N}\sum_{p=1}^{N} \mathbb{1}\Bigl[\, \exists\, l, l' :\; \mathrm{sgn}(g_l^{(p)}) > 0 \;\wedge\; \mathrm{sgn}(g_{l'}^{(p)}) < 0 \,\Bigr] \in [0, 1].
    \end{equation}
   It measures sign conflict, the quantity targeted by the sign-election step of \textsc{Prompt-TIES} and \textsc{Prompt-DARE} (\autoref{sec:gradient}). Sign conflict is $0$ when every parameter receives a unanimous sign across formulations and $1$ when every parameter is contested. 
    
    \item \emph{Cancellation.} Cosine similarity and sign disagreement measure \emph{whether} gradients conflict. Cancellation measures \emph{how much} magnitude the conflict costs. The fact that updates lose magnitude when averaged is a long-recognized symptom of interference \cite{YadavTies2023, YuDare2024, yu2020gradient}. Since there is no standard statistic for this phenomenon, we directly quantify the effect as the portion of the average per-template gradient norm that the mean update fails to retain.

    \begin{equation}
    \mathrm{cancellation} = 1 - \frac{\lVert \bar g\rVert}{\tfrac{1}{T_i}\sum_{l}\lVert g_l\rVert} \in [0, 1].
    \end{equation}

    When all formulations are identical, there is no cancellation, so averaging preserves the full norm. Cancellation approaches $1$ when the per-template gradients cancel each other out, causing the mean update to collapse toward zero.
    
    \item \emph{Effective rank.} The number of distinct directions that the template gradients span is computed as the entropy-based effective rank \cite{roy2007effective} of the cosine similarity Gram matrix $\hat G = [\cos(g_l, g_{l'})]_{l,l'} \in \mathbb{R}^{T_i \times T_i}$. The eigenvalues $\lambda_1, \dots, \lambda_{T_i} \ge 0$ are normalized to $p_r = \lambda_r / \sum_s \lambda_s$. The effective rank is:
    
    \begin{equation}
    \mathrm{eff\text{-}rank} = \exp\Bigl(-\sum_{r=1}^{T_i} p_r \log p_r\Bigr) \in [1, T_i].
    \end{equation}
    
    It is equal to one when all per-template gradients share the same update direction. It grows toward $T_i$ as the formulations spread into mutually distinct directions that cannot be satisfied by a single update. This metric has been used in previous work analyzing gradients, as in \cite{gauch2022few}.
\end{enumerate}

\subsection{Auxiliary objective diagnostics}
\label{sec:aux_metrics}

We compute two per-example diagnostics to check whether the auxiliary objectives of \textsc{PPCL} \cite{qiang-etal-2024-ppcl} and \textsc{CoIN} \cite{yan2024contrastive} produce their intended effects on the model's outputs and hidden states. Both are evaluated in the \textsc{All-in-One-Batch} setup using the \textsc{majority templates} (see \autoref{sec:data_construction}). We report the average of each diagnostic over the evaluation examples in each dataset and over all datasets.

\paragraph{Jensen--Shannon divergence.}
We average the pairwise Jensen--Shannon divergence over all majority templates of an example, mirroring \textsc{PPCL} objective (see \autoref{form:PPCL_loss}) without the scaling term $\lambda_3$.

\begin{equation}
\mathrm{JS}^{\bullet}(i,j) = \binom{|\mathcal{S}_i^\star|}{2}^{-1}
\sum_{\{l,l'\}\subseteq \mathcal{S}_i^\star}
\mathrm{JS}\left(q^{\bullet}_{i,j,l} \,\big\|\, q^{\bullet}_{i,j,l'}\right),
\qquad \bullet \in \{\text{label}, \text{prompt}\},
\end{equation}

where $q^{\bullet}_{i,j,l}$ is the readout distribution:
\begin{itemize}
  \item \texttt{label}: $q^\text{label}_{i,j,l} = \bar p_{i,j,l}$. As in \textsc{PPCL}, the \texttt{label} readout takes the output distribution averaged over the response tokens. It depends on the reference target  $y_{i,j}$ and is therefore not an inference-time signal. Teacher forcing re-anchors the model on the correct prefix at each step. Thus, the metric captures only per-step agreement along the gold path. This suppresses the divergence that compounds under autoregressive decoding. Consequently, it can appear consistent even when the models' generated outputs differ. This makes it an optimistic proxy that we pair with the label-free \texttt{prompt} readout.
  \item \texttt{prompt}: $q^\text{prompt}_{i,j,l} = p_{\mathcal{M}}(\,\cdot \mid x_{i,j,l})$. Takes the distribution of the first response token given the prompt alone. This is truly label-free, but can only answer whether the model begins to answer the same way across formulations.
\end{itemize}

\paragraph{Hidden-state similarity.}
To probe the effect of \textsc{CoIN} (see \autoref{sec:coin}), we compare the hidden states of the last layers across templates. Since \textsc{CoIN}'s batch construction is not used in evaluation, we adapt its positive/negative contrast to the \textsc{All-in-One-Batch} majority-template setup shared with \textsc{PPCL}. Let $h^\bullet_{i,j,l} = h_{\mathcal{M}}(x_{i,j,l}, y_{i,j,l})$ be the hidden state for a prompt-label pair, and let $\mathcal{E}_\mathcal{B}$ be the distinct examples of dataset $i$ in the batch. For each example, we report the contrastive \emph{difference} between the same-example cross-template similarity (\textsc{CoIN}'s positive) and the cross-example similarity (\textsc{CoIN}'s negative):
\begin{equation}
\mathrm{Diff}^{\bullet}(i,j) =
\binom{|\mathcal{S}_i^\star|}{2}^{-1}
\sum_{\{l,l'\}\subseteq \mathcal{S}_i^\star}
\left[
\mathrm{sim}\big(h^{\bullet}_{i,j,l}, h^{\bullet}_{i,j,l'}\big)
-  \frac{1}{|\mathcal{E}_\mathcal{B}|-1}\sum_{j'\neq j} \mathrm{sim}\big(h^{\bullet}_{i,j,l}, h^{\bullet}_{i,j',l'}\big)
\right],
\end{equation}

where $\mathrm{sim}$ is the cosine similarity of \autoref{sec:coin}. A larger $\mathrm{Diff}$ means that formulations of the same example with different templates are closer to one another than to other examples, the alignment \textsc{CoIN} targets. As with \textsc{CoIN} we have two different readout variants $\bullet \in \{\text{label}, \text{prompt}\}$, either the last target token ($\mathrm{Diff}^{\text{label}}$) or the last prompt token ($\mathrm{Diff}^{\text{prompt}}$). Unlike the \texttt{label}, the \texttt{prompt} readouts are directly comparable between the $\mathrm{Diff}$ and $\mathrm{JS}$.

\chapter{Experimental setup}
\section{Data}
\label{sec:data}

Studying whether training can make a model prompt-robust requires two things of the data. First, each example must be available in several semantically equivalent formulations. Second, evaluation must probe genuinely unseen tasks since robustness is only useful if it transfers beyond the training distribution. The setup below satisfies both conditions. A full overview of the datasets and their templates is provided in \autoref{app:dataset}.

\paragraph{Datasets.}
We train on 48 datasets and evaluate on 11 strictly unseen test datasets. The training tasks cover a wide variety of topics, including question answering, summarization, sentiment classification, and topic classification. The held-out tasks include natural language inference, word sense disambiguation, coreference resolution, and commonsense reasoning. Since no evaluation task appears during training, test performance reflects task-level generalization rather than memorization. We adopt this setup because it is frequently used for work on prompt-agnostic language models \cite{zhou-etal-2022-prompt, yan2024contrastive, hejabi2025flipflopconsistencyunsupervisedtraining, wei2025paft}.

\paragraph{Templates.}
We use PromptSource \cite{bach2022promptsource} as the source of prompt templates. PromptSource provides a set of human-written templates covering paraphrastic and structural variation for every dataset. \textcite{chatterjee-etal-2024-posix} show that models are most sensitive to these perturbation types. Each dataset contains between 4 and 15 templates (see \autoref{app:dataset}).

\paragraph{Preprocessing.}
We download all datasets from the HuggingFace Hub \cite{lhoest-etal-2021-datasets}, except for StoryCloze \cite{mostafazadeh2016corpus}, which we obtain from the original release.\footnote{\url{https://cs.rochester.edu/nlp/rocstories/}, last accessed 18.06.2026.} Three training datasets were unavailable in the required version and were replaced: tydiqa \cite{tydiqa} with coqa\_expanded, wikihop \cite{welbl-etal-2018-constructing} with riddle\_sense \cite{lin-etal-2021-riddlesense}, and gigaword \cite{parkerrobertEnglishGigawordFifth2011} with scitldr \cite{cachola2020tldr}.

We manually filter every template against three criteria:
\begin{itemize}
  \item \emph{Applicability.} The template must be compatible to the dataset currently distributed on HuggingFace. We found that several templates were not applicable to the currently available version. We discard these rather than attempting to repair them.
  \item \emph{Task consistency.} The template must preserve the task type of the dataset. Some templates convert an example into a different task. For example, the sentence-completion dataset HellaSwag \cite{zellers2019hellaswag} is reformulated as a topic classification task. Since HellaSwag is an evaluation dataset and topic classification is a training task, such a template would leak training tasks into the test set.
  \item \emph{Solvability.} The template must be solvable without any task-specific context. For example, one template from the topic classification dataset TREC \cite{li-roth-2002-learning}, expects "other location" as the answer to the prompt "What is this question asking for? Where is the Eiffel Tower?" Such unsolvable templates penalize every model, regardless of capability, and would artificially lower worst-case scores while inflating apparent prompt sensitivity. We remove them.
\end{itemize}

After filtering, we discard any dataset with fewer than two valid templates since at least two formulations are required to measure prompt variation. Finally, to balance the contribution of each task, we cap the size of each training set at 10{,}240 examples and the size of each validation set at 512. The test sets remain at full size. The training set contains 417,238 individual examples.

\paragraph{Data splits.}
Validation reuses the training datasets, tasks, and templates while holding out examples, which isolates generalization to unseen inputs. The test split introduces unseen tasks, each with its own templates and examples. Therefore, it measures the out-of-distribution generalization that we ultimately care about. Reporting on both allows us to distinguish in-task generalization from the more challenging cross-task generalization. To improve performance during training, we compute the validation loss only on a subset of 4,096 examples. This set is used for early stopping and gradient interference calculations. After training, we evaluate using the full validation and test splits. We report the results for three splits: the test split evaluated on all templates (\testall{}), the test split evaluated on only the majority templates (\testmaj{}), and the validation split evaluated on only the majority templates (\valmaj{}).

\section{Training}

\paragraph{Models.}
We evaluate the main methods using four base models that vary along two dimensions, model family and scale: Llama3.2-1B and Llama3.1-8B \cite{grattafiori2024llama3herdmodels}, and Qwen3-0.6B and Qwen3-8B \cite{yang2025qwen3technicalreport}. These models allow us to test whether our findings generalize across architectures and whether robustness effects hold across model capacity. We chose Llama and Qwen for three reasons: (1) Their comparable and modest sizes allow for controlled comparisons and repeated experimentation. (2) Both are strong across a wide range of domains, increasing the likelihood that the models can solve the tasks \cite{grattafiori2024llama3herdmodels, yang2025qwen3technicalreport}. (3) They have been used in prior work on prompt-agnostic language models \cite{wei2025paft} and model merging \cite{stoica2025knots}. We use the base variant for every model to ensure that any observed differences in robustness are attributable to our training methods rather than prior instruction tuning and to limit the risk of test-set contamination.

We fully fine-tune the two small models. We apply LoRA \cite{hu2021loralowrankadaptationlarge} to the two 8B models. LoRA reduces computational cost and keeps the number of trainable parameters, and thus the dimensionality of the gradient matrix, manageable for the gradient-merging methods. All models are loaded through HuggingFace Transformers \cite{wolf-etal-2020-transformers}.

\paragraph{Instruction-tuning setup.}
We use TorchTune \cite{torchtune} for instruction tuning. Unless otherwise stated, each method is trained and validated on the majority of templates in each dataset and tested on all templates. \textsc{CoIN} and \textsc{PPCL} require label-consistent majority templates during training. Restricting all methods to the same majority templates allows us to compare them to data-construction approaches without introducing differences in template variance into the comparison. Additionally, since \textsc{CoIN} needs the majority label class to be at least three, we drop all datasets with fewer than three majority labels in majority training. We select the checkpoint with the best average performance and implement early stopping with a patience of 2.

\paragraph{Hyperparameters.}
We inherit the base hyperparameters from each model's TorchTune configuration and only vary the learning rate. Starting from default of $2{\times}10^{-5}$ in TorchTune, we sweep $\{10^{-5}, 3{\times}10^{-5}, 5{\times}10^{-5}, 7{\times}10^{-5}, 9{\times}10^{-5}, 10^{-4}\}$ for each combination of $\mathcal{L}_\text{MIFT}$ and model, giving 20 sweeps in total. The selected learning rates per method are reported in \autoref{tab:lr}. Since all data-construction variants share the standard cross-entropy loss, we perform a single sweep per model for them on \textsc{Single} using each dataset's first template. Each sweep runs for the first 2,000 steps. Afterwards, we keep the learning rate with the best validation checkpoint.

Due to computational constraints we do not run a separate learning-rate sweep for the gradient-merging methods \textsc{Prompt-TIES} and \textsc{Prompt-DARE}.  Instead, we set their learning rate to the median of the learning rates selected for the other methods on the same model. The only hyperparameter we add for \textsc{Prompt-TIES} and \textsc{Prompt-DARE}  is the amount of gradient we trim. We set this to $20\%$ of the gradients, as this value performed well in initial experiments. For \textsc{CoIN} we adopt the values proposed by its authors: $\tau = 0.05$ and $\lambda = 1000$. \textsc{PPCL} reports no stable defaults, so we tune $\lambda_3$, the weight of the Jensen--Shannon term, between values of $\{0.1, 1, 10, 100\}$. We run this sweep on Llama3.2-1B at a learning rate of $5{\times}10^{-5}$ for 2,000 steps. We select $\lambda_3=1$. The full set of hyperparameters is listed in \autoref{tab:shared_hyp}.

We train all methods on the same base examples for one epoch. Thus, the multi-template strategies take $|\mathcal{S}_i|$ times as many optimizer steps per dataset compared to \textsc{Single}. We deliberately do not match compute across runs to reduce the amount of computing power used in our experiments, as additional epochs would merely re-traverse already-seen examples rather than supply new gradient information. Moreover, initial testing showed no performance gains from longer training. Finally, prior work has shown that multi-template training outperforms single-template training, consistent with our findings \cite{sanh2022t0, wei2025paft}.

\begin{table*}[ht]
  \centering
  \caption{Shared hyperparameters for all experiments. $^\dagger$LoRA is only applied to the 8B models.}\label{tab:shared_hyp}
  \begin{tabular}{@{}ll@{}}
    \toprule
    \textbf{Hyperparameter} & \textbf{Value} \\
    \midrule
    Optimizer               & AdamW (fused) \\
    LR schedule             & Constant with linear warmup \\
    Warmup steps            & 100 \\
    Effective batch size    & 256\footnote{Because templates and label spaces differ across datasets, each micro-batch is drawn from a single dataset. We use a gradient-accumulation factor of $4$ to mix datasets within an optimizer step.} \\
    Epochs                  & 1 \\
    Precision               & bfloat16 \\
    Max train samples / dataset & 10\,240 \\
    \midrule
    LoRA rank$^\dagger$     & 16 \\
    LoRA alpha$^\dagger$    & 32 \\
    LoRA dropout$^\dagger$  & 0.0 \\
    LoRA target modules$^\dagger$ & attention (\texttt{q\_proj}, \texttt{v\_proj}, \texttt{o\_proj}) and MLP\footnote{\LlamaB{} additionally applies LoRA to the output embedding (\texttt{lm\_head}); Qwen3-8B does not.} \\
    \midrule
    COIN $\tau$             & 0.05 \\
    COIN $\lambda$          & 1000 \\
    \textsc{PPCL} $\lambda_1$, $\lambda_2$ & 1, 1 \\
    \textsc{PPCL} $\lambda_3$        & 1 \\
    \midrule
    \textsc{Prompt-TIES} $k$ & 0.8 \\
    \textsc{Prompt-DARE} $p$ & 0.2 \\
    \bottomrule
  \end{tabular}
\end{table*}

\begin{table}[ht]
  \centering
  \caption{Learning rates selected per model and method. We sweep one learning rate per $\mathcal{L}_\text{MIFT}$ and model combination.}\label{tab:lr}
  \begin{tabular}{lcccc}
    \toprule
    \textbf{Method} & \LlamaI{} & \LlamaB{} & \QwenI{} & \QwenB{} \\
    \midrule
    \textbf{Standard Loss}        & $5{\times}10^{-5}$ & $5{\times}10^{-5}$ & $3{\times}10^{-5}$ & $1{\times}10^{-5}$ \\
    \textbf{\textsc{CoIN}}        & $7{\times}10^{-5}$ & $9{\times}10^{-5}$ & $5{\times}10^{-5}$ & $5{\times}10^{-5}$ \\
    \textbf{\textsc{PPCL}}        & $3{\times}10^{-5}$ & $1{\times}10^{-4}$ & $1{\times}10^{-5}$ & $1{\times}10^{-5}$ \\
    \textbf{\textsc{Prompt-TIES}} & $5{\times}10^{-5}$ & $9{\times}10^{-5}$ & $3{\times}10^{-5}$ & $1{\times}10^{-5}$ \\
    \textbf{\textsc{Prompt-DARE}} & $5{\times}10^{-5}$ & $9{\times}10^{-5}$ & $3{\times}10^{-5}$ & $1{\times}10^{-5}$ \\
    \bottomrule
  \end{tabular}
\end{table}

\paragraph{Generation.}
For metrics that rely on generated text, we use a greedy decoder with a temperature of zero and no sampling. This makes the outputs deterministic, so differences across prompts reflect model variance rather than decoding noise.

\paragraph{Resources.}
We train all models using NVIDIA H100 80 GB HBM3 GPUs. The 8B models require four of these GPUs for training, while the smaller models require one. A single multi-template training takes six to ten hours. In total, we trained for 3,000 GPU hours.

\section{Evaluation}

\subsection{Text-scoring metrics}
\label{sec:text_scoring_metrics}
Our main metric is \textbf{Rouge-L} \cite{lin-2004-rouge}, which measures the longest common subsequence overlap between the model's generated answer and the reference. Following \textcite{wang2022supernaturalinstructionsgeneralizationdeclarativeinstructions}, we adopt it as the standard score for instruction following on this kind of multi-task benchmark. As a secondary metric, we use \textbf{rank classification accuracy}, which scores each candidate answer option based on its log-likelihood and selects the option with the highest score. This metric was also used in \cite{sanh2022t0}.

The T0-benchmark was originally evaluated using rank classification accuracy. Therefore, making the generative Rouge-L our primary metric is a deliberate choice. First, rank classification accuracy does not reflect how these models are used. Through teacher forcing, the model is handed the candidate set and is asked only for a relative ordering of it. However, a deployed model generates an answer without a candidate set provided. Therefore, robustness measured under rank classification accuracy does not certify the robustness of the system as it is actually deployed. Second, and more importantly for our purpose, rank classification accuracy can mask the sensitivity we set out to measure. It only requires that the correct option remains marginally more likely, while the model's generative behavior --- which token it emits first, the form of the answer, and whether it follows the instruction at all --- shifts substantially across reformulations. Thus, a model can appear prompt-robust under rank classification accuracy yet be brittle in generation. Third, as a constrained task it leaves little headroom on capable base-models. Rank accuracy quickly approaches the ceiling (see \autoref{sec:res_rank}), which compresses the differences between methods that our comparison depends on. Fourth, rank classification accuracy only works for the test split of the dataset because the training and validation splits incorporate many templates without concrete answer sets, for example, all summarization tasks. Nevertheless, we report both for the test set. Rank classification allows us to be comparable to the source benchmark and serves as a means of verifying whether our generation-based conclusions are transferable.

\subsection{Robustness metrics}
We calculate all robustness and diagnostic metrics based on text scoring metrics. For conditional local variance, we select a threshold of $\theta = 0.375$ for Rouge-L and a threshold of 1 for rank classification accuracy. As shown in \autoref{app:clv_threshold}, the choice of this hyperparameter does not change the ranking of the best-performing methods in our results. Additionally, we observe that a higher threshold increases the \clv{} for all methods because we remove examples that perform poorly across all paraphrases and thus have low variance. With this in mind, we chose a threshold of $\theta = 0.375$ as a trade-off between not rewarding poor performance and including as many examples as possible in the calculation.

\subsection{Statistical significance tests}
\label{exp_set:stats}
We assess whether the top-ranked method is significantly better than each competing method for each metric and summary statistic using a paired, one-sided bootstrap test. For a given pair of runs, we first reduce each example to the per-example means across templates for the average statistics, the per-example variance for the local-variance statistics, or the scores of the best- or worst-performing template per dataset for the best- or worst-template statistics. Then, examples are paired across the two runs. We draw $B = 9999$ bootstrap replicates. In each replicate, we resample the examples with replacement within each dataset. Then, we compute the dataset-level mean difference between the two runs and macro-average these differences across datasets. From the bootstrap distribution, we form a one-sided pivotal confidence interval and declare the top run to be significantly better if the lower bound exceeds zero.

We report significance at $\alpha = 0.05$ and correct it for multiple comparisons within each metric using the Bonferroni correction, giving an effective per-test threshold of $\alpha = 0.005$.  For each metric, we report the \texttt{best group}: the top-ranked run and every run that is not significantly worse than it under this test.

\section{Baselines}
\label{sec:baselines_def}
We compare \textsc{Prompt-TIES} and \textsc{Prompt-DARE} to eight baselines. Six of these are data construction approaches, as described in \autoref{sec:data_construction}. \textsc{Single} first template, \textsc{Single} second template, and \textsc{Single} third template are single-template runs that always select the first, second, or third template of every dataset, respectively. We fix the templates because some datasets from the same family share templates. We evaluate the impact of template choice in \autoref{sec:res_template_count_and_choice}. In addition, we include the three multi-template data construction approaches \textsc{All shuffled}, \textsc{One-at-a-Time}, and \textsc{All-in-One-Batch}, which are also described in \autoref{sec:data_construction}. The remaining two baselines are \textsc{PPCL}, as described in \autoref{sec:PPCL} and \textsc{CoIN}, as described in \autoref{sec:coin}. For \textsc{CoIN} we always use the \texttt{label} readout version, following the original CoIN paper \cite{yan2024contrastive}. We analyze the impact of the prompt readout version in \autoref{sec:res_coin_diag}.

\chapter{Results}
\label{ch:results}

This chapter evaluates \textsc{Prompt-TIES} and \textsc{Prompt-DARE} against the baselines described in \autoref{sec:baselines}. We structure it around the contributions laid out in the introduction (see \autoref{intro}). First, we analyze how to measure prompt brittleness (\autoref{sec:globallocalresult}) and find that it is best captured as local, example-level variance since the template-example interaction rather than a fixed template ranking dominates. Next, we characterize gradient interference in multi-prompt instruction fine-tuning (\autoref{sec:gradient_interference}), demonstrating that it is driven by per-parameter sign conflict rather than by cosine misalignment. These findings motivate our proposal of \textsc{Prompt-TIES} and \textsc{Prompt-DARE}, which we then compare to our baselines in terms of performance and robustness (\autoref{sec:res_frontier}). Our data construction methods achieve the highest accuracy, and our gradient-merging methods provide the greatest local robustness for fully fine-tuned models. We probe how far these conclusions hold (\autoref{sec:res_checks}) across evaluation splits, metrics, template counts, and model scale. Finally, we diagnose why \textsc{PPCL} and \textsc{CoIN} underperform (\autoref{sec:res_diag}), tracing each failure to a lack of transfer between the training and test distributions.

\section{Local versus global variance}
\label{sec:globallocalresult}
Prior work quantifies prompt brittleness through the best and worst template of a dataset, providing a global view of robustness \cite{cao2024worst, sclar2024quantifying}. However, we argue that this view is incomplete. If brittleness arises from the interaction between a template and a specific example, rather than from the template itself, then a dataset-level ranking hides most of the sensitivity. This section tests this hypothesis and establishes the local, example-level view adopted throughout the rest of the chapter.

We quantify the interaction through the interaction fraction $\rho^{\text{int}}_i$ defined in \autoref{sec:perf_robust_metrics}. It is the share of per-example brittleness that cannot be explained by a fixed template ranking. A value of zero means that the same template is always the best and another template is always the worst, so the global view remains unchanged. A value approaching one means that the best and worst templates are almost entirely example-specific.

On average, across datasets and models $\rho^{\text{int}}$ is $54\%$ and exceeds $70\%$ on ANLI and WiC (see \autoref{tab:interaction-fraction}). On average, this means that a global best-versus-worst comparison discards more than half of the observed brittleness. The lone exception is HellaSwag ($15\%$), where a few templates perform poorly across all examples. We will return to this in \autoref{sec:res_gradient}, as HellaSwag is the dataset on which our gradient methods do not achieve the lowest \clv{}. The effect weakens with scale, dropping from $56$ to $50$ within the Llama family and from $66$ to $41$ within Qwen, consistent with the regime shift analyzed in \autoref{sec:res_scale}.

Two consequences follow. First, reporting only the best and worst templates understates prompt sensitivity. Therefore, we adopt the \clv{} as our primary robustness metric and only retain the best- and worst-template scores for comparability with previous work. Second, since brittleness is an example-level interaction rather than a property of individual templates, a method that merely reselects or reweights templates cannot eliminate it. The method must act at the level of the individual example. This is precisely what our per-prompt gradient merging does (see \autoref{sec:gradient} and \autoref{sec:res_gradient}).

\begin{table}[htbp]
\centering
\caption{Interaction fraction $\rho^{\text{int}}_i = 1-\mathrm{span}^{\text{glob}}_i/\mathrm{span}^{\text{loc}}_i$ on the test split, averaged over all training runs. High values indicate that the best/worst template is example-specific. Low values indicate uniformly good/bad templates. $k$ is the number of templates.}
\label{tab:interaction-fraction}
\begin{tabular}{lrrrrr}
\toprule
 & & \multicolumn{2}{c}{Llama} & \multicolumn{2}{c}{Qwen} \\
\cmidrule(lr){3-4}\cmidrule(lr){5-6}
Dataset & $k$ & 1B & 8B & 0.6B & 8B \\
\midrule
ANLI-r1    & 15 & 73 & 79 & 88 & 50 \\
ANLI-r2    & 15 & 73 & 81 & 90 & 59 \\
ANLI-r3    & 15 & 77 & 77 & 88 & 60 \\
CB         & 15 & 36 & 31 & 52 & 15 \\
COPA       & 7  & 72 & 37 & 80 & 26 \\
HellaSwag  & 5  & 21 & 7  & 18 & 15 \\
RTE        & 9  & 65 & 70 & 63 & 55 \\
StoryCloze & 5  & 44 & 71 & 75 & 29 \\
WSC        & 10 & 46 & 30 & 54 & 38 \\
WiC        & 9  & 80 & 78 & 93 & 51 \\
WinoGrande & 6  & 38 & 26 & 21 & 47 \\
\midrule
\textbf{Macro-average} & & \textbf{56.8} & \textbf{53.4} & \textbf{65.6} & \textbf{40.5} \\
\bottomrule
\end{tabular}
\end{table}

\section{Per-template gradients interfere}
\label{sec:gradient_interference}

We measure gradient interference on two models from the Llama family that were trained using the \textsc{All-in-One-Batch} setting. For both models, we take measurements every 1,000 optimizer steps on a subset of the validation split. Since we do not optimize for these metrics, we report the average over training. For \LlamaB{} we do not include timestep 0, because the LoRA weights are initialized with zero, and evaluation is done before training. The full raw scores can be found in \autoref{app:gradient_interference}.

\paragraph{Gradients interfere in sign.}
On \LlamaI{}, the mean pairwise cosine similarity between formulations is $0.543$, which is high relative to the similarity of task gradients, as reported in \textcite{yu2020gradient} (see \autoref{tab:grad_interference_steps}). Even the most conflicting template pair drops only to $0.294$. Since cosine similarity is always positive, a cosine-based intervention such as PCGrad \cite{yu2020gradient} would not bring the desired results.

However, at the coordinate level the template gradients disagree sharply in sign. At least two formulations push in opposite directions on $57.2\%$ of the trainable parameters, which is exactly the conflict that sign-based merging methods (\autoref{sec:gradient}) are designed to resolve. Plain averaging of the per-template gradients reduces their magnitude by $19.7\%$, and the gradients have an effective rank of $\approx 3$. Together, these metrics show that the formulations span several competing update directions rather than one shared direction. This pattern remains consistent throughout training, with sign conflict ranging from $56\%$ to $62\%$ and cancellation ranging from $19\%$ to $22\%$.

The LoRA adapters and size of \LlamaB{} do not eliminate this interference. The mean pairwise cosine similarity ($0.549$, most conflicting pair $0.286$), the cancellation ($17.5\%$) and the effective rank ($\approx 3$) are comparable to the \LlamaI{} values. The coordinate-level sign disagreement is slightly higher at $63\%$. Overall, the diagnostics confirm that the interference is comparable in severity for both Llama models. However, \textsc{Prompt-TIES} and \textsc{Prompt-DARE} still fail to successfully resolve gradient interference for \LlamaB{} (see \autoref{sec:res_gradient}).  We interpret these results as indicating that LoRA updates share the same interference but do not share a common coordinate system. Thus, a coordinate-wise merge cannot act on them cleanly \cite{stoica2025knots}. As with \LlamaI{}, the findings are stable across training.

\begin{figure*}[tp]
    \centering
    \includegraphics[width=\textwidth]{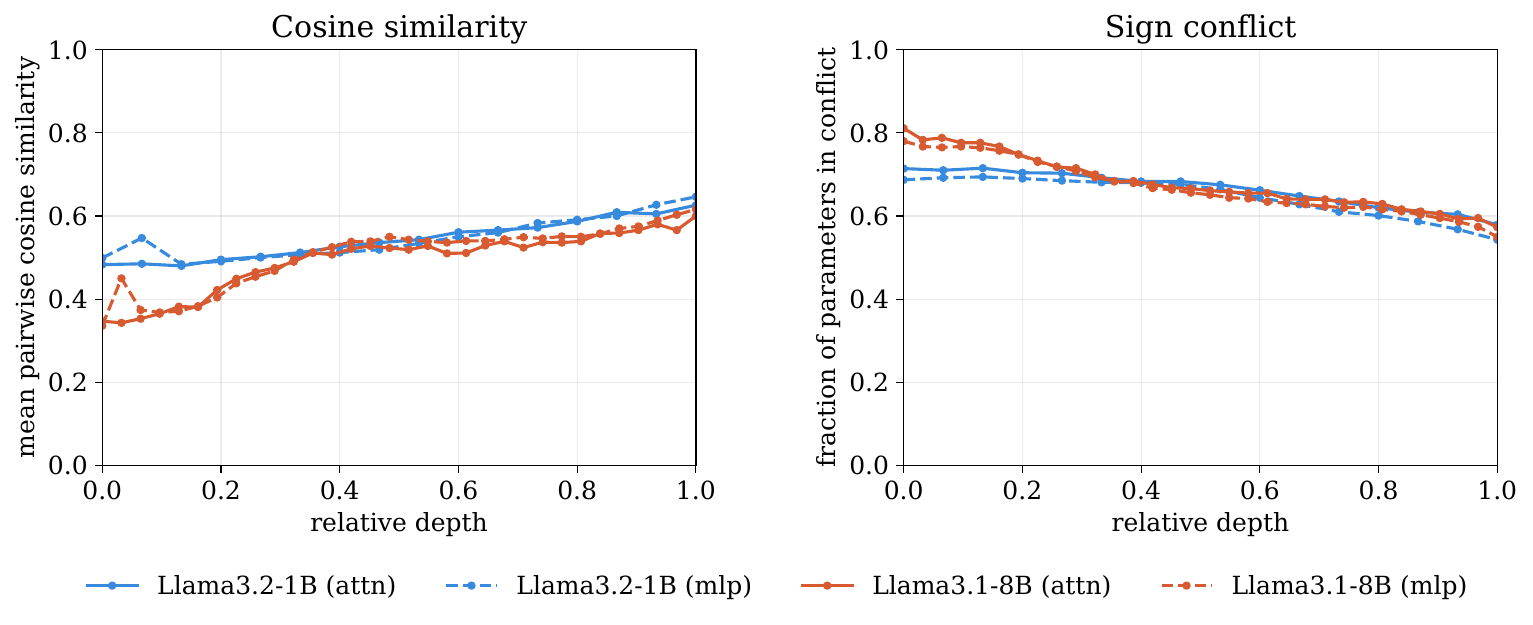}
    \caption{\textbf{Gradient interference declines with depth.} The mean pairwise cosine similarity (left) and sign conflict (right) of the per-template gradients are shown by the relative depth of the transformer block. We separate attention and MLP sublayers. The metrics are averaged over evaluations every 1000 optimizer steps on a subset of the validation split during \textsc{All-in-One-Batch} training. Raw scores can be found in \autoref{tab:grad_interference_layers_1b} and \autoref{tab:grad_interference_layers_8b}.}
    \label{fig:gradient_depth}
\end{figure*}

\paragraph{Interference concentrates in the early layers.}
Resolving the gradients by module shows that the conflict is not uniform across the network (see \autoref{fig:gradient_depth}). Cosine similarity is lowest and sign conflict highest in the first transformer blocks. Then the interference decreases with depth. This effect is stronger in \LlamaB{}, where the first layers have higher interference. The gap in absolute sign conflict and cosine similarity between the first and last layers is $23.4$ percentage points and $0.266$, respectively. Across the later layers, the difference in interference between the models flattens out. We hypothesize that with growing abstraction from the surface form of the prompt the interference reduces. We did not test whether a layer-specific merging rule would improve the results. Interference is roughly the same in the attention and MLP layers. The attention layers only have marginally more interference. Therefore, we do not prioritize either part during gradient interference removal.

\section{The performance--robustness frontier}
\label{sec:res_frontier}

\begin{figure*}[tp]
    \centering
    \includegraphics[width=\textwidth]{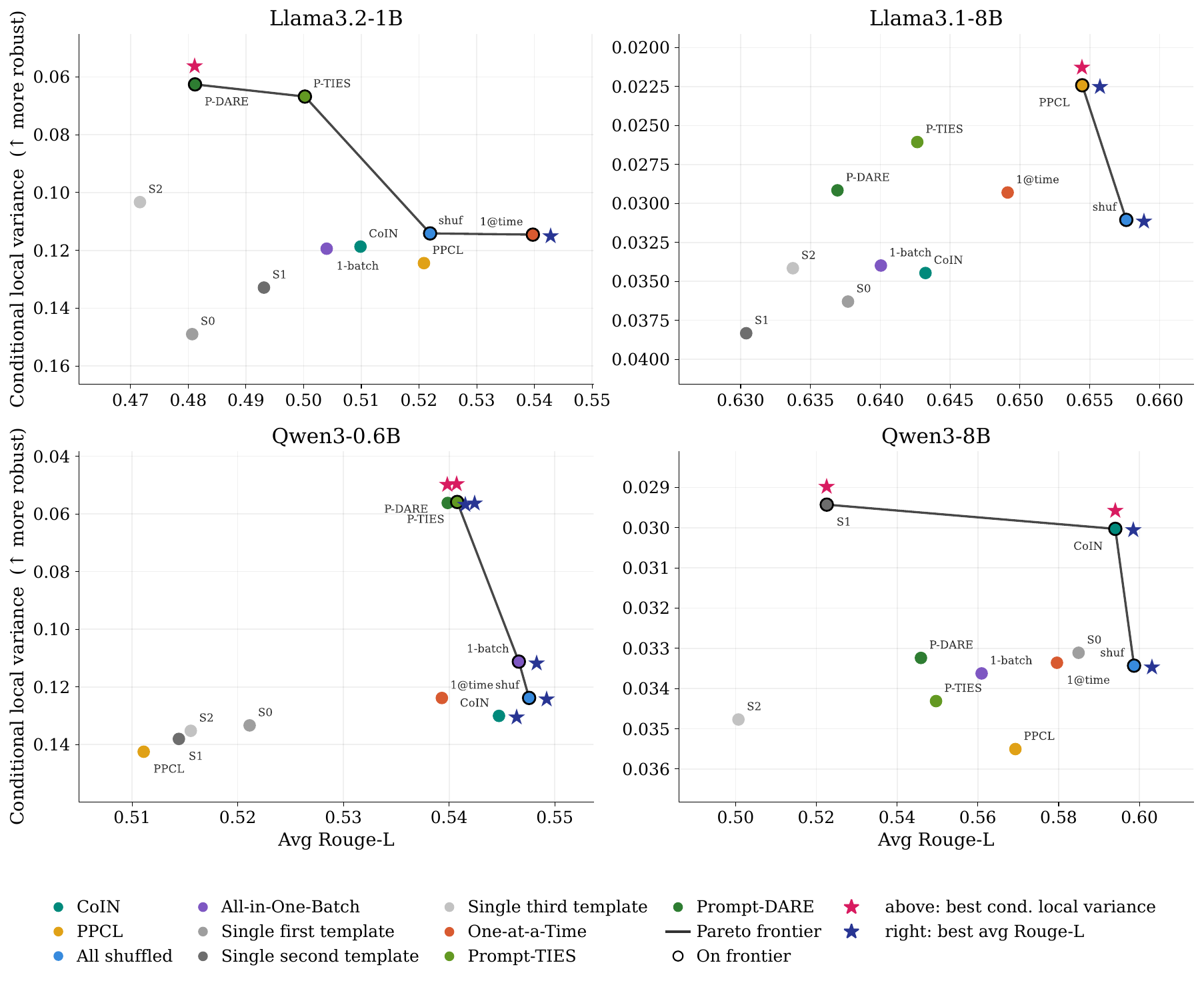}
    \caption{\textbf{Performance--robustness frontier.} Each point represents a training method and is positioned by the mean Rouge-L ($x$, $\rightarrow$ better) and \clv{} ($y$, $\uparrow$ better). The scores are from the \testall{} split. The Pareto-optimal set is outlined in black and connected by a solid line. The stars indicate the statistically best group per model: the \textcolor[HTML]{D81B60}{pink} star for \clv{} and the \textcolor[HTML]{283593}{indigo} star for average Rouge-L. \textsc{Prompt-TIES} and \textsc{Prompt-DARE} are the most robust on the fully-finetuned models.}
    \label{fig:frontier_test}
\end{figure*}

\begin{figure*}[tp]
    \centering
    \includegraphics[width=\textwidth]{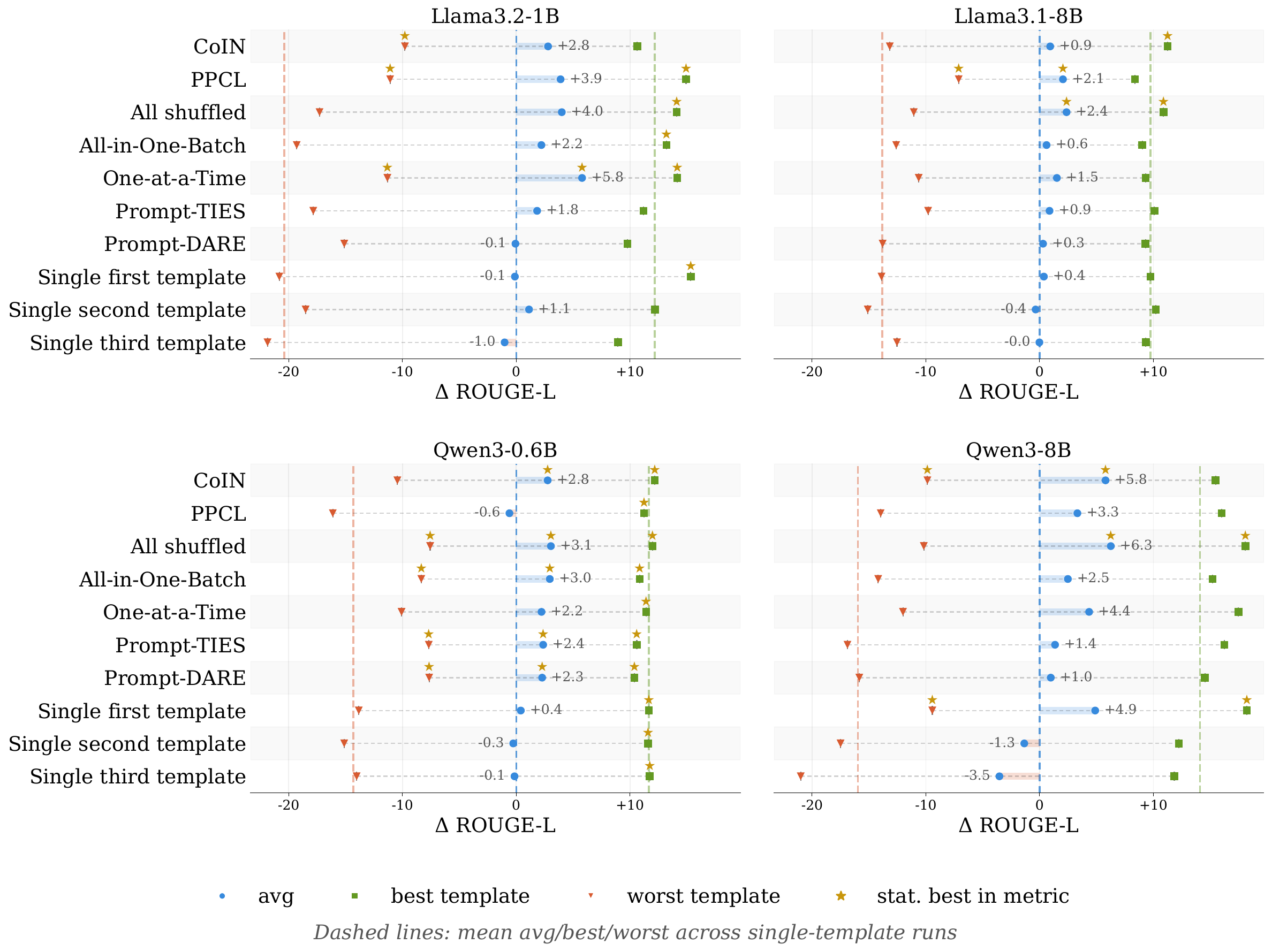}
    \caption{\textbf{Per-method Rouge-L relative to the average \textsc{Single} performance.} We report the scores on the \testall{} split. The circle, square, and triangle mark the average, best- and worst-template score respectively. The dashed span between the worst and best template reflects the method's global sensitivity to prompt variation. The vertical dashed lines mark the average performance of the \textsc{Single} runs. A star above a marker indicates membership in the statistically best group for that statistic.}
    \label{fig:grid_rougeL}
\end{figure*}

We evaluate \textsc{Prompt-TIES} and \textsc{Prompt-DARE} against the baseline methods in terms of performance and robustness, on the \testall{} split. \autoref{fig:frontier_test} plots each method on the plane defined by the average Rouge-L performance and \clv{} for each of the four base models. It defines the Pareto-optimal set. \autoref{fig:grid_rougeL} reports the global variance, as in previous work, through each method's mean, best-template, and worst-template Rouge-L performance \cite{cao2024worst}. All scores are expressed as the Rouge-L difference from the average performance of the \textsc{Single} methods, which abstracts individual model performance. The raw scores underlying both figures are listed in \autoref{tab:test_rougeL_all}.

\subsection{Scale sets the regime}
\label{sec:res_scale}

The two small models exhibit prompt-induced local variance several times higher than the large models across all methods. The best methods for the small models achieve a conditional local variance of $0.06$, whereas the best methods for the large models remain below $0.03$ (see \autoref{fig:frontier_test}). The spread of methods within a model is small compared to the difference in performance between the two model sizes. The same pattern holds for average performance, though gains between \QwenI{} and \QwenB{} are slim. The same applies to global variance, where the scaling effect is strongest within the Llama family. On average, the spread between the worst and best templates decreases from $0.287$ to $0.217$ (see \autoref{fig:grid_rougeL}). For the Qwen family, however, the effect is reversed. We discuss in \autoref{sec:qwen8B} why we hypothesize that \QwenB{} behaves differently. These findings frame every per-method comparison that follows. Method differences are real, but they operate within a regime largely fixed by model size.

\subsection{Gradient merging improves local robustness}
\label{sec:res_gradient}

\paragraph{Gradient merging on the fully fine-tuned models.}
On the two small, fully fine-tuned models, \textsc{Prompt-TIES} and \textsc{Prompt-DARE} roughly halve \clv{} relative to the baselines (see \autoref{fig:frontier_test}). On \LlamaI{}, this improvement comes at the expense of accuracy, while on \QwenI{} the gradient merging approaches are the most performant and robust.

This local robustness does not transfer to global robustness. Although our gradient merging approaches outperform the data construction basis \textsc{All-in-One-Batch} they are not the best models in terms of global robustness. Meaning, on a single example, gradient merging reduces variance. However, across the full set of templates, these changes average out (see \autoref{sec:globallocalresult} for an analysis of local and global variance). We hypothesize that gradient merging approaches target the example-template interaction, but cannot improve worst templates overall. This is evident in the interaction fraction, where across all models \textsc{Prompt-TIES} reduces the average $\rho^{\text{int}}$ to $50.5\%$, compared to $53.5\%$ for \textsc{All shuffled}. Additionally, HellaSwag is the only dataset on which our gradient merging approaches do not reach the lowest \clv{}. This dataset is dominated by global template effects (see \autoref{tab:interaction-fraction}).

\paragraph{Qualitative analysis of gradient merging approaches.}
We qualitatively examine our hypothesis that gradient merging approaches improve \clv{} by reducing template-example interaction. To do so, we manually review the datasets' predictions. From the SuperGLUE COPA dataset, we select six examples with the highest \clv{} in the \LlamaI{}-\textsc{All shuffled} predictions. We compare these predictions (see \autoref{tab:copa-full-ash}) to those of \LlamaI{}-\textsc{Prompt-TIES} (see \autoref{tab:copa-full-ties}).

The models always generate one of the two possible answers. Across all our manual inspections of the model outputs, this general instruction following, in the sense of predicting in the correct format, appears. Beyond that, \textsc{Prompt-TIES} produces more consistent outputs across these examples. \textsc{All shuffled} splits its predictions roughly evenly, as expected given its high local variance, and shows no clear pattern of which template is the best or the worst. \textsc{Prompt-TIES} predicts the same output for all templates in four of the six examples. This consistency can go in either direction, as seen in the last example where the model is wrong for every template.

Finally, \textsc{Prompt-TIES} does not necessarily improve the global variance. The best template score remains at 5 out of 6, though now a new template is the best template. The worst template score increases by one. Although, there are minor improvements in this subset, the examples are selected based on difficulty, so their effect is diluted when averaged with easier examples. This aligns with our findings across the datasets, in which the gradient merging approaches only marginally change global variance.

\paragraph{The LoRA models resist merging.}
For the two larger LoRA-trained models, the robustness advantage of gradient merging disappears. While these models remain among the most locally robust, they are outperformed by approaches with additional loss functions. This aligns with the key finding of KnOTS \cite{stoica2025knots}, that LoRA updates exist in different subspaces and cannot be seamlessly combined without an explicit subspace alignment step. Our coordinate-wise \textsc{Prompt-TIES} and \textsc{Prompt-DARE} operators do not perform this subspace alignment step. As shown in \autoref{sec:gradient_interference}, the interference persists, so an efficient KnOTS-style alignment step in gradient space could restore performance in the LoRA setting. This is the clearest avenue for future work on the method.

\paragraph{Marginal differences between \textsc{Prompt-TIES} and \textsc{Prompt-DARE}.}
Until now, we have only examined \textsc{Prompt-TIES} and \textsc{Prompt-DARE} together. When we separate them, we find that \textsc{Prompt-TIES} has a higher average performance, but there is no clear pattern indicating better robustness.

\subsection{Data construction achieves the highest accuracy}
\label{sec:res_data}

The data-construction strategies of \autoref{sec:data_construction}, which leave the loss objective untouched and vary only the within-batch schedule, have the strongest performance overall.

\paragraph{\textsc{All shuffled} is the strongest accuracy baseline.}
\textsc{All shuffled} is the clear performance winner. It achieves the highest accuracy in three of the four models and has the greatest mean gain among the data schedules on \LlamaB{} ($+2.0$), \QwenI{} ($+2.6$) and \QwenB{} ($+1.4$). Only on \LlamaI{} \textsc{One-at-a-Time} has a higher mean gain ($+5.9$ vs.\ $+4.1$). Additionally, it consistently has one of the best-performing templates. Therefore, the least-engineered, unconstrained use of the templates is a hard baseline for more elaborate methods to clear. While the method is effective at improving performance, it does not reduce variance. It has an average \clv{} across models and, only on \QwenI{}, one of the best worst-templates.

\paragraph{\textsc{One-at-a-Time} rarely leads.}
\textsc{One-at-a-Time} is the data construction approach used in prior work \cite{wei2025paft}. \textcite{wei2025paft} select this approach for their model training without evaluating it against other data-construction approaches. We demonstrate that it is the best data-construction approach only for \LlamaI{}, while on all other models, it performs similarly to or worse than \textsc{All shuffled} across all metrics. This approach may be useful in an online setting where prompts are generated on the fly, but it deprives the model of template variability within the batch.

\paragraph{\textsc{All-in-One-Batch} underperforms.}
Of the three multi-template data construction approaches, packing all formulations of an example into a single batch is the worst. \textsc{All-in-One-Batch} tends to produce the lowest accuracy and the highest prompt-induced variance, both locally and globally. Forcing every rendering of an example into a single gradient step sacrifices example diversity without gaining robustness. We hypothesize that this occurs due to gradient interference, as shown in \autoref{sec:gradient_interference}.

\paragraph{Single-template baselines underperform.}
The single-template baselines tend to have the lowest accuracy and the highest variance of all the methods. The one exception is \QwenB{}, where the second \textsc{Single} template achieves the highest \clv{}. We hypothesize that this is an artifact of \QwenB{} (see \autoref{sec:qwen8B}). More striking is the variability across the single-template choices themselves. The choice of template substantially shifts both accuracy and robustness, with a mean performance difference of $8.4$ between the first and third templates on \QwenB{}. The size of the difference and which template is best differ from model to model, with no template remaining consistent across models. Thus, single-template performance is hard to optimize, especially since the variance is dominated by concrete template-example interactions. We analyze the effect of template selection in more depth in \autoref{sec:res_template_count_and_choice}.

\paragraph{Summary of data construction approaches.}
Our analysis underscores the importance of multi-template training, which aligns with previous work \cite{wei2025paft, wei2022Flan}. Among multi-template data construction approaches \textsc{All shuffled} performs best. We hypothesize that this is due to a trade-off between two opposing effects of placing multiple templates in the same batch. In-batch template variance helps the model generalize across formulations, while gradient interference between templates of the same example hinders generalization. The two extreme schedules illustrate this trade-off: \textsc{One-at-a-Time} offers too little in-batch template variance, and \textsc{All-in-One-Batch} incurs too much gradient interference. \textsc{All shuffled} strikes a balance between the two. It mixes templates from different examples to increase in-batch variance while making it unlikely that the same example will appear under multiple templates within one batch. This finding aligns well with our gradient merging results. They improve \textsc{All-in-One-Batch} by removing gradient interference. If data construction approaches were the only option, we would advise against previous work and recommend the \textsc{All shuffled} approach.

\subsection{Auxiliary objectives rarely reach the frontier}
\label{sec:res_aux}

\paragraph{\textsc{PPCL} is unstable across models.}
\textsc{PPCL} is the least reliable method in the study. While it achieves the highest score on all metrics for \LlamaB{}, it performs poorly elsewhere. It has the highest local variance of any method and performs worse than the \textsc{Single} first template baseline on all metrics on both Qwen models. We analyze this brittleness in \autoref{sec:res_ppcl_diag} and trace it to a failure of cross-task transfer.

\paragraph{\textsc{CoIN} is steadier but usually dominated.}
\textsc{CoIN} is more stable than \textsc{PPCL} and is on the Pareto frontier for \QwenB{}. However, a simple data construction approach always matches or outperforms it on all metrics. It is difficult to justify the structured batch construction and extra loss objective if they do not significantly improve upon the standard loss function. In \autoref{sec:res_coin_diag} we analyze why the representational alignment that \textsc{CoIN} enforces does not translate into generation-time performance.

\subsection{Summary of the frontier}
\label{sec:res_frontier_summary}

The frontier is dominated by two method families. Data construction, especially the \textsc{All shuffled}, achieves the highest accuracy in three out of four models. \textsc{Prompt-TIES} and \textsc{Prompt-DARE} achieve the lowest \clv{} on fully fine-tuned models. Auxiliary-objective methods only reach the frontier in single, model-specific cases (\LlamaB{} for \textsc{PPCL}, \QwenB{} for \textsc{CoIN}) and are otherwise dominated by cheaper data construction methods. Three conclusions follow: 
\begin{enumerate}
\item A practitioner prioritizing accuracy with minimal effort should use \textsc{All shuffled}. To specifically suppress \clv{} when fully fine-tuning, gradient merging approaches should be chosen.
\item The more complex robustness objectives of \textsc{PPCL} and \textsc{CoIN} are difficult to justify. Their advantages are model-specific, and in the case of \textsc{PPCL}, they come at the cost of severe instability elsewhere.
\item The gap between the best and worst templates accounts for between $34\%$ on \LlamaB{} and $57\%$ on \LlamaI{} of the average performance. Thus, although the methods and model scaling improve global robustness, significant variation in performance remains across templates.
\end{enumerate}

\section{Robustness of the conclusions}
\label{sec:res_checks}

In \autoref{sec:res_frontier}, we measure the hard case using an out-of-distribution task, templates, and examples, and scoring with a generative metric.  In this section, we examine the robustness of our findings using the \valmaj{} split and rank classification accuracy.

Additionally, we analyze confounders that could influence our results. First, we examine how much of the results can be explained by the number and choice of templates. Second, we examine why \QwenB{} behaves differently from the other models.

\subsection{In-task generalization on the \valmaj{} split}
\label{sec:res_validation}

\begin{figure}[htbp]
  \centering
  \includegraphics[width=\linewidth]{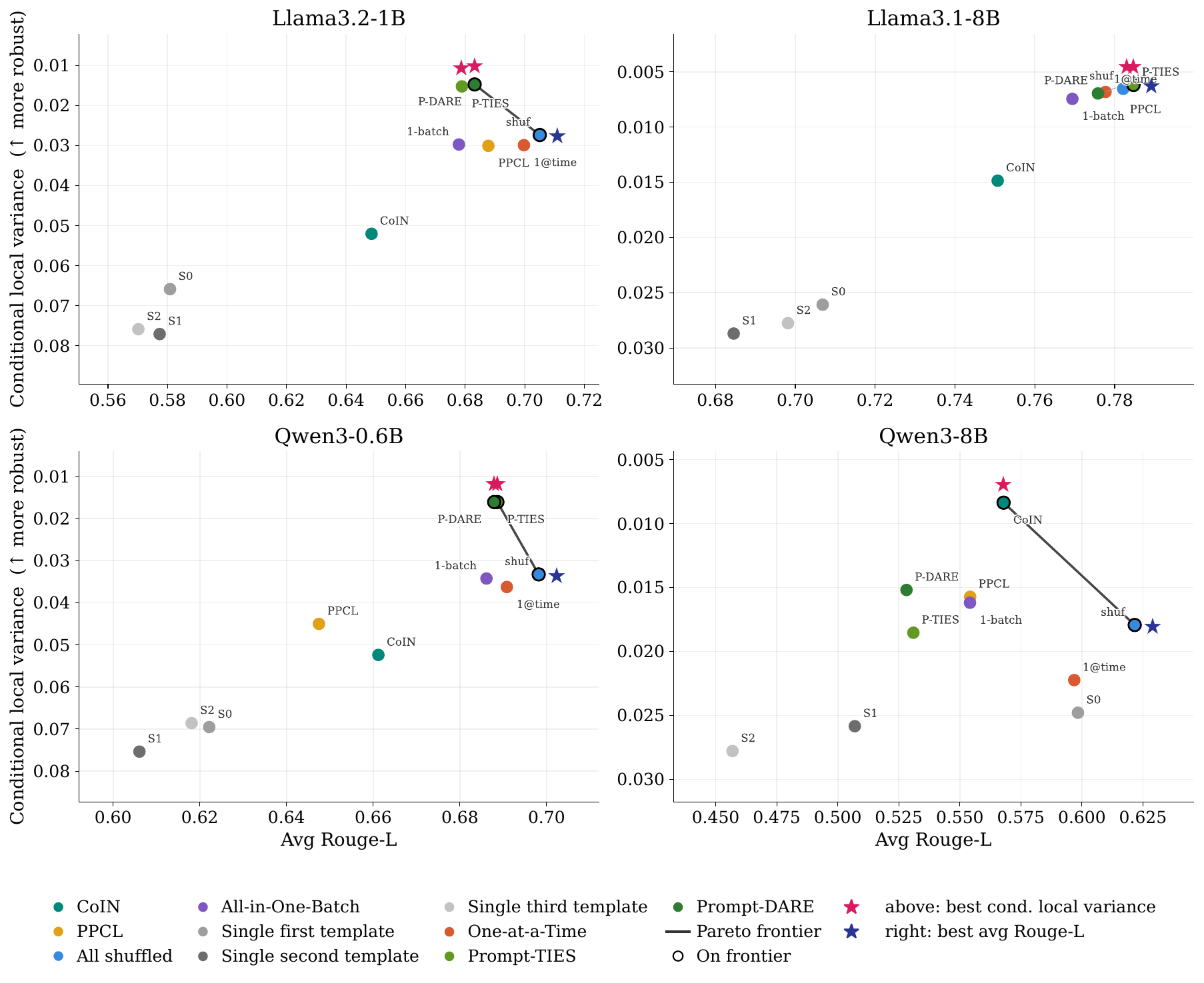}
  \caption{\textbf{Performance--robustness frontier on the \valmaj{} split.} Each point represents a training method and is positioned by the mean Rouge-L ($x$, $\rightarrow$ better) and \clv{} ($y$, $\uparrow$ better). The scores are from the \valmaj{} split. The Pareto-optimal set is outlined in black and connected by a solid line. The stars indicate the statistically best group per model: the \textcolor[HTML]{D81B60}{pink} star for \clv{} and the \textcolor[HTML]{283593}{indigo} star for average Rouge-L. \textsc{Prompt-TIES} and \textsc{Prompt-DARE} are the most robust on the fully-finetuned models.}
  \label{fig:frontier_val}
\end{figure}

The validation split reuses the training datasets, tasks, and templates, holding out only the examples (\autoref{sec:data}). Therefore, it isolates generalization to unseen inputs compared to the harder cross-task and cross-template transfer demanded by the test split. \autoref{fig:frontier_val} repeats the performance-robustness layout on the \valmaj{} split.

\paragraph{Absolute levels are not comparable across splits.}
We do not interpret a performance increase from the test to the validation based on the absolute Rouge-L values, because the two splits are from different tasks. For instance, the Rouge-L score is mechanically lower for a free-form summarization task in the validation set than for a coreference resolution task with short, closed-label answers. Directly comparing the mean Rouge-L score between the splits conflates generalization difficulty with output length. Therefore, we present the validation split results to check the relative ordering of methods.

\paragraph{The per-method ordering stays the same.}
The relative ordering from the test split is essentially preserved. \textsc{All shuffled} remains among the strongest in terms of accuracy. \textsc{Prompt-TIES} and \textsc{Prompt-DARE} again achieve the lowest prompt-induced variance on small models. \textsc{PPCL} and \textsc{CoIN} again fail to improve upon data construction approaches for most models. The robustness of the main findings shows that the results are stable across splits and are not just a phenomenon based on our test split.

\paragraph{What the validation view changes.}
Although the relative ordering remains the same, we observe changes in the validation split. As expected, the difference between the multi- and single-template methods widens because every multi-template method is trained on all the majority templates in a dataset. Therefore, evaluating on those same templates is nearly in-distribution. In contrast, a \textsc{Single} saw only one template per dataset, making the remaining majority templates out-of-distribution. Therefore, the single-multi gap is wider here than on the test set, where all methods face unseen templates, and the gap partly closes. These results confirm that the multi-template method's advantage largely stems from template coverage and that the held-out-task test split is a more demanding and informative measure of robustness.

Additionally, \textsc{PPCL} closes the performance gap to \textsc{CoIN}. On the \testall{} split, \textsc{PPCL} was markedly less stable than \textsc{CoIN}. On the \valmaj{} split, the performance of both is closer together, though they are still outperformed by data construction methods. We hypothesize that \textsc{PPCL}'s consistency penalty does not transfer to held-out tasks and templates, causing instability in the test set. We analyze this hypothesis in \autoref{sec:res_ppcl_diag}.

\subsection{Rank classification accuracy saturates}
\label{sec:res_rank}

The T0 benchmark was originally scored using rank classification accuracy \cite{sanh2022t0}. Although our main results are reported using Rouge-L,  we also report rank classification accuracy to ensure comparability.

\begin{figure*}[tp]
    \centering
    \includegraphics[width=\textwidth]{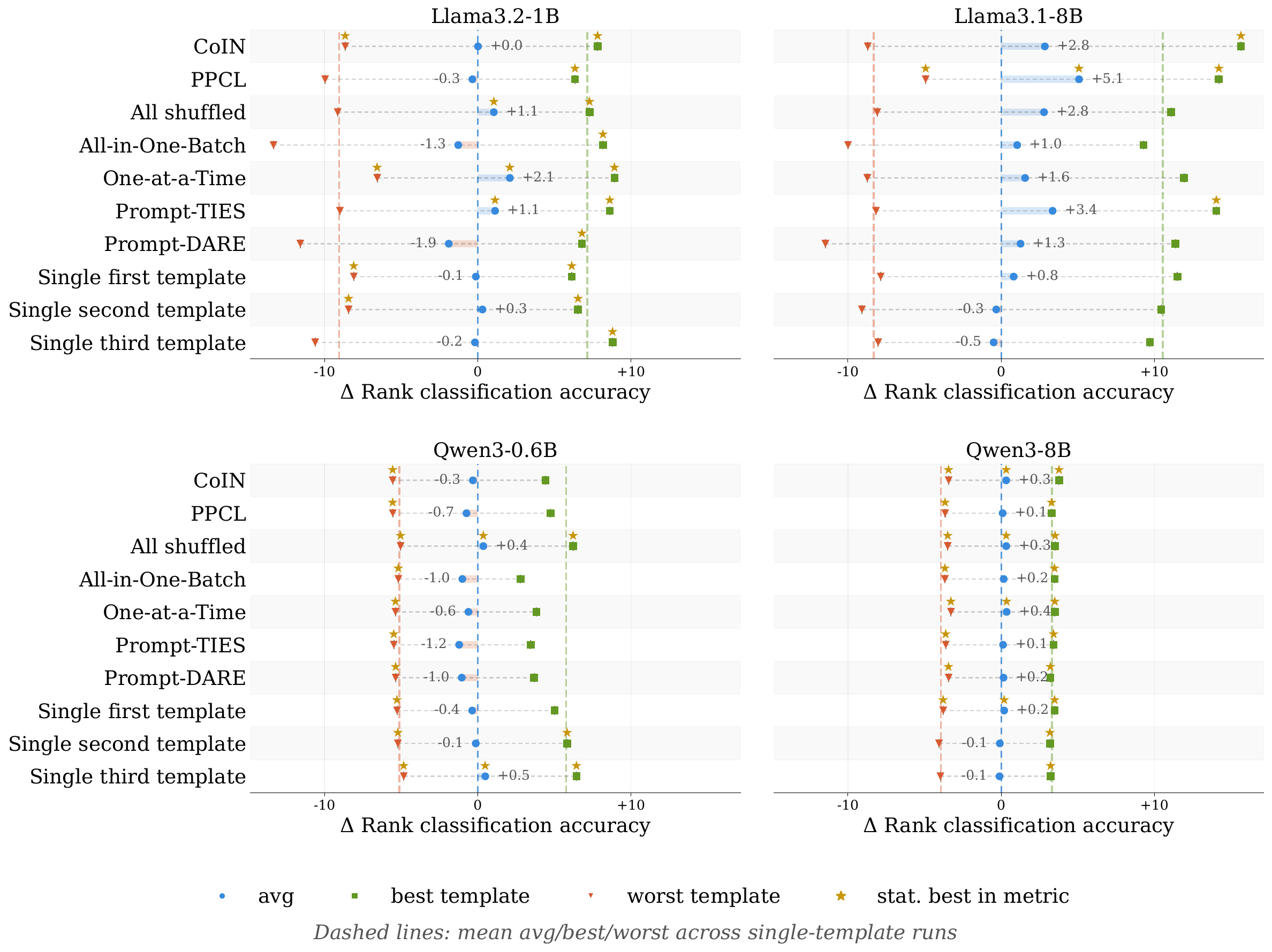}
    \caption{\textbf{Per-method rank classification accuracy relative to the average \textsc{Single} performance.} We report the scores on the \testall{} split. The circle, square, and triangle mark the average, best- and worst-template score respectively. The dashed span between the worst and best template reflects the method's global sensitivity to prompt variation. The vertical dashed lines mark the average performance of the \textsc{Single} runs. A star above a marker indicates membership in the statistically best group for that statistic. Almost every method on \QwenB{} falls within a fraction of a point of the baseline and into the statistically best group. This illustrates the ceiling effect that motivates our use of generative Rouge-L as the primary metric.}
    \label{fig:rank_acc}
\end{figure*}

\autoref{fig:rank_acc} shows that the difference between the worst and best templates decreases across all models. Additionally, the differences between the single template approaches shrink, especially for the Qwen models. For \QwenB{}, most methods are statistically the same for all metrics. Overall, these results demonstrate the anticipated failure mode in \autoref{sec:text_scoring_metrics}: rank classification provides the model with the answer set and only asks for a relative ordering. A capable model can maintain that ordering while adapting its generative behavior based on the prompt formulation.

\subsection{Template count versus template choice}
\label{sec:res_template_count_and_choice}

The validation split attributed much of the multi-template advantage to template coverage (see \autoref{sec:res_validation}). Thus, we investigate whether the performance gains are driven by the sheer number of training formulations or by the inclusion of important templates. To separate these two factors, we conduct a controlled sweep within the \textsc{All shuffled} family on the two fully fine-tuned models. For the count dimension, we train on the first one, two, and three templates per dataset, on the full majority set ($5.3$ templates per dataset, on average), and on all available templates ($6.3$ templates, on average).\footnote{Averages taken from  (\autoref{sec:data})}. For the choice dimension we only use one template, but vary which one we use. We compare the three fixed positions from our baselines, a per-dataset one-time random draw, and the empirically best and worst single template, which are selected post hoc from the full pool of templates based on the \textsc{All shuffled} run.

\begin{figure}[htbp]
  \centering
  \includegraphics[width=\linewidth]{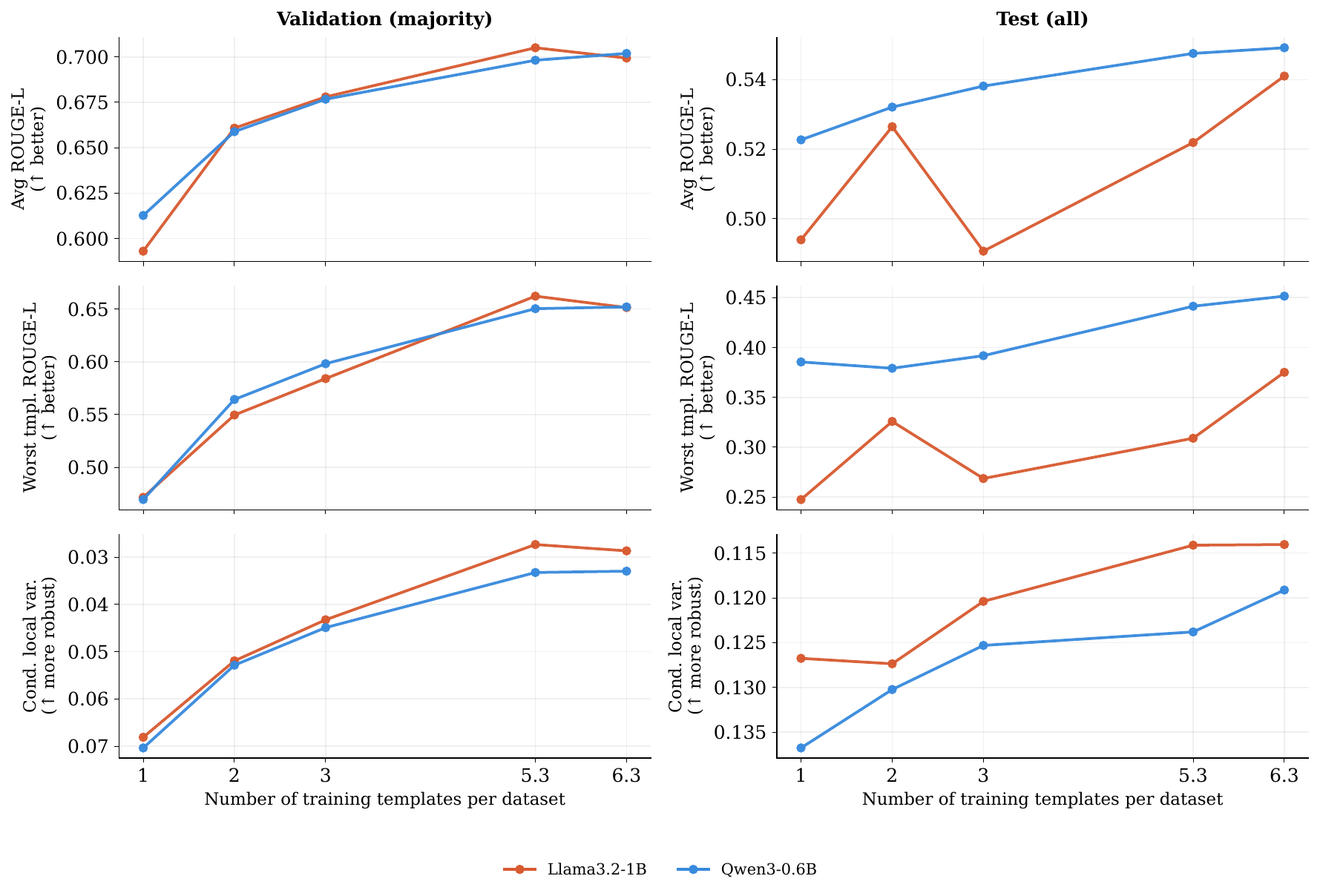}
  \caption{\textbf{Robustness scales with template count.} Average Rouge-L (top, $\uparrow$ better), worst-template Rouge-L (middle, $\uparrow$ better) and \clv{} (bottom, $\downarrow$ better) on the \testall{} split (left) and the \valmaj{} split (right), as a function of the number of training templates per dataset. One line per fully fine-tuned model. Training on more templates increases the performance and robustness.}
  \label{fig:template_count_test}
\end{figure}

\begin{figure}[htbp]
  \centering
  \includegraphics[width=\linewidth]{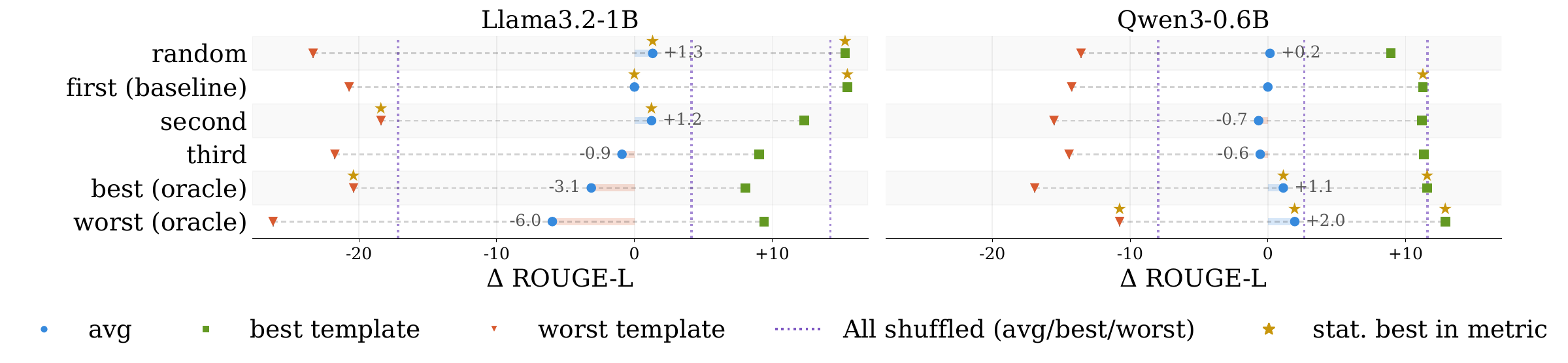}
  \caption{\textbf{At a fixed count template choice is unstable.} We report the scores on the \testall{} split. The circle, square, and triangle mark the average, best- and worst-template score respectively. The dashed span between the worst and best template reflects the method's global sensitivity to prompt variation. The vertical dashed lines mark the performance of the \textsc{All shuffled} run. A star above a marker indicates membership in the statistically best group for that statistic. For readability, we removed "single" and "template" from the method names. Template selection is unreliable and cannot be transferred between models.}
  \label{fig:template_choice}
\end{figure}

\paragraph{Robustness scales with template count.}
Adding templates increases the mean and worst-template performance while decreasing the \clv{} on the \testall{} split (\autoref{fig:template_count_test}). For \LlamaI{}, however, we see a higher performance with two template training than with three template training. This is likely because template number two improves performance drastically, while template number three lowers it for single-template training. This means the added benefit of one new template may decrease if the added template is much worse than the initial set (see \autoref{fig:template_choice}). 

This trend is clearer for the \valmaj{} split (see \autoref{fig:template_count_test}), where the saturating nature of the effect becomes apparent. The curve is monotonically increasing, but the performance gains achieved by adding templates decrease as more templates are added. Therefore, the marginal value of an extra formulation is front-loaded, as it is important for the model to see variance in the templates, but the variance becomes less and less effective.

\paragraph{Template choice does not generalize.}
To isolate the effect of the number of templates from the choice of templates, we fixed the number of templates to one and tested different template selection metrics (see \autoref{fig:template_choice}).

The oracle best and worst template runs chosen with full knowledge of the validation dataset scores do not provide stable gains across models. While they show moderate gains on \QwenI{}, they are detrimental to performance on \LlamaI{}. This is not just a test set artifact. The performance of the \LlamaI{} worst template run on the \valmaj{} split is $19.5$ points lower than the single first template run (see \autoref{tab:val_rougeL_template}). We hypothesize this is due to unstable training based on template selection. Additionally, template selection is not transferable across models, as the best single templates agree on only $\approx$30\% of datasets, and the worst agree on $\approx$43\%.

Random template selection is the most stable, hinting at the difficulty of selecting good training templates. Importantly, none of the methods can match the performance of \textsc{All shuffled}, as shown by the dashed lines in \autoref{fig:template_choice}. This means that template choice cannot make up for the lack of template count.\footnote{We acknowledge that we only tried automatic template selection. With more carefully selected, model- and dataset-specific templates, the scores could be improved.}

Overall, we found that adding more templates had a positive impact on all metrics, even though the marginal gains decreased. However, selecting better templates is a brittle and model-dependent process.

\subsection{Why does \QwenB{} behave differently?}
\label{sec:qwen8B}

\begin{figure*}[tp]
    \centering
    \includegraphics[width=\textwidth]{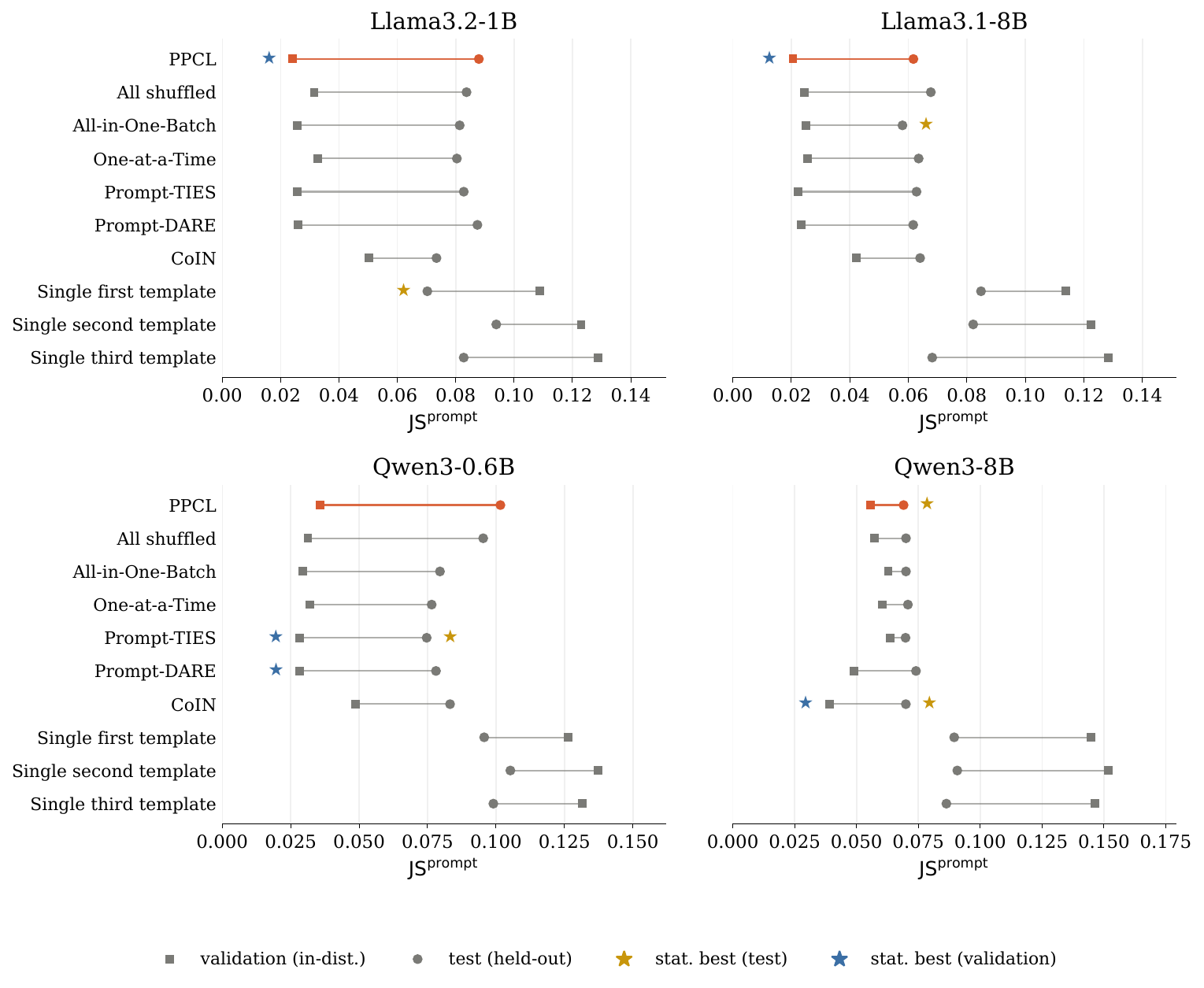}
    \caption{\textbf{\textsc{PPCL} only aligns the output distributions in validation.} The cross-template Jensen--Shannon divergence of the output distributions on the \texttt{prompt} readout (as defined in \autoref{sec:aux_metrics}). A lower Jensen--Shannon divergence is better. Squares show the score on the \valmaj{} split, while circles show the score on the \testmaj{} split. A star next to a marker indicates membership in the statistically best group for that readout. \textsc{PPCL} achieves for the Llama models substantially lower divergence in the validation set, but this advantage disappears in the held-out test set.}
    \label{fig:jsdiv_per_model}
\end{figure*}

\QwenB{} is often an outlier. For instance, the performance difference between model sizes in the Qwen model family is smaller than in the Llama family, and the single template methods have a wider range. The \textsc{Single} second template method even has the highest \clv{} (see \autoref{fig:frontier_test}). Additionally, the scores saturate strongly on the rank classification accuracy metric, with every non-single template run having the best worst- and best-template. Finally, \QwenB{} has a much smaller spread in Jensen--Shannon divergence and contrastive difference than the other models (see \autoref{fig:margin_per_model} and \autoref{fig:jsdiv_per_model}). We hypothesize that this is due to not training the output layer of the model. We applied LoRA to the output layer of \LlamaB{} as it would not adhere to the chat template otherwise. \QwenB{} already uses the correct chat template, even as a base model. Therefore, we used the standard TorchTune implementation. Training the output layer appears to be crucial for adapting base models to multi-template instruction fine-tuning; thus, the results for \QwenB{} should be viewed with caution.

\section{Why the auxiliary objectives fail}
\label{sec:res_diag}

As discussed in \autoref{sec:res_frontier}, \textsc{PPCL} and \textsc{CoIN} rarely reach the frontier and fail to improve upon simpler data construction approaches. This is unexpected since both methods have been shown to perform well in previous work \cite{qiang-etal-2024-ppcl, yan2024contrastive}. Therefore, we analyze why in our setting their performance is poor, separating the question into two parts. First, we test whether the training works as intended, i.e., whether each auxiliary objective effectively moves its target quantity. We examine this for \textsc{CoIN} in \autoref{sec:res_coin_diag} and for \textsc{PPCL} in \autoref{sec:res_ppcl_diag}. Second, we test whether the targeted quantity is a good objective for prompt robustness (\autoref{sec:res_diag_correlation}). There, we analyze the correlation between the diagnostic metrics and robustness, as well as the causal relationship between them. The raw scores for this section can be found in \autoref{tab:test_rougeL_template} and \autoref{tab:val_rougeL_template}.

\subsection{\textsc{CoIN} aligns label rather than prompt representations}
\label{sec:res_coin_diag}

The rationale behind \textsc{CoIN}'s contrastive objective follows from previous work showing that models tend to cluster their representations based on lexical similarity rather than semantics and that their robustness improves when they cluster based on semantics \cite{aissi-etal-2025-reinforcement}. \textsc{CoIN} therefore adds a contrastive objective that rewards the model for organizing its representations by semantic meaning. To test whether this objective shifts the target, we measure the \emph{contrastive difference} in evaluation: the difference in cosine similarity between the representations of semantically equivalent inputs and lexically similar inputs (see \autoref{sec:aux_metrics} for the exact definition).

\begin{figure*}[tp]
    \centering
    \includegraphics[width=\textwidth]{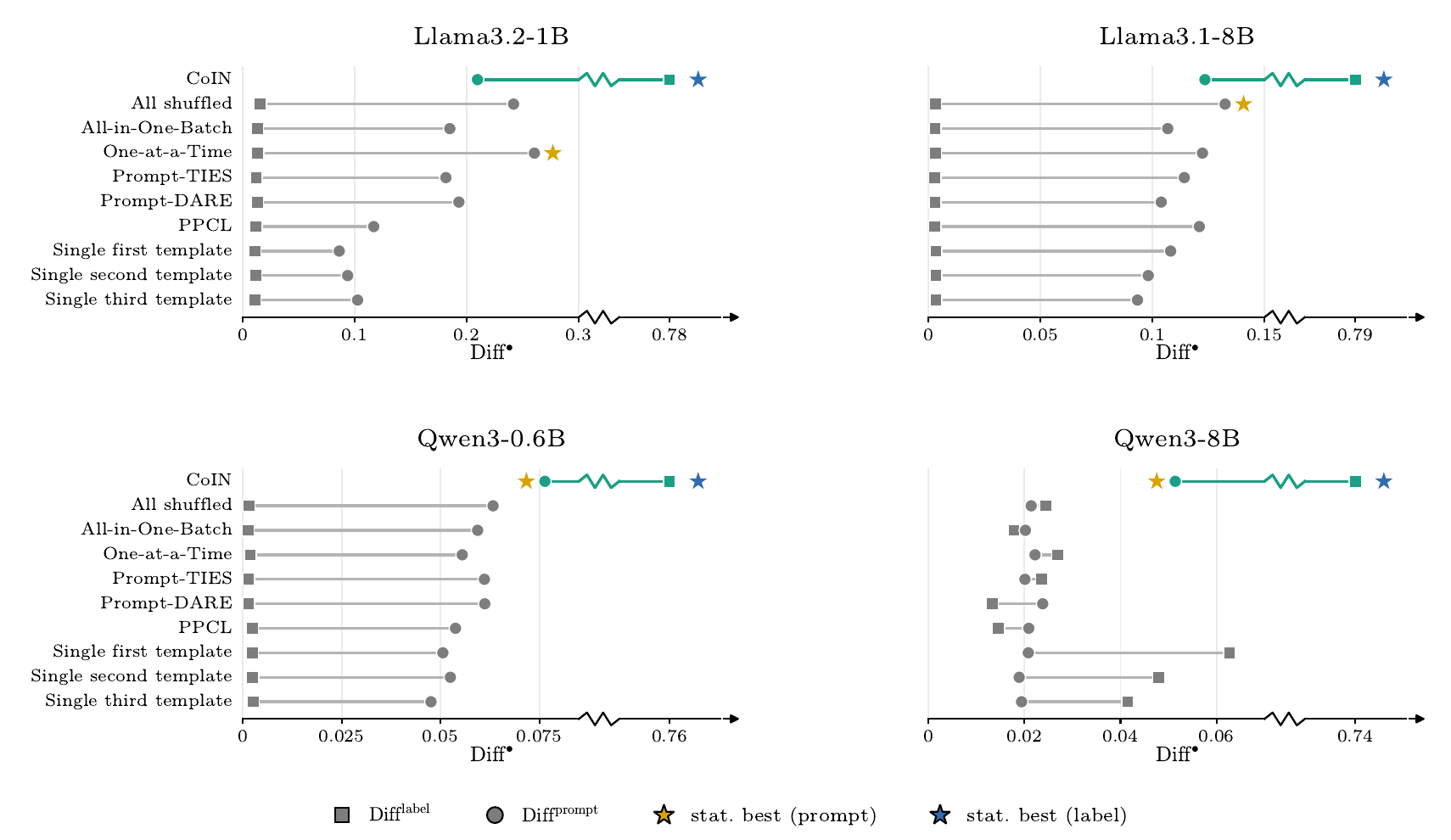}
    \caption{\textbf{\textsc{CoIN} only aligns the label readout.} The hidden-state difference (as defined in \autoref{sec:aux_metrics}) on the \testmaj{} split. A higher difference is better. Squares show the difference at the \texttt{label} readout, while circles show the difference at the \texttt{prompt} readout. A star next to a marker indicates membership in the statistically best group for that readout. \textsc{CoIN} attains a large gap at the \texttt{label} readout but not at the \texttt{prompt} readout, showing that its representational alignment does not carry over to the deployment-relevant generation mode.}
    \label{fig:margin_per_model}
\end{figure*}

\textsc{CoIN} achieves the largest contrastive difference of any method at the \texttt{label} readout (see \autoref{fig:margin_per_model}), meaning it clusters by semantic meaning rather than lexical surface form. Since the \texttt{label} readout is also the point at which the loss is applied, the objective shifts its target as intended. All other methods are near zero at the \texttt{label} readout, consistent with previous work \cite{aissi-etal-2025-reinforcement}. However, this gap shrinks to within roughly $0.01$ at the \texttt{prompt} readout on the Qwen models. On the Llama models \textsc{CoIN} separates even worse than competing data construction approaches. Interestingly, for every other method, the separation is stronger at the \texttt{prompt} readout, though it remains small. These findings demonstrate that \textsc{CoIN} can separate prompts when paired with their labels, but not when the prompt stands alone. We observe no large differences between the \testmaj{} and \valmaj{} splits.

\subsection{\textsc{PPCL} enforces consistency only in-distribution}
\label{sec:res_ppcl_diag}

The rationale behind \textsc{PPCL} is to align the output distribution across semantically equivalent prompt formulations \cite{qiang-etal-2024-ppcl}. To test if the target moves, we measure the cross-template Jensen--Shannon divergence of the output distributions, which is the quantity that \textsc{PPCL} penalizes directly (see \autoref{sec:aux_metrics}). Lower divergence indicates more consistent generation.

On the Llama models, \textsc{PPCL} achieves the lowest Jensen--Shannon divergence on the \valmaj{} set, but this advantage does not transfer to the \testmaj{} set. Consequently, \textsc{PPCL} has the largest span of Jensen--Shannon divergence between the \valmaj{} and \testmaj{} sets. While the objective enforces agreement across formulations on which it was trained, it does not produce a model that is consistent across unseen formulations and tasks. On the Qwen family, \textsc{PPCL} does not achieve the lowest Jensen--Shannon divergence, even on the \valmaj{} set. Yet it again shows the largest gap between \valmaj{} and \testmaj{} on \QwenI{}. We observe no significant difference between the \texttt{prompt} and \texttt{label} readouts. We attribute this to the labels being short, which causes both readouts to be dominated by the distribution of the first output token.

Therefore, the \textsc{PPCL} objective moves its target only within the distribution and does so less reliably than \textsc{CoIN} moves its own. We hypothesize that this occurs because the output distribution is a quantity that every model shapes by default and is thus a less distinctive lever than the hidden-state representation.

\subsection{Do the diagnostics predict robustness?}
\label{sec:res_diag_correlation}

We have established that \textsc{CoIN} and \textsc{PPCL} do move their respective targets, at least in distribution. Now, we turn to the second question: Are these quantities good objectives for achieving robustness? We test whether they correlate with higher performance and robustness to check if the reasons for choosing these objectives actually hold or if they are not effective levers for targeting robustness. To do so, we correlate each diagnostic with task performance across methods. We remove single methods from the correlation because they artificially increase correlation. There are three single method runs, which all have poor performance, high Jensen--Shannon divergence, and low contrastive difference. For each model, we compute the Spearman rank correlation, $\rho$, between the diagnostic metric and robustness or performance over the entire dataset. Note that this measures correlation, not causation. Below, we test whether this correlation can motivate new design decisions and demonstrate causation. \autoref{tab:corr} reports the mean $\rho$ over the four models for each diagnostic and readout.

\begin{table}[htbp]
\centering
\caption{Mean cross-method Spearman correlation $\bar\rho$ between each diagnostic and each performance statistic, averaged over the four models. The diagnostics are the hidden-state difference ($\mathrm{Diff}$) and the Jensen--Shannon divergence ($\mathrm{JS}$), both defined in \autoref{sec:aux_metrics}. The performance statistics are the Rouge-L accuracy (avg) and the local variance (\clv). We report each correlation per readout position on the \testmaj{} and \valmaj{} splits.}

\label{tab:corr}
\begin{tabular}{llrrrr}
\toprule
 &  & \multicolumn{2}{c}{test} & \multicolumn{2}{c}{val} \\
\cmidrule(lr){3-4}\cmidrule(lr){5-6}
 &  & \multicolumn{1}{c}{avg} & \multicolumn{1}{c}{clv} & \multicolumn{1}{c}{avg} & \multicolumn{1}{c}{clv} \\
metric & readout & $\overline{\rho}$ & $\overline{\rho}$ & $\overline{\rho}$ & $\overline{\rho}$ \\
\midrule
$\mathrm{Diff}$ & prompt & +0.38 & +0.06 & +0.26 & -0.23 \\
 & label & +0.08 & +0.60 & -0.15 & +0.49 \\
$\mathrm{JS}$ & prompt & +0.17 & +0.13 & -0.34 & +0.71 \\
 & label & +0.05 & -0.06 & -0.21 & +0.66 \\
\bottomrule
\end{tabular}
\end{table}

\paragraph{Correlation of contrastive difference and performance.}
The \texttt{prompt} readout for the contrastive difference is positively correlated with accuracy ($+0.38$ on the test set and $+0.26$ on the validation set) and negatively correlated with robustness on the validation set (see \autoref{tab:corr}). This means that organizing the hidden representation semantically rather than lexically corresponds with higher accuracy and robustness, as in \cite{aissi-etal-2025-reinforcement}. The \texttt{label} readout is not correlated with accuracy ($+0.08$ on the test set, $-0.15$ on the validation set) and decisively, the correlation to \clv{} becomes positive, meaning it is correlated with high variance. This is consistent with \textsc{CoIN}'s own behavior. \textsc{CoIN} can contrast prompts in the \texttt{label} readout, but it cannot align them in the \texttt{prompt} readout, resulting in poor performance. Overall, we find that the contrastive difference at the \texttt{label} readout is a misleading proxy.

\paragraph{Correlation of Jensen--Shannon divergence and performance.}
In validation, lower cross-template Jensen--Shannon divergence correlates with lower \clv{}, while the correlation with performance is much weaker, though still present. On the test split there is little to no correlation. As analyzed in \autoref{sec:res_ppcl_diag}, we observe no meaningful difference in correlation between the readout points. These findings demonstrate that Jensen--Shannon divergence helps align model outputs with training templates, thereby decreasing variability in answers and \clv{}. However, this benefit does not extend to new templates and tasks.

\paragraph{Retargeting the contrast to the prompt readout.}
We test whether switching to the \texttt{prompt} readout, which correlates with higher performance and lower variance, actually results in a better model (see \autoref{tab:corr}). To test this hypothesis, we use $\text{CoIN}_\text{prompt}$, which is identical to $\text{CoIN}_\text{label}$ except that the contrastive loss pools the hidden state at the last prompt token rather than the last target token (see \autoref{sec:coin}). Due to computational restrictions, we train $\text{CoIN}_\text{prompt}$ on the two small, fully fine-tuned models. $\text{CoIN}_\text{prompt}$ significantly increases the contrastive difference at the \texttt{prompt} readout and achieves the highest score of all models ($0.61$ on both models, up from $0.21$ and $0.08$ from $\text{CoIN}_\text{label}$). Its label-readout difference collapses to nearly zero. This mirrors the behavior of $\text{CoIN}_\text{label}$, with each variant winning the readout it is trained on.

However, neither the robustness nor the accuracy improves. Rouge-L does not change on \LlamaI{} ($-0.4$ pp) and drops sharply on \QwenI{} ($-4.4$ pp). The \clv{} is unchanged or slightly worse, and on both models $\text{CoIN}_\text{prompt}$ falls short of the \textsc{All shuffled}. Aligning the prompt representation is thus correlated with robustness across methods, but it does not cause it. This means that \textsc{CoIN}'s contrastive objective does not become competitive merely by changing the token on which the loss is conditioned. Our proposed gradient merging methods outperform all baselines on \clv{} on fully-finetuned models.

\chapter{Conclusion}
The goal of this thesis is to improve prompt robustness by reducing the gradient interference that arises in multi-prompt training. Previous studies have attempted to improve prompt robustness at train-time by either adding additional phrasings to the training data or by adding auxiliary loss objectives. However, we found that the former introduces gradient interference and that the latter is brittle.

First, we define how prompt robustness should be measured. In our experiments, brittleness does not stem from a few uniformly weak templates, but rather from the interaction between a template and the specific example to which it is applied. Therefore, we treat local variance --- how much a single example's answer fluctuates across phrasings --- as the robustness target that best reflects the user experience rather than the performance of the worst template alone. Global metrics average out much of this variance. $54\%$ of local variance stems from template-example interaction and is lost in the global measure.

We characterize gradient interference in multi-template training, a common approach to improving prompt robustness. Across trainable parameters, $57\%$ have at least two per-template gradients pushing in opposite directions. The naive average discards approximately one-fifth of the gradient magnitude through cancellation. Compared to multi-task training, the cosine similarity between gradients is comparatively high, so the interference is dominated by per-parameter sign conflict and concentrated in the early layers of the network.

Based on these findings, we adopt sign-based interference-resolution operators from model merging for gradient merging. We introduce \textsc{Prompt-TIES} and \textsc{Prompt-DARE}, as the per-template gradient merging variants of the model-merging methods \textsc{TIES} and \textsc{DARE}. Both methods first sparsify the gradients, then elect a majority sign per parameter, and finally restrict the merge to parameters that agree with that sign.

We conduct a controlled comparison of these methods against train-time prompt-robustness approaches spanning all three established classes: data construction, consistency regularization (\textsc{PPCL}), and contrastive alignment (\textsc{CoIN}). Data construction approaches, especially \textsc{All shuffled}, improve performance but not the \clv{}. The performance of \textsc{PPCL} and \textsc{CoIN} varies strongly from model to model. Our gradient-merging models roughly halve the \clv{} at little cost to accuracy on fully fine-tuned models by directly reducing the template-example interaction that causes brittleness. However, on the LoRA-trained 8B models this advantage disappears, consistent with the subspace mismatch of LoRA updates reported by KnOTS \cite{stoica2025knots}.

We further diagnose why the purpose-built auxiliary objectives of \textsc{PPCL} and \textsc{CoIN} rarely improve robustness. Both objectives improve the targeted metric during training but do not generalize beyond it. Additionally, while the targets of the interventions, particularly the hidden-representation alignment of \textsc{CoIN}, correlate with improved performance and robustness, training on these metrics does not causally enhance either.

For practitioners, this yields a simple rule of thumb. If the goal is raw performance, data construction is sufficient. However, if local per-example robustness is important and the model is fully fine-tuned, adding \textsc{Prompt-TIES} or \textsc{Prompt-DARE} significantly improves robustness. Overall, performance and robustness increase with scale. However, prompt brittleness does not disappear. Across all methods, the gap between a model's best and worst performing phrasings remains large: between $34$ and $57\%$ of its performance. This leaves ample opportunity for further research into training-time prompt robustness, and we hope this thesis will encourage continued progress.

\chapter{Limitations}
Although our work demonstrates the potential to enhance the robustness of prompts for large language models through gradient merging, it is not without significant limitations.

\paragraph{Instruction fine-tuning only.}
Modern large language training involves multiple stages. Reinforcement learning after supervised fine-tuning could be crucial in enhancing the model's prompt robustness. Additionally, we observe approaches that address prompt robustness \cite{aissi-etal-2025-reinforcement} and gradient merging \cite{lu2024online} during this stage of model training. However, since we fine-tune using data with relatively short labels, reinforcement learning on our checkpoints is difficult. Future work should investigate the effects of prompt robustness approaches throughout the entire training process.

\paragraph{Model scale and family coverage.}
Our experiments are limited to models with up to 8 billion parameters across two families of models (Llama and Qwen). Whether the observed patterns and benefits of gradient merging approaches hold at larger scales, where models may be capable of leveraging richer training signals, remains an open question. Evaluating additional architectures would strengthen the generalizability of our conclusions.

\paragraph{Benchmark scope and evaluation protocol.}
We adopt the training and evaluation split from \textcite{sanh2022t0}, which is well-established in the field of prompt robustness. However, it consists predominantly of classification and short-answer tasks with closed label sets. Prompt sensitivity may manifest differently in such settings than in open-ended generation. Evaluating on more diverse benchmarks would provide a more complete picture of prompt robustness. These datasets would also enable us to further analyze downstream training methods.

\paragraph{Method selection within each category.}
We evaluate one representative method from each of the consistency regularization and contrastive alignment categories. We selected \textsc{PPCL} and \textsc{CoIN}, respectively, based on their central standing in these categories. Alternative methods in these categories \cite{sun2024evaluating, hejabi2025flipflopconsistencyunsupervisedtraining, Liu2025ImprovingCL, aissi-etal-2025-reinforcement} may have different capabilities and trade-offs. Our conclusions apply to the specific instantiations studied, not the broader methodological families.

\paragraph{Hyperparameter optimization.}
We performed a grid search over learning rates for all baselines and tuned the \textsc{PPCL} regularization weight on one model. However, a more exhaustive search, particularly for the \textsc{PPCL} and \textsc{CoIN} auxiliary loss weights, could yield stronger baselines and affect the reported relative comparisons. However, the computational cost of such a search across all models and methods exceeded our resource budget. Additionally, we did not tune the learning rate or trim factor for the gradient merging approaches. Our trim factor was chosen conservatively at $20\%$. \textsc{DARE} shows that dropping up to $95\%$ of the task vectors only incurs small performance drops. With a higher trim factor, we can achieve much more efficient gradient merging because materializing the $T_i \times N$ gradient matrix will not be necessary with strongly trimmed gradients.

\paragraph{Prompt templates.}
Our work is limited by our use of prompt templates as the sole source of variance. Although \textcite{chatterjee-etal-2024-posix} show that models are most sensitive to paraphrastic and structural variations enforced through the use of prompt templates, there are many other types of variance that are not covered. For example, we do not account for human error, such as typos, which are seen in the real world and increase variance, as discussed in \textcite{zhu2024promptbench}.

\chapter{Future work}
We hope that our in-depth analysis of gradient merging for prompt-agnostic training will pave the way for future research on prompt-agnostic language models. We envision several avenues for future research:
\begin{enumerate}
    \item We demonstrate that our gradient merging approaches do not work well on LoRA-trained models, which is consistent with the findings of KnoTS \cite{stoica2025knots}. We found the parameter-wise single vector composition of KnoTS to be too costly for merging gradients at each time step. Thus, we encourage future researchers to develop efficient solutions to address the subspace problem in LoRA.
    \item An open question is how many gradients can be removed while still improving prompt robustness. \textsc{DARE} swept their trim factor to show that $90\%$ to $95\%$ of the task vectors can be dropped while retaining performance. Similar experiments would be interesting for prompt-induced gradients. We hypothesize that these too will be very redundant.
    \item We invite future researchers to port \textsc{Prompt-TIES} and \textsc{Prompt-DARE} into reinforcement learning and test whether the added robustness benefits from supervised fine-tuning transfer through reinforcement learning.
\end{enumerate}

\makeatletter
    \def\toclevel@chapter{-1}
    \def\toclevel@section{0}
\makeatother

\pagestyle{plain}

\renewcommand*{\bibfont}{\small}
\printbibheading
\addcontentsline{toc}{chapter}{Bibliography}
\printbibliography[heading = none]

\addchap{Appendix}
\renewcommand{\thesection}{\Alph{section}}
\setcounter{section}{0}
\counterwithin{table}{section}
\counterwithin{figure}{section}

\section{AI usage}
During the planning stage of this thesis I used ChatGPT and Claude to refine search terms for the literature review. While preparing the code for the project I used GitHub Copilot and Claude Code, GitHub Copilot to review my hand-written code and Claude to help create the plotting scripts. For the final draft I used DeepL Write and Claude for language editing, spell checking, and streamlining. No confidential or personal data was entered into any of these tools. I verified all factual claims, citations, and quotations against the original sources, and the research design, analysis, and argument are my own work.

\section{Dataset statistics}
\label{app:dataset}

\begin{table*}[ht]
  \centering
  \footnotesize
  \caption{
    The 48 training datasets grouped by task type. \emph{Templates} is the total number of PromptSource templates per dataset. \emph{Filtered} is the count after removing manually identified bad templates, and \emph{Majority} is the size of the majority templates (see \autoref{sec:data}). The number of samples is based on the training split, limited to 10,240. We use the evaluation splits of the datasets for validation, with a sample limit of 512 and the same templates.
  }\label{tab:train_datasets}
  \begin{tabular}{@{}llrrrr@{}}
    \toprule
    \textbf{Task} & \textbf{Dataset} & \textbf{Samples} & \textbf{Templates} & \textbf{Filtered} & \textbf{Majority}\\
    \midrule
  Paraphrase        & GLUE (MRPC)            &  3,668 &  7 &  5 &  4 \\
                    & GLUE (QQP)             & 10,240 &  6 &  6 &  4 \\
                    & PAWS                   & 10,240 & 12 & 10 & 10 \\
  \midrule
  Extractive QA     & HotpotQA               & 10,240 &  5 &  5 &  5 \\
                    & TriviaQA               & 10,240 &  5 &  4 &  1 \\
                    & WebQuestions           &  3,778 &  5 &  5 &  1 \\
                    & WikiQA                 & 10,240 & 11 &  5 &  4 \\
                    & AdvQA (BiDAF)          & 10,000 &  5 &  4 &  4 \\
                    & AdvQA (BERT)           & 10,000 &  5 &  4 &  4 \\
                    & AdvQA (RoBERTa)        & 10,000 &  5 &  4 &  4 \\
                    & DuoRC (Self)           & 10,240 &  9 &  5 &  1 \\
                    & DuoRC (Paraphrase)     & 10,240 &  9 &  5 &  1 \\
                    & ROPES                  & 10,240 & 12 & 12 & 12 \\
                    & SQuAD v2               & 10,240 & 12 &  4 &  2 \\
                    & SuperGLUE (ReCoRD)     & 10,240 & 20 &  7 &  1 \\
                    & Quoref                 & 10,240 & 11 & 10 &  1 \\
                    & CoQA                   & 10,240 &  7 &  7 &  6 \\
  \midrule
  MC QA             & ARC-Challenge          &  1,119 &  6 &  5 &  3 \\
                    & ARC-Easy               &  2,251 &  6 &  5 &  3 \\
                    & CoS-E                  &  9,741 & 11 &  6 &  3 \\
                    & CosmosQA               & 10,240 & 13 &  8 &  4 \\
                    & DREAM                  &  6,116 &  5 &  4 &  2 \\
                    & OpenBookQA             &  4,957 &  7 &  7 &  6 \\
                    & PIQA                   & 10,240 & 11 &  6 &  3 \\
                    & QASC                   &  8,134 &  8 &  6 &  6 \\
                    & QuAIL                  & 10,240 & 13 & 11 &  6 \\
                    & QuaRel                 &  1,941 &  5 &  5 &  5 \\
                    & QuaRTz                 &  2,696 &  8 &  8 &  8 \\
                    & RACE (High)            & 10,240 &  8 &  5 &  3 \\
                    & RACE (Middle)          & 10,240 &  8 &  5 &  3 \\
                    & SciQ                   & 10,240 &  5 &  5 &  5 \\
                    & SuperGLUE (BoolQ)      &  9,427 & 10 &  5 &  3 \\
                    & SuperGLUE (MultiRC)    & 10,240 & 10 & 10 & 10 \\
                    & WIQA                   & 10,240 &  8 &  4 &  2 \\
  \midrule
  Sentiment         & Amazon Polarity        & 10,240 &  9 &  9 &  3 \\
                    & App Reviews            & 10,240 &  4 &  4 &  1 \\
                    & IMDB                   & 10,240 & 11 & 11 &  7 \\
                    & Rotten Tomatoes        &  8,530 & 10 & 10 &  7 \\
                    & Yelp Reviews           & 10,240 &  7 &  7 &  5 \\
  \midrule
  Summarization     & CommonGen              & 10,240 &  9 &  7 &  7 \\
                    & CNN/DailyMail          & 10,240 &  9 &  7 &  7 \\
                    & SciTLDR                &  1,992 &  6 &  5 &  5 \\
                    & MultiNews              & 10,236 &  6 &  5 &  5 \\
                    & SAMSum                 & 10,240 &  7 &  6 &  5 \\
                    & XSum                   & 10,240 & 10 & 10 & 10 \\
  \midrule
  Topic Classification& AG News                & 10,240 &  7 &  7 &  4 \\
                    & DBpedia                & 10,240 &  4 &  4 &  4 \\
                    & TREC                   &  5,452 & 18 &  5 &  5 \\
  \midrule
  \multicolumn{2}{@{}l}{\textbf{Total (48 datasets)}} & \textbf{417,238} & & & \\
    \bottomrule
  \end{tabular}
\end{table*}

\begin{table*}[ht]
  \centering
  \small
  \caption{
      The 11 held-out test datasets grouped by task type. \emph{Templates} is the total number of PromptSource templates per dataset. \emph{Filtered} is the count after removing manually identified bad templates, and \emph{Majority} is the size of the majority templates (see \autoref{sec:data}).
  }\label{tab:test_datasets}
  \begin{tabular}{@{}llrrrr@{}}
    \toprule
    \textbf{Task} & \textbf{Dataset} & \textbf{Samples} & \textbf{Templates} & \textbf{Filtered} & \textbf{Majority}\\
    \midrule
  NLI               & SuperGLUE (CB)         &     56 & 15 & 15 &  8 \\
                    & SuperGLUE (RTE)        &    277 & 10 &  9 &  9 \\
                    & ANLI (R1)              &  1,000 & 15 & 15 &  8 \\
                    & ANLI (R2)              &  1,000 & 15 & 15 &  8 \\
                    & ANLI (R3)              &  1,200 & 15 & 15 &  8 \\
  \midrule
  WSD               & SuperGLUE (WiC)        &    638 & 10 &  9 &  8 \\
  \midrule
  Coreference       & SuperGLUE (WSC)        &    104 & 10 & 10 &  7 \\
  \midrule
  Commonsense       & WinoGrande             &  1,767 &  6 &  6 &  5 \\
                    & SuperGLUE (COPA)       &    100 & 12 &  7 &  7 \\
                    & StoryCloze             &  1,871 &  6 &  5 &  5 \\
                    & HellaSwag              & 10,003 & 11 &  5 &  3 \\
  \midrule
  \multicolumn{2}{@{}l}{\textbf{Total (11 datasets)}} & \textbf{18,016} & & & \\
    \bottomrule
  \end{tabular}
\end{table*}

\clearpage
\section{Additional results}
\label{app:results}

In this section we report the raw scores underlying the plots in the paper. We report for the main experiments: the \testall{} Rouge-L scores in \autoref{tab:test_rougeL_all}; the \testmaj{} Rouge-L scores in \autoref{tab:test_rougeL_majority}; the \testall{} rank classification accuracy scores in \autoref{tab:test_rank_all}; and the \valmaj{} Rouge-L scores in \autoref{tab:val_rougeL_majority}.

For the template-count and template-choice experiments we report: the \testall{} Rouge-L scores in \autoref{tab:test_rougeL_template} and the \valmaj{} Rouge-L scores in \autoref{tab:val_rougeL_template}

\begin{table}[htbp]
  \centering
  \caption{ROUGE-L scores on the \testall{} split. \textbf{Bold} marks values in the statistically best group per column.}
  \label{tab:test_rougeL_all}
  \small
  \begin{tabular}{lcccc}
    \toprule
    Method & worst  & avg & best  & clv  \\
    \midrule
    \multicolumn{5}{l}{\textbf{\LlamaI{}}} \\
    \midrule
    \textsc{Single} first template & 0.2736 & 0.4807 & \textbf{0.6353} & 0.1490 \\
    \textsc{Single} second template & 0.2967 & 0.4931 & 0.6039 & 0.1329 \\
    \textsc{Single} third template & 0.2633 & 0.4716 & 0.5713 & 0.1033 \\
    \textsc{All shuffled} & 0.3090 & 0.5219 & \textbf{0.6229} & 0.1141 \\
    \textsc{One-at-a-Time} & \textbf{0.3686} & \textbf{0.5397} & \textbf{0.6234} & 0.1145 \\
    \textsc{All-in-One-Batch} & 0.2889 & 0.5040 & \textbf{0.6139} & 0.1194 \\
    \textsc{PPCL} & \textbf{0.3710} & 0.5209 & \textbf{0.6311} & 0.1244 \\
    \textsc{CoIN} & \textbf{0.3840} & 0.5099 & 0.5883 & 0.1187 \\
    \textsc{Prompt-TIES} & 0.3035 & 0.5002 & 0.5938 & 0.0668 \\
    \textsc{Prompt-DARE} & 0.3307 & 0.4812 & 0.5796 & \textbf{0.0626} \\
    \midrule
    \multicolumn{5}{l}{\textbf{\LlamaB{}}} \\
    \midrule
    \textsc{Single} first template & 0.4949 & 0.6377 & 0.7313 & 0.0363 \\
    \textsc{Single} second template & 0.4828 & 0.6304 & 0.7360 & 0.0383 \\
    \textsc{Single} third template & 0.5085 & 0.6337 & 0.7273 & 0.0342 \\
    \textsc{All shuffled} & 0.5233 & \textbf{0.6576} & \textbf{0.7429} & 0.0311 \\
    \textsc{One-at-a-Time} & 0.5276 & 0.6491 & 0.7272 & 0.0293 \\
    \textsc{All-in-One-Batch} & 0.5078 & 0.6400 & 0.7242 & 0.0340 \\
    \textsc{PPCL} & \textbf{0.5628} & \textbf{0.6545} & 0.7178 & \textbf{0.0224} \\
    \textsc{CoIN} & 0.5022 & 0.6432 & \textbf{0.7464} & 0.0345 \\
    \textsc{Prompt-TIES} & 0.5359 & 0.6426 & 0.7349 & 0.0261 \\
    \textsc{Prompt-DARE} & 0.4959 & 0.6369 & 0.7270 & 0.0292 \\
    \midrule
    \multicolumn{5}{l}{\textbf{\QwenI{}}} \\
    \midrule
    \textsc{Single} first template & 0.3787 & 0.5211 & \textbf{0.6337} & 0.1333 \\
    \textsc{Single} second template & 0.3659 & 0.5144 & \textbf{0.6330} & 0.1380 \\
    \textsc{Single} third template & 0.3768 & 0.5156 & \textbf{0.6345} & 0.1352 \\
    \textsc{All shuffled} & \textbf{0.4414} & \textbf{0.5476} & \textbf{0.6369} & 0.1238 \\
    \textsc{One-at-a-Time} & 0.4163 & 0.5393 & \textbf{0.6313} & 0.1238 \\
    \textsc{All-in-One-Batch} & \textbf{0.4336} & \textbf{0.5466} & \textbf{0.6256} & 0.1112 \\
    \textsc{PPCL} & 0.3558 & 0.5111 & \textbf{0.6294} & 0.1425 \\
    \textsc{CoIN} & 0.4126 & \textbf{0.5447} & \textbf{0.6389} & 0.1300 \\
    \textsc{Prompt-TIES} & \textbf{0.4402} & \textbf{0.5408} & \textbf{0.6230} & \textbf{0.0559} \\
    \textsc{Prompt-DARE} & \textbf{0.4405} & \textbf{0.5399} & \textbf{0.6210} & \textbf{0.0562} \\
    \midrule
    \multicolumn{5}{l}{\textbf{\QwenB{}}} \\
    \midrule
    \textsc{Single} first template & \textbf{0.4417} & 0.5849 & \textbf{0.7182} & 0.0331 \\
    \textsc{Single} second template & 0.3611 & 0.5226 & 0.6585 & \textbf{0.0294} \\
    \textsc{Single} third template & 0.3261 & 0.5007 & 0.6544 & 0.0348 \\
    \textsc{All shuffled} & 0.4342 & \textbf{0.5987} & \textbf{0.7170} & 0.0334 \\
    \textsc{One-at-a-Time} & 0.4159 & 0.5796 & 0.7109 & 0.0334 \\
    \textsc{All-in-One-Batch} & 0.3942 & 0.5609 & 0.6882 & 0.0336 \\
    \textsc{PPCL} & 0.3964 & 0.5693 & 0.6962 & 0.0355 \\
    \textsc{CoIN} & \textbf{0.4374} & \textbf{0.5940} & 0.6908 & \textbf{0.0300} \\
    \textsc{Prompt-TIES} & 0.3672 & 0.5496 & 0.6984 & 0.0343 \\
    \textsc{Prompt-DARE} & 0.3775 & 0.5459 & 0.6813 & 0.0332 \\
    \bottomrule
  \end{tabular}
\end{table}

\begin{table*}[htbp]
  \centering
  \caption{ROUGE-L scores on the \testmaj{} split. \textbf{Bold} marks values in the statistically best group per column.}
  \label{tab:test_rougeL_majority}
  \footnotesize
  \begin{tabular}{lcccccccc}
    \toprule
    Method & worst  & avg & best  & clv & $\mathrm{JS}^{\text{prompt}}$ & $\mathrm{JS}^{\text{label}}$ & $\mathrm{Diff}^{\text{prompt}}$ & $\mathrm{Diff}^{\text{label}}$ \\
    \midrule
    \multicolumn{9}{l}{\textbf{\LlamaI{}}} \\
    \midrule
    \textsc{Single} first template & 0.3962 & 0.4553 & 0.5025 & 0.0693 & \textbf{0.0703} & 0.0325 & 0.0861 & 0.0107 \\
    \textsc{Single} second template & 0.3789 & 0.4940 & \textbf{0.5744} & 0.1043 & 0.0939 & 0.0419 & 0.0936 & 0.0117 \\
    \textsc{Single} third template & 0.3675 & 0.4714 & 0.5121 & 0.0572 & 0.0827 & 0.0388 & 0.1025 & 0.0109 \\
    \textsc{All shuffled} & 0.4562 & \textbf{0.5343} & \textbf{0.5861} & 0.0904 & 0.0837 & 0.0375 & 0.2417 & 0.0156 \\
    \textsc{One-at-a-Time} & \textbf{0.4837} & \textbf{0.5445} & \textbf{0.5875} & 0.0842 & 0.0804 & 0.0344 & \textbf{0.2603} & 0.0134 \\
    \textsc{All-in-One-Batch} & 0.4145 & 0.5166 & \textbf{0.5883} & 0.0800 & 0.0814 & 0.0361 & 0.1848 & 0.0132 \\
    \textsc{PPCL} & 0.4612 & 0.5241 & \textbf{0.5904} & 0.0797 & 0.0879 & 0.0400 & 0.1169 & 0.0119 \\
    \textsc{CoIN} & 0.4443 & 0.5139 & 0.5556 & 0.0987 & 0.0734 & \textbf{0.0307} & 0.2095 & \textbf{0.7792} \\
    \textsc{Prompt-TIES} & 0.4236 & 0.5184 & \textbf{0.5767} & \textbf{0.0322} & 0.0828 & 0.0370 & 0.1813 & 0.0120 \\
    \textsc{Prompt-DARE} & 0.3840 & 0.4752 & 0.5349 & 0.0329 & 0.0874 & 0.0410 & 0.1929 & 0.0132 \\
    \midrule
    \multicolumn{9}{l}{\textbf{\LlamaB{}}} \\
    \midrule
    \textsc{Single} first template & 0.5596 & 0.6316 & 0.6738 & 0.0194 & 0.0848 & 0.0379 & 0.1081 & 0.0033 \\
    \textsc{Single} second template & 0.5375 & 0.6198 & 0.6629 & 0.0224 & 0.0822 & 0.0409 & 0.0981 & 0.0033 \\
    \textsc{Single} third template & 0.5739 & 0.6281 & 0.6696 & 0.0183 & 0.0682 & 0.0314 & 0.0933 & 0.0033 \\
    \textsc{All shuffled} & \textbf{0.6211} & \textbf{0.6550} & \textbf{0.6867} & 0.0145 & 0.0677 & 0.0301 & \textbf{0.1324} & 0.0031 \\
    \textsc{One-at-a-Time} & 0.6053 & 0.6428 & 0.6718 & 0.0147 & 0.0635 & 0.0278 & 0.1223 & 0.0031 \\
    \textsc{All-in-One-Batch} & 0.5926 & 0.6348 & 0.6598 & 0.0151 & \textbf{0.0580} & \textbf{0.0258} & 0.1068 & 0.0030 \\
    \textsc{PPCL} & \textbf{0.6139} & 0.6466 & 0.6739 & \textbf{0.0121} & 0.0617 & 0.0283 & 0.1209 & 0.0029 \\
    \textsc{CoIN} & 0.5816 & 0.6235 & 0.6688 & 0.0163 & 0.0640 & 0.0286 & 0.1234 & \textbf{0.7896} \\
    \textsc{Prompt-TIES} & 0.5986 & 0.6320 & 0.6679 & 0.0131 & 0.0628 & 0.0284 & 0.1142 & 0.0029 \\
    \textsc{Prompt-DARE} & 0.5996 & 0.6375 & 0.6695 & 0.0134 & 0.0617 & 0.0278 & 0.1039 & 0.0030 \\
    \midrule
    \multicolumn{9}{l}{\textbf{\QwenI{}}} \\
    \midrule
    \textsc{Single} first template & 0.4306 & 0.5232 & \textbf{0.5819} & 0.1238 & 0.0957 & 0.0856 & 0.0506 & 0.0024 \\
    \textsc{Single} second template & 0.4177 & 0.5141 & 0.5757 & 0.1206 & 0.1052 & 0.0941 & 0.0525 & 0.0025 \\
    \textsc{Single} third template & 0.4142 & 0.5106 & 0.5700 & 0.1187 & 0.0990 & 0.0894 & 0.0476 & 0.0027 \\
    \textsc{All shuffled} & \textbf{0.5089} & \textbf{0.5596} & \textbf{0.6027} & 0.0971 & 0.0953 & 0.0770 & 0.0633 & 0.0016 \\
    \textsc{One-at-a-Time} & 0.4892 & 0.5498 & \textbf{0.5968} & 0.0973 & 0.0765 & \textbf{0.0650} & 0.0555 & 0.0019 \\
    \textsc{All-in-One-Batch} & \textbf{0.5133} & \textbf{0.5609} & \textbf{0.6019} & 0.0891 & 0.0795 & 0.0660 & 0.0594 & 0.0013 \\
    \textsc{PPCL} & 0.4301 & 0.5207 & 0.5835 & 0.1072 & 0.1016 & 0.0913 & 0.0538 & 0.0024 \\
    \textsc{CoIN} & \textbf{0.5091} & \textbf{0.5542} & \textbf{0.5962} & 0.1003 & 0.0832 & 0.0709 & \textbf{0.0764} & \textbf{0.7572} \\
    \textsc{Prompt-TIES} & \textbf{0.5009} & \textbf{0.5456} & \textbf{0.5951} & \textbf{0.0427} & \textbf{0.0747} & \textbf{0.0638} & 0.0611 & 0.0015 \\
    \textsc{Prompt-DARE} & \textbf{0.4980} & 0.5434 & \textbf{0.5864} & \textbf{0.0418} & 0.0781 & 0.0662 & 0.0612 & 0.0015 \\
    \midrule
    \multicolumn{9}{l}{\textbf{\QwenB{}}} \\
    \midrule
    \textsc{Single} first template & 0.5465 & 0.6168 & 0.6826 & 0.0220 & 0.0895 & 0.0814 & 0.0208 & 0.0627 \\
    \textsc{Single} second template & 0.4940 & 0.5639 & 0.6413 & 0.0200 & 0.0907 & 0.0787 & 0.0189 & 0.0480 \\
    \textsc{Single} third template & 0.5093 & 0.5860 & 0.6636 & 0.0244 & 0.0863 & 0.0781 & 0.0194 & 0.0415 \\
    \textsc{All shuffled} & 0.5743 & \textbf{0.6406} & \textbf{0.7052} & 0.0211 & 0.0700 & \textbf{0.0619} & 0.0214 & 0.0244 \\
    \textsc{One-at-a-Time} & 0.5544 & 0.6322 & \textbf{0.7008} & 0.0226 & 0.0708 & 0.0630 & 0.0222 & 0.0269 \\
    \textsc{All-in-One-Batch} & 0.5443 & 0.6158 & 0.6834 & 0.0222 & 0.0700 & 0.0623 & 0.0202 & 0.0178 \\
    \textsc{PPCL} & 0.5383 & 0.6086 & 0.6755 & 0.0215 & \textbf{0.0690} & \textbf{0.0616} & 0.0209 & 0.0146 \\
    \textsc{CoIN} & \textbf{0.5899} & 0.6330 & 0.6720 & \textbf{0.0139} & \textbf{0.0699} & \textbf{0.0632} & \textbf{0.0514} & \textbf{0.7355} \\
    \textsc{Prompt-TIES} & 0.5360 & 0.6099 & 0.6836 & 0.0223 & 0.0698 & \textbf{0.0616} & 0.0201 & 0.0236 \\
    \textsc{Prompt-DARE} & 0.5387 & 0.6109 & 0.6782 & 0.0219 & 0.0740 & 0.0662 & 0.0238 & 0.0133 \\
    \bottomrule
  \end{tabular}
\end{table*}

\begin{table}[htbp]
  \centering
  \caption{Rank classification accuracy scores on the \testall{} split. \textbf{Bold} marks values in the statistically best group per column.}
  \label{tab:test_rank_all}
  \small
  \begin{tabular}{lcccc}
    \toprule
    Method & worst  & avg & best  & clv \\
    \midrule
    \multicolumn{5}{l}{\textbf{\LlamaI{}}} \\
    \midrule
    \textsc{Single} first template & \textbf{0.4025} & 0.4821 & \textbf{0.5447} & 0.1564 \\
    \textsc{Single} second template & \textbf{0.3991} & 0.4864 & \textbf{0.5488} & \textbf{0.1562} \\
    \textsc{Single} third template & 0.3773 & 0.4815 & \textbf{0.5714} & 0.1648 \\
    \textsc{All shuffled} & 0.3919 & \textbf{0.4939} & \textbf{0.5564} & 0.1650 \\
    \textsc{One-at-a-Time} & \textbf{0.4178} & \textbf{0.5043} & \textbf{0.5726} & 0.1715 \\
    \textsc{All-in-One-Batch} & 0.3500 & 0.4706 & \textbf{0.5651} & 0.1695 \\
    \textsc{PPCL} & 0.3837 & 0.4799 & \textbf{0.5469} & 0.1736 \\
    \textsc{CoIN} & \textbf{0.3969} & 0.4835 & \textbf{0.5616} & 0.1631 \\
    \textsc{Prompt-TIES} & 0.3934 & \textbf{0.4947} & \textbf{0.5696} & \textbf{0.1513} \\
    \textsc{Prompt-DARE} & 0.3676 & 0.4645 & \textbf{0.5514} & 0.1630 \\
    \midrule
    \multicolumn{5}{l}{\textbf{\LlamaB{}}} \\
    \midrule
    \textsc{Single} first template & 0.4578 & 0.5446 & 0.6514 & 0.1646 \\
    \textsc{Single} second template & 0.4455 & 0.5332 & 0.6409 & 0.1589 \\
    \textsc{Single} third template & 0.4561 & 0.5315 & 0.6334 & 0.1536 \\
    \textsc{All shuffled} & 0.4555 & 0.5644 & 0.6474 & 0.1535 \\
    \textsc{One-at-a-Time} & 0.4491 & 0.5520 & 0.6557 & 0.1553 \\
    \textsc{All-in-One-Batch} & 0.4365 & 0.5469 & 0.6293 & 0.1435 \\
    \textsc{PPCL} & \textbf{0.4871} & \textbf{0.5873} & \textbf{0.6785} & \textbf{0.1300} \\
    \textsc{CoIN} & 0.4493 & 0.5649 & \textbf{0.6929} & \textbf{0.1312} \\
    \textsc{Prompt-TIES} & 0.4547 & 0.5700 & \textbf{0.6769} & 0.1494 \\
    \textsc{Prompt-DARE} & 0.4217 & 0.5491 & 0.6502 & 0.1407 \\
    \midrule
    \multicolumn{5}{l}{\textbf{\QwenI{}}} \\
    \midrule
    \textsc{Single} first template & \textbf{0.4286} & 0.4776 & 0.5314 & \textbf{0.0650} \\
    \textsc{Single} second template & \textbf{0.4291} & 0.4800 & \textbf{0.5396} & \textbf{0.0642} \\
    \textsc{Single} third template & \textbf{0.4330} & \textbf{0.4862} & \textbf{0.5457} & \textbf{0.0647} \\
    \textsc{All shuffled} & \textbf{0.4309} & \textbf{0.4849} & \textbf{0.5435} & \textbf{0.0641} \\
    \textsc{One-at-a-Time} & \textbf{0.4276} & 0.4752 & 0.5195 & \textbf{0.0648} \\
    \textsc{All-in-One-Batch} & \textbf{0.4295} & 0.4712 & 0.5092 & \textbf{0.0638} \\
    \textsc{PPCL} & \textbf{0.4258} & 0.4739 & 0.5287 & \textbf{0.0662} \\
    \textsc{CoIN} & \textbf{0.4258} & 0.4781 & 0.5256 & \textbf{0.0654} \\
    \textsc{Prompt-TIES} & \textbf{0.4265} & 0.4691 & 0.5158 & \textbf{0.0628} \\
    \textsc{Prompt-DARE} & \textbf{0.4277} & 0.4708 & 0.5180 & \textbf{0.0636} \\
    \midrule
    \multicolumn{5}{l}{\textbf{\QwenB{}}} \\
    \midrule
    \textsc{Single} first template & \textbf{0.4568} & \textbf{0.4966} & \textbf{0.5295} & 0.0589 \\
    \textsc{Single} second template & 0.4541 & 0.4937 & \textbf{0.5265} & 0.0598 \\
    \textsc{Single} third template & 0.4550 & 0.4935 & \textbf{0.5269} & 0.0592 \\
    \textsc{All shuffled} & \textbf{0.4597} & \textbf{0.4979} & \textbf{0.5298} & \textbf{0.0571} \\
    \textsc{One-at-a-Time} & \textbf{0.4619} & \textbf{0.4982} & \textbf{0.5296} & \textbf{0.0567} \\
    \textsc{All-in-One-Batch} & \textbf{0.4579} & 0.4963 & \textbf{0.5295} & 0.0578 \\
    \textsc{PPCL} & \textbf{0.4580} & 0.4956 & \textbf{0.5275} & 0.0579 \\
    \textsc{CoIN} & \textbf{0.4603} & \textbf{0.4979} & \textbf{0.5324} & \textbf{0.0565} \\
    \textsc{Prompt-TIES} & \textbf{0.4584} & 0.4958 & \textbf{0.5288} & 0.0579 \\
    \textsc{Prompt-DARE} & \textbf{0.4603} & 0.4962 & \textbf{0.5268} & \textbf{0.0570} \\
    \bottomrule
  \end{tabular}
\end{table}

\begin{table*}[htbp]
  \centering
  \caption{ROUGE-L scores on the \valmaj{} split. \textbf{Bold} marks values in the statistically best group per column.}
  \label{tab:val_rougeL_majority}
  \footnotesize
  \begin{tabular}{lcccccccc}
    \toprule
    Method & worst  & avg & best  & clv & $\mathrm{JS}^{\text{prompt}}$ & $\mathrm{JS}^{\text{label}}$ & $\mathrm{Diff}^{\text{prompt}}$ & $\mathrm{Diff}^{\text{label}}$ \\
    \midrule
    \multicolumn{9}{l}{\textbf{\LlamaI{}}} \\
    \midrule
    \textsc{Single} first template & 0.4717 & 0.5808 & 0.6531 & 0.0659 & 0.1088 & 0.0466 & 0.2051 & 0.0147 \\
    \textsc{Single} second template & 0.4402 & 0.5774 & 0.6722 & 0.0771 & 0.1230 & 0.0558 & 0.2142 & 0.0149 \\
    \textsc{Single} third template & 0.4139 & 0.5702 & 0.6553 & 0.0759 & 0.1287 & 0.0584 & 0.2142 & 0.0142 \\
    \textsc{All shuffled} & \textbf{0.6622} & \textbf{0.7050} & \textbf{0.7242} & 0.0274 & 0.0315 & 0.0145 & \textbf{0.4265} & 0.0186 \\
    \textsc{One-at-a-Time} & 0.6533 & 0.6997 & \textbf{0.7210} & 0.0299 & 0.0326 & 0.0148 & 0.4141 & 0.0165 \\
    \textsc{All-in-One-Batch} & 0.6289 & 0.6779 & 0.7003 & 0.0297 & 0.0257 & 0.0121 & 0.3320 & 0.0174 \\
    \textsc{PPCL} & 0.6406 & 0.6878 & 0.7104 & 0.0301 & \textbf{0.0242} & \textbf{0.0112} & 0.2542 & 0.0154 \\
    \textsc{CoIN} & 0.5553 & 0.6485 & 0.6921 & 0.0520 & 0.0503 & 0.0240 & 0.3684 & \textbf{0.7907} \\
    \textsc{Prompt-TIES} & 0.6320 & 0.6789 & 0.7011 & \textbf{0.0152} & 0.0257 & 0.0121 & 0.3348 & 0.0158 \\
    \textsc{Prompt-DARE} & 0.6356 & 0.6832 & 0.7058 & \textbf{0.0147} & 0.0260 & 0.0123 & 0.3455 & 0.0161 \\
    \midrule
    \multicolumn{9}{l}{\textbf{\LlamaB{}}} \\
    \midrule
    \textsc{Single} first template & 0.5561 & 0.7068 & 0.7749 & 0.0261 & 0.1139 & 0.0450 & 0.2405 & 0.0048 \\
    \textsc{Single} second template & 0.5434 & 0.6846 & 0.7735 & 0.0287 & 0.1224 & 0.0549 & 0.2408 & 0.0048 \\
    \textsc{Single} third template & 0.5492 & 0.6982 & 0.7698 & 0.0278 & 0.1284 & 0.0533 & 0.2337 & 0.0047 \\
    \textsc{All shuffled} & 0.7389 & 0.7821 & 0.8009 & 0.0065 & 0.0245 & 0.0112 & \textbf{0.3149} & 0.0045 \\
    \textsc{One-at-a-Time} & 0.7317 & 0.7777 & 0.7980 & 0.0068 & 0.0256 & 0.0115 & 0.2944 & 0.0046 \\
    \textsc{All-in-One-Batch} & 0.7223 & 0.7694 & 0.7907 & 0.0075 & 0.0250 & 0.0111 & 0.2634 & 0.0043 \\
    \textsc{PPCL} & \textbf{0.7403} & 0.7831 & 0.8016 & \textbf{0.0062} & \textbf{0.0206} & \textbf{0.0095} & 0.2738 & 0.0042 \\
    \textsc{CoIN} & 0.6657 & 0.7507 & 0.7863 & 0.0149 & 0.0423 & 0.0198 & 0.2843 & \textbf{0.8064} \\
    \textsc{Prompt-TIES} & \textbf{0.7418} & \textbf{0.7847} & \textbf{0.8037} & \textbf{0.0062} & 0.0224 & 0.0105 & 0.2711 & 0.0043 \\
    \textsc{Prompt-DARE} & 0.7332 & 0.7758 & 0.7958 & 0.0070 & 0.0234 & 0.0104 & 0.2701 & 0.0043 \\
    \midrule
    \multicolumn{9}{l}{\textbf{\QwenI{}}} \\
    \midrule
    \textsc{Single} first template & 0.4895 & 0.6222 & 0.6910 & 0.0695 & 0.1264 & 0.0557 & 0.1027 & 0.0035 \\
    \textsc{Single} second template & 0.4599 & 0.6061 & 0.6904 & 0.0754 & 0.1372 & 0.0625 & 0.1058 & 0.0039 \\
    \textsc{Single} third template & 0.4992 & 0.6181 & 0.6925 & 0.0686 & 0.1315 & 0.0573 & 0.0992 & 0.0039 \\
    \textsc{All shuffled} & \textbf{0.6503} & \textbf{0.6982} & \textbf{0.7206} & 0.0333 & 0.0313 & 0.0144 & 0.1351 & 0.0024 \\
    \textsc{One-at-a-Time} & 0.6385 & 0.6908 & \textbf{0.7162} & 0.0363 & 0.0321 & 0.0149 & 0.1267 & 0.0028 \\
    \textsc{All-in-One-Batch} & 0.6349 & 0.6861 & 0.7107 & 0.0342 & 0.0295 & 0.0135 & 0.1283 & 0.0019 \\
    \textsc{PPCL} & 0.5921 & 0.6475 & 0.6768 & 0.0450 & 0.0357 & 0.0167 & 0.1070 & 0.0037 \\
    \textsc{CoIN} & 0.5717 & 0.6612 & 0.7036 & 0.0524 & 0.0486 & 0.0227 & \textbf{0.1622} & \textbf{0.7924} \\
    \textsc{Prompt-TIES} & 0.6385 & 0.6886 & 0.7137 & \textbf{0.0161} & \textbf{0.0282} & \textbf{0.0132} & 0.1361 & 0.0024 \\
    \textsc{Prompt-DARE} & 0.6372 & 0.6879 & 0.7131 & \textbf{0.0161} & \textbf{0.0283} & 0.0133 & 0.1348 & 0.0023 \\
    \midrule
    \multicolumn{9}{l}{\textbf{\QwenB{}}} \\
    \midrule
    \textsc{Single} first template & 0.4685 & 0.5984 & 0.6927 & 0.0248 & 0.1449 & 0.0739 & 0.0298 & 0.0581 \\
    \textsc{Single} second template & 0.3742 & 0.5070 & 0.6155 & 0.0258 & 0.1518 & 0.0797 & 0.0270 & 0.0488 \\
    \textsc{Single} third template & 0.3266 & 0.4569 & 0.5725 & 0.0278 & 0.1463 & 0.0694 & 0.0279 & 0.0428 \\
    \textsc{All shuffled} & \textbf{0.5188} & \textbf{0.6216} & \textbf{0.7014} & 0.0179 & 0.0574 & 0.0278 & 0.0331 & 0.0259 \\
    \textsc{One-at-a-Time} & 0.4741 & 0.5968 & 0.6938 & 0.0222 & 0.0606 & 0.0299 & 0.0339 & 0.0294 \\
    \textsc{All-in-One-Batch} & 0.4664 & 0.5541 & 0.6296 & 0.0162 & 0.0628 & 0.0281 & 0.0311 & 0.0194 \\
    \textsc{PPCL} & 0.4737 & 0.5542 & 0.6253 & 0.0157 & 0.0556 & 0.0245 & 0.0321 & 0.0157 \\
    \textsc{CoIN} & \textbf{0.5158} & 0.5679 & 0.5926 & \textbf{0.0084} & \textbf{0.0390} & \textbf{0.0170} & \textbf{0.0582} & \textbf{0.7351} \\
    \textsc{Prompt-TIES} & 0.4388 & 0.5310 & 0.6235 & 0.0185 & 0.0636 & 0.0290 & 0.0308 & 0.0259 \\
    \textsc{Prompt-DARE} & 0.4565 & 0.5281 & 0.5945 & 0.0152 & 0.0491 & 0.0226 & 0.0353 & 0.0147 \\
    \bottomrule
  \end{tabular}
\end{table*}

\begin{table}[htbp]
  \centering
  \caption{ROUGE-L scores on the \testall{} split for the template-count and template-choice experiments. \textbf{Bold} marks values in the statistically best group per column.}
  \label{tab:test_rougeL_template}
  \small
  \begin{tabular}{lcccc}
    \toprule
    Method & worst & avg & best & clv \\
    \midrule
    \multicolumn{5}{l}{\textbf{\LlamaI{}}} \\
    \midrule
    \textsc{Single} first template & 0.2736 & 0.4807 & \textbf{0.6353} & 0.1490 \\
    \textsc{Single} second template & 0.2967 & 0.4931 & 0.6039 & 0.1329 \\
    \textsc{Single} third template & 0.2633 & 0.4716 & 0.5713 & \textbf{0.1033} \\
    \textsc{Single} random template & 0.2474 & 0.4939 & \textbf{0.6337} & 0.1268 \\
    \textsc{Single} best template & 0.2768 & 0.4493 & 0.5613 & 0.1423 \\
    \textsc{Single} worst template & 0.2183 & 0.4211 & 0.5747 & 0.1474 \\
    \textsc{Single} first two templates & 0.3258 & \textbf{0.5265} & \textbf{0.6330} & 0.1274 \\
    \textsc{Single} first three templates & 0.2685 & 0.4907 & 0.5894 & 0.1204 \\
    \textsc{All shuffled} majority templates & 0.3090 & 0.5219 & \textbf{0.6229} & 0.1141 \\
    \textsc{All shuffled} all templates & \textbf{0.3751} & \textbf{0.5411} & \textbf{0.6302} & 0.1140 \\
    \midrule
    \multicolumn{5}{l}{\textbf{\QwenI{}}} \\
    \midrule
    \textsc{Single} first template & 0.3787 & 0.5211 & \textbf{0.6337} & 0.1333 \\
    \textsc{Single} second template & 0.3659 & 0.5144 & 0.6330 & 0.1380 \\
    \textsc{Single} third template & 0.3768 & 0.5156 & 0.6345 & 0.1352 \\
    \textsc{Single} random template & 0.3855 & 0.5227 & 0.6103 & 0.1368 \\
    \textsc{Single} best template & 0.3519 & 0.5323 & \textbf{0.6368} & 0.1416 \\
    \textsc{Single} worst template & 0.4135 & \textbf{0.5406} & \textbf{0.6500} & \textbf{0.1202} \\
    \textsc{Single} first two templates & 0.3791 & 0.5321 & \textbf{0.6340} & 0.1302 \\
    \textsc{Single} first three templates & 0.3918 & 0.5382 & \textbf{0.6474} & 0.1253 \\
    \textsc{All shuffled} majority templates & \textbf{0.4414} & \textbf{0.5476} & \textbf{0.6369} & 0.1238 \\
    \textsc{All shuffled} all templates & \textbf{0.4515} & \textbf{0.5492} & \textbf{0.6358} & \textbf{0.1191} \\
    \bottomrule
  \end{tabular}
\end{table}

\begin{table}[htbp]
  \centering
  \caption{ROUGE-L scores on the \valmaj{} split for the template-count and template-choice experiments. \textbf{Bold} marks values in the statistically best group per column.}
  \label{tab:val_rougeL_template}
  \small
  \begin{tabular}{lcccc}
    \toprule
    Method & worst & avg & best & clv \\
    \midrule
    \multicolumn{5}{l}{\textbf{\LlamaI{}}} \\
    \midrule
    \textsc{Single} first template & 0.4717 & 0.5808 & 0.6531 & 0.0659 \\
    \textsc{Single} second template & 0.4402 & 0.5774 & 0.6722 & 0.0771 \\
    \textsc{Single} third template & 0.4139 & 0.5702 & 0.6553 & 0.0759 \\
    \textsc{Single} random template & 0.4716 & 0.5931 & 0.6646 & 0.0681 \\
    \textsc{Single} best template & 0.4393 & 0.5575 & 0.6358 & 0.0676 \\
    \textsc{Single} worst template & 0.2844 & 0.3861 & 0.4599 & 0.0944 \\
    \textsc{Single} first two templates & 0.5496 & 0.6608 & 0.7100 & 0.0519 \\
    \textsc{Single} first three templates & 0.5841 & 0.6780 & 0.7148 & 0.0433 \\
    \textsc{All shuffled} majority templates & \textbf{0.6622} & \textbf{0.7050} & \textbf{0.7242} & \textbf{0.0274} \\
    \textsc{All shuffled} all templates & 0.6514 & 0.6994 & \textbf{0.7200} & 0.0287 \\
    \midrule
    \multicolumn{5}{l}{\textbf{\QwenI{}}} \\
    \midrule
    \textsc{Single} first template & 0.4895 & 0.6222 & 0.6910 & 0.0695 \\
    \textsc{Single} second template & 0.4599 & 0.6061 & 0.6904 & 0.0754 \\
    \textsc{Single} third template & 0.4992 & 0.6181 & 0.6925 & 0.0686 \\
    \textsc{Single} random template & 0.4695 & 0.6127 & 0.6877 & 0.0704 \\
    \textsc{Single} best template & 0.4462 & 0.5901 & 0.6846 & 0.0810 \\
    \textsc{Single} worst template & 0.4314 & 0.5634 & 0.6586 & 0.0832 \\
    \textsc{Single} first two templates & 0.5643 & 0.6589 & 0.7045 & 0.0529 \\
    \textsc{Single} first three templates & 0.5982 & 0.6767 & 0.7128 & 0.0449 \\
    \textsc{All shuffled} majority templates & \textbf{0.6503} & 0.6982 & \textbf{0.7206} & \textbf{0.0333} \\
    \textsc{All shuffled} all templates & \textbf{0.6520} & \textbf{0.7019} & \textbf{0.7243} & \textbf{0.0330} \\
    \bottomrule
  \end{tabular}
\end{table}

\clearpage
\section{Qualitative analysis}
\label{app:qualitative}

\begin{table}[p]\centering\scriptsize\renewcommand{\arraystretch}{1.25}
\caption{Examples from the SuperGlue COPA datasets with the highest \clv{} in the \LlamaI{}-\textsc{All shuffled} predictions. Each row is one example, and each column is a prompt template with the placeholders for the example. The COPA task involves selecting the most plausible cause or effect based on a premise and two answer choices. Since the model always generates one of the alternatives, each cell is simplified to the alternative predicted by the template (in bold when correct). The dataset distinguishes between cause and effect questions, which changes the templates slightly. We denote cause examples with C and effect examples with E. We report the number of correct predictions over templates and examples. The per-example view shows local variance, and the per-template view shows global variance.}
\label{tab:copa-full-ash}
\begin{tabular}{p{4.0cm}|ccccccc|c}
\toprule
\textbf{Example} & \rotatebox{90}{\parbox{4.6cm}{\raggedright \textbf{best\_option}\newline\emph{$\langle$P$\rangle$} What's the best option? \mbox{-- C1/E1} \mbox{-- C2/E2} We are looking for a cause/an effect}} & \rotatebox{90}{\parbox{4.6cm}{\raggedright \textbf{cause\_effect}\newline\emph{$\langle$P$\rangle$} Select the most plausible cause:/effect: \mbox{-- $\langle$C1/E1$\rangle$} \mbox{-- $\langle$C2/E2$\rangle$}}} & \rotatebox{90}{\parbox{4.6cm}{\raggedright \textbf{choose}\newline\emph{$\langle$P$\rangle$} because\dots/so\dots{} Choose between: \mbox{-- $\langle$C1/E1$\rangle$} \mbox{-- $\langle$C2/E2$\rangle$}}} & \rotatebox{90}{\parbox{4.6cm}{\raggedright \textbf{exercise}\newline Exercise: choose the most plausible alternative. \emph{$\langle$P$\rangle$} because\dots/so\dots{} \mbox{-- $\langle$C1/E1$\rangle$} \mbox{-- $\langle$C2/E2$\rangle$}}} & \rotatebox{90}{\parbox{4.6cm}{\raggedright \textbf{more likely}\newline Pick the more likely continuation to the following sentence: \emph{$\langle$P$\rangle$} as a result of:/as a consequence: \mbox{-- $\langle$C1/E1$\rangle$} \mbox{-- $\langle$C2/E2$\rangle$}}} & \rotatebox{90}{\parbox{4.6cm}{\raggedright \textbf{plausible\_alt.}\newline\emph{$\langle$P$\rangle$} This happened because\dots/As a consequence\dots{} Help me pick the more plausible option: \mbox{-- $\langle$C1/E1$\rangle$} \mbox{-- $\langle$C2/E2$\rangle$}}} & \rotatebox{90}{\parbox{4.6cm}{\raggedright \textbf{i\_am\_hesitating}\newline\emph{$\langle$P$\rangle$} I am hesitating between two options. Help me choose the more likely cause:/effect: \mbox{-- $\langle$C1/E1$\rangle$} \mbox{-- $\langle$C2/E2$\rangle$}}} & \rotatebox{90}{\textbf{\# templates right}} \\
\midrule
\textbf{P:} The man received a parking ticket\newline C1: He parallel parked on the street\newline \textbf{C2: The parking meter expired} & C1 & \textbf{C2} & \textbf{C2} & \textbf{C2} & C1 & C1 & \textbf{C2} & 4 \\
\midrule
\textbf{P:} My view of the movie screen was blocked\newline C1: The couple behind me was whispering\newline \textbf{C2: A tall person was sitting in front of me} & \textbf{C2} & \textbf{C2} & C1 & C1 & \textbf{C2} & C1 & C1 & 3 \\
\midrule
\textbf{P:} The bride got cold feet before the wedding\newline E1: The wedding guests brought gifts\newline \textbf{E2: She called the wedding off} & E1 & \textbf{E2} & E1 & E1 & \textbf{E2} & \textbf{E2} & E1 & 3 \\
\midrule
\textbf{P:} I spent the day at the pool\newline E1: I sprained my ankle\newline \textbf{E2: My face got sunburned} & \textbf{E2} & \textbf{E2} & E1 & E1 & E1 & E1 & \textbf{E2} & 3 \\
\midrule
\textbf{P:} The girl found a bug in her cereal\newline E1: She poured milk in the bowl\newline \textbf{E2: She lost her appetite} & E1 & \textbf{E2} & E1 & E1 & \textbf{E2} & \textbf{E2} & \textbf{E2} & 4 \\
\midrule
\textbf{P:} The couple got engaged\newline \textbf{E1: They planned a wedding}\newline E2: They took some time apart & E2 & E2 & \textbf{E1} & \textbf{E1} & E2 & E2 & \textbf{E1} & 3 \\
\midrule
\textbf{\# examples right} & \textbf{2} & \textbf{5} & \textbf{2} & \textbf{2} & \textbf{3} & \textbf{2} & \textbf{4} & 20 \\
\bottomrule
\end{tabular}
\end{table}

\begin{table}[p]\centering\scriptsize\renewcommand{\arraystretch}{1.25}
\caption{Examples from the SuperGlue COPA datasets with the highest \clv{} in the \LlamaI{}-\textsc{All shuffled} predictions. Here we list the predictions from \LlamaI{}-\textsc{Prompt-TIES}. Each row is one example, and each column is a prompt template with the placeholders for the example. The COPA task involves selecting the most plausible cause or effect based on a premise and two answer choices. Since the model always generates one of the alternatives, each cell is simplified to the alternative predicted by the template (in bold when correct). The dataset distinguishes between cause and effect questions, which changes the templates slightly. We denote cause examples with C and effect examples with E. We report the number of correct predictions over templates and examples. The per-example view shows local variance, and the per-template view shows global variance.}
\label{tab:copa-full-ties}
\begin{tabular}{p{4.0cm}|ccccccc|c}
\toprule
\textbf{Example} & \rotatebox{90}{\parbox{4.6cm}{\raggedright \textbf{best\_option}\newline\emph{$\langle$P$\rangle$} What's the best option? \mbox{-- $\langle$C1/E1$\rangle$} \mbox{-- $\langle$C2/E2$\rangle$} We are looking for a cause/an effect}} & \rotatebox{90}{\parbox{4.6cm}{\raggedright \textbf{cause\_effect}\newline\emph{$\langle$P$\rangle$} Select the most plausible cause:/effect: \mbox{-- $\langle$C1/E1$\rangle$} \mbox{-- $\langle$C2/E2$\rangle$}}} & \rotatebox{90}{\parbox{4.6cm}{\raggedright \textbf{choose}\newline\emph{$\langle$P$\rangle$} because\dots/so\dots{} Choose between: \mbox{-- $\langle$C1/E1$\rangle$} \mbox{-- $\langle$C2/E2$\rangle$}}} & \rotatebox{90}{\parbox{4.6cm}{\raggedright \textbf{exercise}\newline Exercise: choose the most plausible alternative. \emph{$\langle$P$\rangle$} because\dots/so\dots{} \mbox{-- $\langle$C1/E1$\rangle$} \mbox{-- $\langle$C2/E2$\rangle$}}} & \rotatebox{90}{\parbox{4.6cm}{\raggedright \textbf{more likely}\newline Pick the more likely continuation to the following sentence: \emph{$\langle$P$\rangle$} as a result of:/as a consequence: \mbox{-- $\langle$C1/E1$\rangle$} \mbox{-- $\langle$C2/E2$\rangle$}}} & \rotatebox{90}{\parbox{4.6cm}{\raggedright \textbf{plausible\_alt.}\newline\emph{$\langle$P$\rangle$} This happened because\dots/As a consequence\dots{} Help me pick the more plausible option: \mbox{-- $\langle$C1/E1$\rangle$} \mbox{-- $\langle$C2/E2$\rangle$}}} & \rotatebox{90}{\parbox{4.6cm}{\raggedright \textbf{i\_am\_hesitating}\newline\emph{$\langle$P$\rangle$} I am hesitating between two options. Help me choose the more likely cause:/effect: \mbox{-- $\langle$C1/E1$\rangle$} \mbox{-- $\langle$C2/E2$\rangle$}}} & \rotatebox{90}{\textbf{\# templates right}} \\
\midrule
\textbf{P:} The man received a parking ticket\newline C1: He parallel parked on the street\newline \textbf{C2: The parking meter expired} & C1 & \textbf{C2} & \textbf{C2} & \textbf{C2} & C1 & \textbf{C2} & \textbf{C2} & 5 \\
\midrule
\textbf{P:} My view of the movie screen was blocked\newline C1: The couple behind me was whispering\newline \textbf{C2: A tall person was sitting in front of me} & \textbf{C2} & \textbf{C2} & \textbf{C2} & \textbf{C2} & \textbf{C2} & \textbf{C2} & \textbf{C2} & 7 \\
\midrule
\textbf{P:} The bride got cold feet before the wedding\newline E1: The wedding guests brought gifts\newline \textbf{E2: She called the wedding off} & \textbf{E2} & \textbf{E2} & \textbf{E2} & \textbf{E2} & \textbf{E2} & \textbf{E2} & \textbf{E2} & 7 \\
\midrule
\textbf{P:} I spent the day at the pool\newline E1: I sprained my ankle\newline \textbf{E2: My face got sunburned} & \textbf{E2} & \textbf{E2} & \textbf{E2} & \textbf{E2} & \textbf{E2} & \textbf{E2} & \textbf{E2} & 7 \\
\midrule
\textbf{P:} The girl found a bug in her cereal\newline E1: She poured milk in the bowl\newline \textbf{E2: She lost her appetite} & E1 & E1 & E1 & E1 & \textbf{E2} & \textbf{E2} & \textbf{E2} & 3 \\
\midrule
\textbf{P:} The couple got engaged\newline \textbf{E1: They planned a wedding}\newline E2: They took some time apart & E2 & E2 & E2 & E2 & E2 & E2 & E2 & 0 \\
\midrule
\textbf{\# examples right} & \textbf{3} & \textbf{4} & \textbf{4} & \textbf{4} & \textbf{4} & \textbf{5} & \textbf{5} & 29 \\
\bottomrule
\end{tabular}
\end{table}

\clearpage
\section{Impact of the threshold choice on conditional local variance}
\label{app:clv_threshold}

In this section, we demonstrate the impact on the \clv{} by choosing a Rouge-L threshold. We report the \clv{} based on the choice of $\theta$ for each method. All results are on the \testall{} split. The results are split by model: \LlamaI{} \autoref{tab:clv-solvable-llama-1b}, \LlamaB{} \autoref{tab:clv-solvable-llama-8b}, \QwenI{} \autoref{tab:clv-solvable-qwen-1b}, and \QwenB{} \autoref{tab:clv-solvable-qwen-8b}. We demonstrate that changing the threshold alters the \clv{} but leaves the ranking of the best model unchanged. Note that we do not compute significant scores and rather only bold the lowest value.
\begin{table}[htbp]
\centering
\small
\caption{We report the \clv{} based on the choice of $\theta$ for each method on \LlamaI{}. All results are on the \testall{} split. \textbf{Bold} marks the run with the lowest conditional local variance.}
\label{tab:clv-solvable-llama-1b}
\begin{tabular}{l c ccccccc}
\toprule
 & & \multicolumn{7}{c}{Conditional local variance at solvability threshold $\theta $} \\
\cmidrule(lr){3-9}
Method & Local & $0.375$ & $0.5$ & $0.625$ & $0.75$ & $0.875$ & $0.9999$ & $1$ \\
\midrule
\textbf{dare\_ties} & \textbf{0.0616} & \textbf{0.0626} & \textbf{0.0626} & \textbf{0.0857} & \textbf{0.0857} & \textbf{0.0902} & \textbf{0.0905} & \textbf{0.0905} \\
ties & 0.0651 & 0.0668 & 0.0669 & 0.0913 & 0.0914 & 0.0951 & 0.0952 & 0.0952 \\
single\_2 & 0.0892 & 0.1033 & 0.1035 & 0.1102 & 0.1106 & 0.1111 & 0.1120 & 0.1120 \\
all\_shuffled & 0.0974 & 0.1141 & 0.1142 & 0.1201 & 0.1203 & 0.1205 & 0.1206 & 0.1206 \\
coin & 0.0982 & 0.1187 & 0.1188 & 0.1225 & 0.1229 & 0.1231 & 0.1243 & 0.1243 \\
one\_at\_a\_time & 0.0985 & 0.1145 & 0.1146 & 0.1186 & 0.1187 & 0.1188 & 0.1188 & 0.1188 \\
all\_in\_one\_batch & 0.1122 & 0.1194 & 0.1196 & 0.1252 & 0.1255 & 0.1257 & 0.1267 & 0.1267 \\
ppcl & 0.1166 & 0.1244 & 0.1245 & 0.1289 & 0.1290 & 0.1290 & 0.1291 & 0.1291 \\
single\_1 & 0.1251 & 0.1329 & 0.1330 & 0.1395 & 0.1403 & 0.1409 & 0.1417 & 0.1417 \\
single\_0 & 0.1290 & 0.1490 & 0.1491 & 0.1498 & 0.1499 & 0.1499 & 0.1499 & 0.1499 \\
\bottomrule
\end{tabular}
\end{table}

\begin{table}[htbp]
\centering
\small
\caption{We report the \clv{} based on the choice of $\theta$ for each method on \LlamaB{}. All results are on the \testall{} split. \textbf{Bold} marks the run with the lowest conditional local variance.}
\label{tab:clv-solvable-llama-8b}
\begin{tabular}{l c ccccccc}
\toprule
 & & \multicolumn{7}{c}{Conditional local variance at solvability threshold $\theta $} \\
\cmidrule(lr){3-9}
Method & Local & $0.375$ & $0.5$ & $0.625$ & $0.75$ & $0.875$ & $0.9999$ & $1$ \\
\midrule
\textbf{ppcl} & \textbf{0.0221} & \textbf{0.0224} & \textbf{0.0225} & \textbf{0.0246} & \textbf{0.0250} & \textbf{0.0304} & \textbf{0.0312} & \textbf{0.0312} \\
ties & 0.0257 & 0.0261 & 0.0264 & 0.0288 & 0.0302 & 0.0373 & 0.0393 & 0.0393 \\
dare\_ties & 0.0288 & 0.0292 & 0.0293 & 0.0325 & 0.0341 & 0.0441 & 0.0471 & 0.0471 \\
one\_at\_a\_time & 0.0289 & 0.0293 & 0.0293 & 0.0316 & 0.0328 & 0.0401 & 0.0425 & 0.0425 \\
all\_shuffled & 0.0309 & 0.0311 & 0.0311 & 0.0330 & 0.0343 & 0.0409 & 0.0438 & 0.0438 \\
all\_in\_one\_batch & 0.0336 & 0.0340 & 0.0340 & 0.0368 & 0.0387 & 0.0483 & 0.0513 & 0.0513 \\
single\_2 & 0.0338 & 0.0342 & 0.0342 & 0.0364 & 0.0375 & 0.0461 & 0.0490 & 0.0490 \\
coin & 0.0342 & 0.0345 & 0.0345 & 0.0369 & 0.0377 & 0.0450 & 0.0477 & 0.0477 \\
single\_0 & 0.0360 & 0.0363 & 0.0363 & 0.0391 & 0.0398 & 0.0488 & 0.0511 & 0.0511 \\
single\_1 & 0.0381 & 0.0383 & 0.0383 & 0.0404 & 0.0416 & 0.0514 & 0.0543 & 0.0543 \\
\bottomrule
\end{tabular}
\end{table}

\begin{table}[htbp]
\centering
\small
\caption{We report the \clv{} based on the choice of $\theta$ for each method on \QwenI{}. All results are on the \testall{} split. \textbf{Bold} marks the run with the lowest conditional local variance.}
\label{tab:clv-solvable-qwen-1b}
\begin{tabular}{l c ccccccc}
\toprule
 & & \multicolumn{7}{c}{Conditional local variance at solvability threshold $\theta $} \\
\cmidrule(lr){3-9}
Method & Local & $0.375$ & $0.5$ & $0.625$ & $0.75$ & $0.875$ & $0.9999$ & $1$ \\
\midrule
\textbf{ties} & \textbf{0.0531} & \textbf{0.0559} & \textbf{0.0560} & \textbf{0.0728} & \textbf{0.0728} & \textbf{0.0763} & \textbf{0.0766} & \textbf{0.0766} \\
dare\_ties & 0.0534 & 0.0562 & 0.0563 & 0.0735 & 0.0734 & 0.0772 & 0.0775 & 0.0775 \\
all\_in\_one\_batch & 0.0944 & 0.1112 & 0.1113 & 0.1178 & 0.1178 & 0.1179 & 0.1179 & 0.1179 \\
all\_shuffled & 0.1121 & 0.1238 & 0.1239 & 0.1268 & 0.1269 & 0.1269 & 0.1269 & 0.1269 \\
one\_at\_a\_time & 0.1134 & 0.1238 & 0.1239 & 0.1285 & 0.1285 & 0.1285 & 0.1287 & 0.1287 \\
coin & 0.1177 & 0.1300 & 0.1301 & 0.1345 & 0.1345 & 0.1346 & 0.1347 & 0.1347 \\
single\_0 & 0.1200 & 0.1333 & 0.1334 & 0.1373 & 0.1373 & 0.1373 & 0.1374 & 0.1374 \\
single\_2 & 0.1220 & 0.1352 & 0.1353 & 0.1376 & 0.1376 & 0.1376 & 0.1377 & 0.1377 \\
single\_1 & 0.1243 & 0.1380 & 0.1381 & 0.1409 & 0.1409 & 0.1409 & 0.1409 & 0.1409 \\
ppcl & 0.1272 & 0.1425 & 0.1426 & 0.1449 & 0.1449 & 0.1450 & 0.1450 & 0.1450 \\
\bottomrule
\end{tabular}
\end{table}

\begin{table}[htbp]
\centering
\small
\caption{We report the \clv{} based on the choice of $\theta$ for each method on \QwenB{}. All results are on the \testall{} split. \textbf{Bold} marks the run with the lowest conditional local variance.}
\label{tab:clv-solvable-qwen-8b}
\begin{tabular}{l c ccccccc}
\toprule
 & & \multicolumn{7}{c}{Conditional local variance at solvability threshold $\theta $} \\
\cmidrule(lr){3-9}
Method & Local & $0.375$ & $0.5$ & $0.625$ & $0.75$ & $0.875$ & $0.9999$ & $1$ \\
\midrule
\textbf{single\_1} & \textbf{0.0290} & \textbf{0.0294} & \textbf{0.0300} & \textbf{0.0327} & 0.0347 & \textbf{0.0417} & \textbf{0.0456} & \textbf{0.0456} \\
coin & 0.0298 & 0.0300 & 0.0303 & 0.0333 & \textbf{0.0345} & 0.0443 & 0.0557 & 0.0557 \\
single\_0 & 0.0329 & 0.0331 & 0.0333 & 0.0354 & 0.0365 & 0.0439 & 0.0467 & 0.0467 \\
dare\_ties & 0.0330 & 0.0332 & 0.0336 & 0.0359 & 0.0382 & 0.0461 & 0.0497 & 0.0497 \\
one\_at\_a\_time & 0.0332 & 0.0334 & 0.0335 & 0.0355 & 0.0371 & 0.0450 & 0.0477 & 0.0477 \\
all\_shuffled & 0.0334 & 0.0334 & 0.0335 & 0.0353 & 0.0364 & 0.0433 & 0.0459 & 0.0459 \\
all\_in\_one\_batch & 0.0335 & 0.0336 & 0.0339 & 0.0358 & 0.0377 & 0.0453 & 0.0492 & 0.0492 \\
ties & 0.0341 & 0.0343 & 0.0346 & 0.0370 & 0.0395 & 0.0487 & 0.0513 & 0.0513 \\
single\_2 & 0.0343 & 0.0348 & 0.0355 & 0.0385 & 0.0413 & 0.0496 & 0.0491 & 0.0491 \\
ppcl & 0.0354 & 0.0355 & 0.0357 & 0.0381 & 0.0400 & 0.0487 & 0.0527 & 0.0527 \\
\bottomrule
\end{tabular}
\end{table}
\clearpage
\section{Gradient interference measurements}
\label{app:gradient_interference}

This section reports the raw values behind \autoref{sec:gradient_interference}. We measure the four gradient-interference statistics of \autoref{sec:grad_interference_measure}: cosine similarity ($\mathrm{cos}$), sign disagreement ($\mathrm{sign\text{-}conflict}$), cancellation ($\mathrm{cancellation}$) and effective rank ($\mathrm{eff\text{-}rank}$), on the \textsc{All-in-One-Batch} runs of the two Llama-family models. \autoref{tab:grad_interference_steps} tracks the four statistics over training, and \autoref{tab:grad_interference_layers_1b} and \autoref{tab:grad_interference_layers_8b} resolve cosine similarity and sign disagreement by layer and module.

\begin{table}[htbp]
  \centering
  \small
  \caption{Gradient interference metrics (see \autoref{sec:grad_interference_measure}) were measured every 1,000 optimizer steps on a subset of the validation split during \textsc{All-in-One-Batch} training. The \textbf{Avg.} column reports the mean over all steps for \LlamaI{}, and the mean over all steps excluding $t{=}0$ for \LlamaB{}. $t{=}0$ is an outlier for \LlamaB{} as the LoRA adapters are zero-initialized, and evaluation occurs before training.}
  \label{tab:grad_interference_steps}
  \begin{tabular}{lrrrrrrrr}
    \toprule
    Statistic & $t{=}0$ & $1001$ & $2002$ & $3003$ & $4004$ & $5005$ & $6006$ & \textbf{Avg.} \\
    \midrule
    \multicolumn{9}{l}{\textbf{\LlamaI{}}} \\
    \midrule
    $\mathrm{cos}$            & 0.531 & 0.532 & 0.541 & 0.546 & 0.550 & 0.551 & 0.549 & 0.543 \\
    $\mathrm{cos}_{\min}$     & 0.289 & 0.294 & 0.293 & 0.292 & 0.299 & 0.295 & 0.295 & 0.294 \\
    $\mathrm{sign\text{-}conflict}$ & 0.615 & 0.571 & 0.564 & 0.565 & 0.566 & 0.560 & 0.561 & 0.572 \\
    $\mathrm{cancellation}$   & 0.217 & 0.204 & 0.202 & 0.194 & 0.189 & 0.185 & 0.191 & 0.197 \\
    $\mathrm{eff\text{-}rank}$ & 3.273 & 2.988 & 2.852 & 2.836 & 2.802 & 2.805 & 2.810 & 2.909 \\
    \midrule
    \multicolumn{9}{l}{\textbf{\LlamaB{}}} \\
    \midrule
    $\mathrm{cos}$            & 0.502 & 0.514 & 0.564 & 0.567 & 0.556 & 0.551 & 0.544 & 0.549 \\
    $\mathrm{cos}_{\min}$     & 0.274 & 0.251 & 0.313 & 0.311 & 0.295 & 0.272 & 0.272 & 0.286 \\
    $\mathrm{sign\text{-}conflict}$ & 0.379 & 0.667 & 0.640 & 0.634 & 0.634 & 0.629 & 0.630 & 0.639 \\
    $\mathrm{cancellation}$   & 0.232 & 0.204 & 0.169 & 0.167 & 0.161 & 0.169 & 0.181 & 0.175 \\
    $\mathrm{eff\text{-}rank}$ & 3.505 & 3.148 & 2.860 & 2.815 & 2.974 & 2.854 & 2.985 & 2.939 \\
    \bottomrule
  \end{tabular}
\end{table}

\begin{table}[htbp]
  \centering
  \small
  \caption{The per-layer cosine similarity ($\mathrm{cos}$) and sign disagreement ($\mathrm{sign\text{-}conflict}$) (see \autoref{sec:grad_interference_measure}) for \LlamaI{}, split into the attention and feed-forward sublayers. These values were measured every 1,000 optimizer steps on a subset of the validation split during \textsc{All-in-One-Batch} training and averaged over training.}
  \label{tab:grad_interference_layers_1b}
  \begin{tabular}{lrrrr}
    \toprule
    & \multicolumn{2}{c}{Attention} & \multicolumn{2}{c}{Feed-forward} \\
    \cmidrule(lr){2-3}\cmidrule(lr){4-5}
    Layer & $\mathrm{cos}$ & $\mathrm{sign\text{-}conflict}$ & $\mathrm{cos}$ & $\mathrm{sign\text{-}conflict}$ \\
    \midrule
    0  & 0.483 & 0.714 & 0.499 & 0.687 \\
    1  & 0.485 & 0.710 & 0.547 & 0.692 \\
    2  & 0.480 & 0.715 & 0.484 & 0.694 \\
    3  & 0.495 & 0.704 & 0.491 & 0.690 \\
    4  & 0.502 & 0.703 & 0.500 & 0.685 \\
    5  & 0.512 & 0.692 & 0.507 & 0.681 \\
    6  & 0.526 & 0.683 & 0.512 & 0.680 \\
    7  & 0.536 & 0.683 & 0.519 & 0.676 \\
    8  & 0.543 & 0.675 & 0.534 & 0.664 \\
    9  & 0.561 & 0.662 & 0.550 & 0.644 \\
    10 & 0.566 & 0.648 & 0.560 & 0.628 \\
    11 & 0.572 & 0.635 & 0.583 & 0.610 \\
    12 & 0.587 & 0.620 & 0.591 & 0.601 \\
    13 & 0.609 & 0.609 & 0.600 & 0.587 \\
    14 & 0.605 & 0.604 & 0.627 & 0.568 \\
    15 & 0.626 & 0.579 & 0.646 & 0.543 \\
    \bottomrule
  \end{tabular}
\end{table}

\begin{table}[htbp]
  \centering
  \footnotesize
  \caption{The per-layer cosine similarity ($\mathrm{cos}$) and sign disagreement ($\mathrm{sign\text{-}conflict}$) (see \autoref{sec:grad_interference_measure}) for \LlamaB{}, split into the attention and feed-forward sublayers. These values were measured every 1,000 optimizer steps on a subset of the validation split during \textsc{All-in-One-Batch} training and averaged over training (except for step $t=0$, see \autoref{tab:grad_interference_steps} for details).}
  \label{tab:grad_interference_layers_8b}
  \begin{tabular}{lrrrr}
    \toprule
    & \multicolumn{2}{c}{Attention} & \multicolumn{2}{c}{Feed-forward} \\
    \cmidrule(lr){2-3}\cmidrule(lr){4-5}
    Layer & $\mathrm{cos}$ & $\mathrm{sign\text{-}conflict}$ & $\mathrm{cos}$ & $\mathrm{sign\text{-}conflict}$ \\
    \midrule
    0  & 0.347 & 0.811 & 0.336 & 0.780 \\
    1  & 0.343 & 0.783 & 0.450 & 0.767 \\
    2  & 0.353 & 0.788 & 0.374 & 0.765 \\
    3  & 0.365 & 0.776 & 0.368 & 0.767 \\
    4  & 0.382 & 0.776 & 0.371 & 0.764 \\
    5  & 0.381 & 0.767 & 0.383 & 0.757 \\
    6  & 0.422 & 0.748 & 0.404 & 0.748 \\
    7  & 0.449 & 0.733 & 0.438 & 0.730 \\
    8  & 0.465 & 0.718 & 0.454 & 0.719 \\
    9  & 0.475 & 0.715 & 0.468 & 0.708 \\
    10 & 0.490 & 0.700 & 0.495 & 0.693 \\
    11 & 0.511 & 0.683 & 0.513 & 0.684 \\
    12 & 0.507 & 0.684 & 0.525 & 0.679 \\
    13 & 0.521 & 0.676 & 0.538 & 0.667 \\
    14 & 0.528 & 0.668 & 0.539 & 0.663 \\
    15 & 0.523 & 0.666 & 0.550 & 0.656 \\
    16 & 0.519 & 0.661 & 0.543 & 0.651 \\
    17 & 0.528 & 0.658 & 0.539 & 0.644 \\
    18 & 0.510 & 0.655 & 0.536 & 0.642 \\
    19 & 0.511 & 0.655 & 0.540 & 0.634 \\
    20 & 0.529 & 0.641 & 0.540 & 0.631 \\
    21 & 0.539 & 0.640 & 0.544 & 0.628 \\
    22 & 0.524 & 0.640 & 0.549 & 0.623 \\
    23 & 0.537 & 0.633 & 0.546 & 0.620 \\
    24 & 0.536 & 0.634 & 0.551 & 0.622 \\
    25 & 0.539 & 0.629 & 0.550 & 0.616 \\
    26 & 0.558 & 0.616 & 0.557 & 0.611 \\
    27 & 0.559 & 0.611 & 0.570 & 0.603 \\
    28 & 0.566 & 0.605 & 0.575 & 0.595 \\
    29 & 0.580 & 0.593 & 0.589 & 0.586 \\
    30 & 0.566 & 0.595 & 0.603 & 0.574 \\
    31 & 0.600 & 0.573 & 0.616 & 0.550 \\
    \bottomrule
  \end{tabular}
\end{table}

\end{document}

%% file: core/preamble/document_settings.tex
\newif\ifprintVersion   
\newif\ifprofessionalPrint 
\newif\iffancyTheorems  
\newif\ifboldNumberSets 
\newif\ifbachelorThesis 

\printVersionfalse
\professionalPrintfalse
\fancyTheoremstrue
\boldNumberSetstrue
\bachelorThesisfalse

\newcommand*{\printTitle}{}
\newcommand*{\myTitle}[1]{\renewcommand*{\printTitle}{#1}}
\newcommand*{\printTitleBold}{\textbf{\printTitle}}

\newcommand*{\printAuthor}{}
\newcommand*{\myName}[1]{\renewcommand*{\printAuthor}{#1}}

\newcommand*{\printProgram}{}
\newcommand*{\myProgram}[1]{\renewcommand*{\printProgram}{#1}}

\newcommand*{\printDateReceived}{}
\newcommand*{\dateOfHandingIn}[1]{\renewcommand*{\printDateReceived}{#1}}

\newcommand*{\printSubject}{}
\newcommand*{\mySubject}[1]{\renewcommand*{\printSubject}{#1}}

\newcommand*{\printKeywords}{}
\newcommand*{\myKeywords}[1]{\renewcommand*{\printKeywords}{#1}}

\newcommand*{\printNameOfSupervisor}{}
\newcommand*{\nameOfMySupervisor}[1]{\renewcommand*{\printNameOfSupervisor}{#1}}

\newcommand*{\printAdditionalExaminers}{}
\newcommand*{\additionalExaminers}[1]{\renewcommand*{\printAdditionalExaminers}{#1}}

\newlength{\extraborderlength}
\newcommand*{\extraBorder}[1]{\setlength{\extraborderlength}{#1}}

\newlength{\mybindingcorrection}
\newcommand*{\bindingCorrection}[1]{\setlength{\mybindingcorrection}{#1}}

%% file: core/preamble/format.tex
\usepackage[utf8]{inputenc} 
\usepackage[T1]{fontenc}    
\usepackage
[
    ngerman,         
    main = american, 
]
{babel}                     

\input glyphtounicode
\usepackage{calc} 

\newlength{\myparindent}
\newlength{\myparskip}
\setfootnoterule{0 cm}

\deffootnote[1.2 em]{1.2 em}{0 em}{\makebox[1.4 em][l]{\textbf{\thefootnotemark}}}

\makeatletter%
\@removefromreset{footnote}{chapter}%
\makeatother

\usepackage[dvipsnames]{xcolor} 

\definecolor{stroke1}{HTML}{2574A9} 

\colorlet{captionlabel}{black}
\colorlet{footerpagenr}{black}
\colorlet{footerchapter}{stroke1}
\colorlet{footerchaptername}{black}
\colorlet{footersection}{stroke1}
\colorlet{footersectionname}{black}
\colorlet{chapternumber}{stroke1}

\newlength{\mypaperwidth}
\newlength{\mypaperheight}
\newlength{\mybodywidth}
\newlength{\mybodyheight}
\newlength{\myoutermargin}
\ifprintVersion
    \ifprofessionalPrint
    \else
    \fi
\else
\fi

\newlength{\mytopmargin}
\ifprintVersion
    \ifprofessionalPrint
    \fi
\fi

\newlength{\myinnermargin}
\ifprintVersion
    \ifprofessionalPrint
    \fi
\fi

\newlength{\mybottommargin}
\ifprintVersion
    \ifprofessionalPrint
    \fi
\fi

\newcommand{\goldenratio}{1.618}

\newlength{\myheadsep} 
\ifprintVersion
    \ifprofessionalPrint
    \fi
\fi

\newlength{\myfootskip} 
\ifprintVersion
    \ifprofessionalPrint
    \fi
\fi

\newlength{\mymargininnersep} 
\newlength{\mymarginoutersep} 
\ifprintVersion
    \ifprofessionalPrint
    \fi
\fi

\newlength{\mymarginwidth} 
\newlength{\mymarginwidthwithinnersep} 
\usepackage
[
    \ifprintVersion
        \ifprofessionalPrint
            paperwidth = \mypaperwidth + 2\extraborderlength + \mybindingcorrection,
            paperheight = \mypaperheight + 2\extraborderlength,
        \else
            paperwidth = \mypaperwidth,
            paperheight = \mypaperheight,
        \fi
    \else
        paperwidth = \mypaperwidth,
        paperheight = \mypaperheight,
    \fi
    textwidth = \mybodywidth,
    textheight = \mybodyheight,
    outer = \myoutermargin,
    top = \mytopmargin,
    headsep = \myheadsep,
    footskip = \myfootskip,
    marginparsep = \mymargininnersep,
    marginparwidth = \mymarginwidth,
]
{geometry} 

\usepackage
[
]
{scrlayer-scrpage} 

\clearpairofpagestyles

\KOMAoptions
{%
    headwidth = \textwidth + \mymarginwidthwithinnersep,%
    footwidth = \myoutermargin : \textwidth,%
}

\automark[chapter]{chapter}
\automark*[section]{}

\lehead%
{%
    \begin{minipage}[b]{\mymarginwidth}%
        \small\raggedleft\normalfont\textsf{\textbf{\color{footerchapter}\chaptername\ \thechapter}}
    \end{minipage}
}
\cehead{\hspace*{\mymarginwidthwithinnersep}\parbox{\textwidth}{\raggedright\leftmark}}

\rohead%
{%
    \Ifstr{\rightmark}{\leftmark}%
    {%
        \begin{minipage}[b]{\mymarginwidth}%
            \small\raggedright\normalfont\textsf{\textbf{\color{footersection}Chapter\ \thechapter}}%
        \end{minipage}%
    }%
    {%
        \begin{minipage}[b]{\mymarginwidth}%
            \small\raggedright\normalfont\textsf{\textbf{\color{footersection}Section\ \thesection}}%
        \end{minipage}%
    }%
}
\cohead{\hspace*{-\mymarginwidthwithinnersep}\parbox{\textwidth}{\raggedleft\rightmark}}

\lefoot*%
{%
    \vspace*{1 ex}%
    {\color{stroke1}\rule{\myoutermargin - \mymargininnersep}{0.5 mm}}\\
    \begin{minipage}[b]{\myoutermargin - \mymargininnersep}%
        \raggedleft\normalfont\color{footerpagenr}\textbf{\thepage}%
    \end{minipage}%
}
\rofoot*%
{%
    \vspace*{1 ex}%
    {\color{stroke1}\rule{\myoutermargin - \mymargininnersep}{0.5 mm}}\\
    \begin{minipage}[b]{\myoutermargin - \mymargininnersep}%
        \raggedright\normalfont\color{footerpagenr}\textbf{\thepage}%
    \end{minipage}%
}

\usepackage{caption}
\usepackage{tikz} 

\newlength{\mytmpa}
\newlength{\mytmpb}

\renewcommand*{\partlineswithprefixformat}[3]%
{%
    #2
    \thispagestyle{empty}
    \ifprintVersion
        \setlength{\mytmpa}{0.618\mypaperwidth + \mybindingcorrection + \extraborderlength}%
        \setlength{\mytmpb}{0.382\mypaperheight + \extraborderlength}%
    \else
        \setlength{\mytmpa}{0.618\mypaperwidth}%
        \setlength{\mytmpb}{0.382\mypaperheight}%
    \fi
    \begin{tikzpicture}[overlay, remember picture]%
        \node [inner sep = 0, outer sep = 0, anchor = north] at (current page.north west)%
        {%
            \begin{tikzpicture}[overlay, remember picture]%
                \draw[color = stroke1, line width = 0.7 mm] (\mytmpa, 0) -- (\mytmpa, -\mytmpb);%
            \end{tikzpicture}%
        };%
        \node (align) [align = right, below = \mytmpb - 2 ex, inner sep = 0, outer sep = 0, anchor = north west] at (current page.north west)%
        {%
            \hspace{\mytmpa}\hspace{0.5 em}\partname\ \thepart\\[1 ex]
            \color{stroke1}#3%
        };%
    \end{tikzpicture}%
}
\RedeclareSectionCommand%
[%
    font = \normalfont\Huge\sffamily,
    prefixfont = \normalfont\Huge\sffamily,
]
{part}

\usepackage{etoolbox}
\usepackage[strict]{changepage}
\usetikzlibrary{calc}

\makeatletter
\newcommand*{\mycheckoddpage}{\checkoddpage\if@twoside\else\oddpagetrue\fi}
\makeatother

\newbool{chapterHasANumber}
\newbool{chapterHasAStar}
\newcommand*{\thischapterlabel}{}
\renewcommand*{\chapterlinesformat}[3]%
{%
    \Ifnumbered{#1}{\setbool{chapterHasANumber}{true}}{\setbool{chapterHasANumber}{false}}%
    \Ifstr{#2}{}{\setbool{chapterHasAStar}{true}}{\setbool{chapterHasAStar}{false}}%
    \ifboolexpr{bool{chapterHasANumber} and not bool{chapterHasAStar}}%
    {%
        \renewcommand*{\thischapterlabel}{\color{chapternumber}\thechapter}
    }%
    {%
        \renewcommand*{\thischapterlabel}{\vphantom{0}}
    }%
    \mycheckoddpage
    \begin{tikzpicture}[overlay, remember picture]%
        \ifoddpage%
            \coordinate (anchor) at ($(current page.north west) + (\myinnermargin, -\mytopmargin)$);
            \node [inner sep = 0, outer sep = 0, anchor = north west] (numbernode) at (anchor)%
            {%
                \sffamily\fontsize{60}{60}\selectfont\thischapterlabel
            };%
            \draw[color = stroke1, line width = 0.7 mm] ($(numbernode.south west)-(0, 1 ex)$) -- ++(\paperwidth, 0);%
            \ifboolexpr{bool{chapterHasANumber} and not bool{chapterHasAStar}}%
            {%
                \node (align) [text width = \textwidth - 2 cm, align = right, right = \mybodywidth, inner sep = 0, outer sep = 0, anchor = east] at (numbernode.west) {#3};
            }%
            {%
                \node (align) [align = left, inner sep = 0, outer sep = 0, anchor = west] at (numbernode.west) {#3};
            }%
        \else%
            \coordinate (anchor) at ($(current page.north east) - (\myinnermargin, \mytopmargin)$);
            \node [inner sep = 0, outer sep = 0, anchor = north east] (numbernode) at (anchor)%
            {%
                \sffamily\fontsize{60}{60}\selectfont\thischapterlabel
            };%
            \draw[color = stroke1, line width = 0.7 mm] ($(numbernode.south east)-(0, 1 ex)$) -- ++(-\paperwidth, 0);%
            \ifboolexpr{bool{chapterHasANumber} and not bool{chapterHasAStar}}%
            {%
                \node (align) [text width = \textwidth - 2 cm, align = left, left = \mybodywidth, inner sep = 0, outer sep = 0, anchor = west] at (numbernode.east) {#3};
            }%
            {%
                \node (align) [align = right, inner sep = 0, outer sep = 0, anchor = east] at (numbernode.east) {#3};
            }%
        \fi%
    \end{tikzpicture}%
}
\RedeclareSectionCommand%
[%
    font = \color{stroke1}\normalfont\huge\sffamily,
    afterskip = 20 pt,
]
{chapter}

\BeforeStartingTOC[toc]{\pagestyle{plain}}
\AfterStartingTOC{\thispagestyle{plain}}

%% file: core/bibliography/bibliography_format.tex
\usepackage
[
    sortcites,              
    style = alphabetic,     
    defernumbers,           
    safeinputenc,           
    backref = true,         
    backrefstyle = three,   
    hyperref = true,        
    maxbibnames = 99,       
    maxcitenames = 2,       
]
{biblatex} 

\renewbibmacro{in:}%
{%
    \ifentrytype{article}{}{\printtext{\bibstring{in}\intitlepunct}}%
}

\renewbibmacro*{volume+number+eid}%
{%
    \printfield{volume}%
    \iffieldundef{number}{}{\addcolon}%
    \printfield{number}%
    \setunit*{\addcomma\space}%
    \printfield{eid}%
}

\DefineBibliographyStrings{english}%
{%
    backrefpage  = {\lowercase{s}ee page}, 
    backrefpages = {\lowercase{s}ee pages} 
}

\DeclareFieldFormat[article]{title}{\textbf{\color{stroke1}#1}}
\DeclareFieldFormat[inproceedings]{title}{\textbf{\color{stroke1}#1}}
\DeclareFieldFormat[thesis]{title}{\textbf{\color{stroke1}#1}}
\DeclareFieldFormat[book]{title}{\textbf{\color{stroke1}#1}}
\DeclareFieldFormat[unpublished]{title}{\textbf{\color{stroke1}#1}}
\DeclareFieldFormat[report]{title}{\textbf{\color{stroke1}#1}}
\DeclareFieldFormat[inbook]{chapter}{\textbf{\color{stroke1}#1}}
\DeclareFieldFormat[inbook]{title}{#1}
\DeclareFieldFormat{pages}{#1}

\newtoggle{authorend}
\togglefalse{authorend}

\DeclareBibliographyDriver{article}%
{%
  \usebibmacro{bibindex}%
  \usebibmacro{begentry}%
  \iftoggle{authorend}{}{\usebibmacro{author/translator+others}}%
  \setunit{\labelnamepunct}\newblock
  \usebibmacro{title}%
  \newunit
  \printlist{language}%
  \newunit\newblock
  \usebibmacro{byauthor}%
  \newunit\newblock
  \usebibmacro{bytranslator+others}%
  \newunit\newblock
  \printfield{version}%
  \newunit\newblock
  \usebibmacro{in:}%
  \usebibmacro{journal+issuetitle}%
  \newunit
  \usebibmacro{byeditor+others}%
  \newunit
  \usebibmacro{note+pages}%
  \newunit\newblock
  \iftoggle{bbx:isbn}
  {\printfield{issn}}
  {}%
  \newunit\newblock
  \usebibmacro{doi+eprint+url}%
  \newunit\newblock
  \usebibmacro{addendum+pubstate}%
  \setunit{\bibpagerefpunct}\newblock
  \usebibmacro{pageref}%
  \newunit\newblock
  \iftoggle{bbx:related}
  {\usebibmacro{related:init}%
    \usebibmacro{related}}
  {}%
  \usebibmacro{finentry}%
  \iftoggle{authorend}{\usebibmacro{author/translator+others}}{}%
}

\DeclareBibliographyDriver{inbook}%
{%
  \usebibmacro{bibindex}%
  \usebibmacro{begentry}%
  \iftoggle{authorend}{}{\usebibmacro{author/translator+others}}%
  \setunit{\labelnamepunct}\newblock
  \usebibmacro{chapter+pages}%
  \newunit
  \printlist{language}%
  \newunit\newblock
  \usebibmacro{byauthor}%
  \newunit\newblock
  \usebibmacro{in:}%
  \usebibmacro{bybookauthor}%
  \newunit\newblock
  \usebibmacro{maintitle+booktitle}%
  \newunit\newblock
  \usebibmacro{byeditor+others}%
  \newunit\newblock
  \printfield{edition}%
  \newunit
  \iffieldundef{maintitle}
  {\printfield{volume}%
    \printfield{part}}
  {}%
  \newunit
  \printfield{volumes}%
  \newunit\newblock
  \usebibmacro{series+number}%
  \newunit\newblock
  \printfield{note}%
  \newunit\newblock
  \usebibmacro{publisher+location+date}%
  \newunit\newblock
  \newunit\newblock
  \iftoggle{bbx:isbn}
  {\printfield{isbn}}
  {}%
  \newunit\newblock
  \usebibmacro{doi+eprint+url}%
  \newunit\newblock
  \usebibmacro{addendum+pubstate}%
  \setunit{\bibpagerefpunct}\newblock
  \usebibmacro{pageref}%
  \newunit\newblock
  \iftoggle{bbx:related}
  {\usebibmacro{related:init}%
    \usebibmacro{related}}
  {}%
  \usebibmacro{finentry}%
  \iftoggle{authorend}{\usebibmacro{author/translator+others}}{}%
}

\DeclareBibliographyDriver{inproceedings}%
{%
  \usebibmacro{bibindex}%
  \usebibmacro{begentry}%
  \iftoggle{authorend}{}{\usebibmacro{author/translator+others}}%
  \setunit{\labelnamepunct}\newblock
  \usebibmacro{title}%
  \newunit
  \printlist{language}%
  \newunit\newblock
  \usebibmacro{byauthor}%
  \newunit\newblock
  \usebibmacro{in:}%
  \usebibmacro{maintitle+booktitle}%
  \newunit\newblock
  \usebibmacro{event+venue+date}%
  \newunit\newblock
  \usebibmacro{byeditor+others}%
  \newunit\newblock
  \iffieldundef{maintitle}
  {\printfield{volume}%
    \printfield{part}}
  {}%
  \newunit
  \printfield{volumes}%
  \newunit\newblock
  \usebibmacro{series+number}%
  \newunit\newblock
  \printfield{note}%
  \newunit\newblock
  \printlist{organization}%
  \newunit
  \usebibmacro{publisher+location+date}%
  \newunit\newblock
  \usebibmacro{chapter+pages}%
  \newunit\newblock
  \iftoggle{bbx:isbn}
  {\printfield{isbn}}
  {}%
  \newunit\newblock
  \usebibmacro{doi+eprint+url}%
  \newunit\newblock
  \usebibmacro{addendum+pubstate}%
  \setunit{\bibpagerefpunct}\newblock
  \usebibmacro{pageref}%
  \newunit\newblock
  \iftoggle{bbx:related}
  {\usebibmacro{related:init}%
    \usebibmacro{related}}
  {}%
  \usebibmacro{finentry}%
  \iftoggle{authorend}{\usebibmacro{author/translator+others}}{}%
}

\DeclareBibliographyDriver{thesis}%
{%
  \usebibmacro{bibindex}%
  \usebibmacro{begentry}%
  \iftoggle{authorend}{}{\usebibmacro{author}}%
  \setunit{\labelnamepunct}\newblock
  \usebibmacro{title}%
  \newunit
  \printlist{language}%
  \newunit\newblock
  \usebibmacro{byauthor}%
  \newunit\newblock
  \printfield{note}%
  \newunit\newblock
  \printfield{type}%
  \newunit
  \usebibmacro{institution+location+date}%
  \newunit\newblock
  \usebibmacro{chapter+pages}%
  \newunit
  \printfield{pagetotal}%
  \newunit\newblock
  \iftoggle{bbx:isbn}
  {\printfield{isbn}}
  {}%
  \newunit\newblock
  \usebibmacro{doi+eprint+url}%
  \newunit\newblock
  \usebibmacro{addendum+pubstate}%
  \setunit{\bibpagerefpunct}\newblock
  \usebibmacro{pageref}%
  \newunit\newblock
  \iftoggle{bbx:related}
  {\usebibmacro{related:init}%
    \usebibmacro{related}}
  {}%
  \usebibmacro{finentry}%
  \iftoggle{authorend}{\usebibmacro{author}}{}%
}

\providebool{bbx:subentry}
\newbibmacro*{citenum}%
{
  \printtext[bibhyperref]{
    \printfield{prefixnumber}%
    \printfield{labelnumber}%
    \ifbool{bbx:subentry}
    {\printfield{entrysetcount}}
    {}}%
}

\DeclareCiteCommand{\conline}[\mkbibbrackets]
{\usebibmacro{prenote}}
{\usebibmacro{citeindex}%
  \usebibmacro{citenum}}
{\multicitedelim}
{\usebibmacro{postnote}}

%% file: packages_and_commands/standard_packages.tex
\usepackage
[
    babel = true, 
]
{microtype}           
\usepackage{csquotes} 

\usepackage{amsmath}
\usepackage{amssymb}
\usepackage{amsthm}
\usepackage{thmtools}
\usepackage{mathtools}
\usepackage{thm-restate}
\usepackage{dsfont}        
\usepackage{braceMnSymbol} 

\usepackage
[
    ttscale = 0.85, 
]
{libertine} 
\usepackage
[
    libertine,    
    slantedGreek, 
    vvarbb,       
    libaltvw,     
]
{newtxmath} 
\usepackage{url} 
\usepackage{bm}  

\usepackage{graphicx} 
\usepackage
[
    subrefformat = simple, 
    labelformat = simple,  
]
{subcaption}         
\usepackage{wrapfig} 

\columnsep = \mymargininnersep

\usepackage{array}     
\usepackage{booktabs}  
\usepackage{longtable} 
\usepackage{pdflscape} 

\usepackage{enumitem} 

\usepackage
[
    ruled,         
    vlined,        
    linesnumbered, 
]
{algorithm2e} 

\usepackage{xspace}   
\usepackage
[
    shortcuts, 
]
{extdash}             
\usepackage{setspace} 

\usepackage{xparse}    
\usepackage{footnote}  
\usepackage{afterpage} 
\usepackage
[
    textsize = scriptsize, 
]
{todonotes}            


\usepackage
[
    bookmarks = true,                 
    bookmarksopen = false,            
    bookmarksnumbered = true,         
    pdfstartpage = 1,                 
    pdftitle = {{\printTitle}},       
    pdfauthor = {{\printAuthor}},     
    pdfsubject = {{\printSubject}},   
    pdfkeywords = {{\printKeywords}}, 
    breaklinks = true,                
    \ifprintVersion
        hidelinks,                    
    \else
    colorlinks = true,            
    allcolors = stroke1,          
    \fi
]
{hyperref} 

\usepackage
[
    noabbrev,   
    nameinlink, 
]
{cleveref} 

%% file: packages_and_commands/general_commands.tex
\newcommand*{\colloquialDegreeName}{Master}

\newcommand*{\degreeAbbreviation}{M.}

\ifbachelorThesis
    \renewcommand*{\colloquialDegreeName}{Bachelor}
    
    \renewcommand*{\degreeAbbreviation}{B.}
\fi

\makeatletter
\def\IfEmptyTF#1%
{%
    \if\relax\detokenize{#1}\relax%
        \expandafter\@firstoftwo%
    \else%
        \expandafter\@secondoftwo%
    \fi%
}
\makeatother

\NewDocumentCommand{\mathOrText}{m}
{%
    \ensuremath{#1}\xspace%
}

\let\originalleft\left
\let\originalright\right
\renewcommand{\left}{\mathopen{}\mathclose\bgroup\originalleft}
\renewcommand{\right}{\aftergroup\egroup\originalright}

\makeatletter
\DeclareRobustCommand{\bfseries}%
{%
    \not@math@alphabet\bfseries\mathbf%
    \fontseries\bfdefault\selectfont%
    \boldmath%
}
\makeatother

\xspaceaddexceptions{]\}}

\allowdisplaybreaks

\crefname{ineq}{inequality}{inequalities}
\creflabelformat{ineq}{#2{\upshape(#1)}#3}

\crefname{term}{term}{terms}
\creflabelformat{term}{#2{\upshape(#1)}#3}

\let\oldfootnote\footnote

\newlength{\spaceBeforeFootnote} 
\newlength{\spaceAfterFootnote}  

\RenewDocumentCommand{\footnote}{o o o m}%
{%
    \IfNoValueTF{#1}%
    {%
        \oldfootnote{#4}%
    }%
    {%
        \setlength{\spaceBeforeFootnote}{\IfEmptyTF{#1}{0}{#1} em}%
        \IfNoValueTF{#2}%
        {%
            \hspace*{\spaceBeforeFootnote}\oldfootnote{#4}%
        }%
        {%
            \setlength{\spaceAfterFootnote}{\IfEmptyTF{#2}{0}{#2} em}%
            \hspace*{\spaceBeforeFootnote}\IfNoValueTF{#3}{\oldfootnote{#4}}{\oldfootnote[#3]{#4}}\hspace*{\spaceAfterFootnote}%
        }%
    }%
}

\makesavenoteenv{figure}
\makesavenoteenv{table}
\makesavenoteenv{tabular}

\iffancyTheorems
    \declaretheoremstyle
    [
        spaceabove = \topsep,
        spacebelow = \topsep,
        headfont = \bfseries,
        headformat = \textcolor{stroke1}{$\blacktriangleright$} \NAME~\NUMBER \NOTE,
        notefont = \bfseries,
        notebraces = {(}{)},
        bodyfont = \normalfont,
        postheadspace = 0.5 em,
        qed = \textcolor{stroke1}{\bfseries$\blacktriangleleft$},
    ]
    {myTheoremStyle}

    \declaretheorem
    [
        style = myTheoremStyle,
        name = Conjecture,
        numberwithin = chapter,
    ]
    {conjecture}
    
    \declaretheorem
    [
        style = myTheoremStyle,
        name = Proposition,
        sharenumber = conjecture,
    ]
    {proposition}
    
    \declaretheorem
    [
        style = myTheoremStyle,
        name = Claim,
        sharenumber = conjecture,
    ]
    {claim}
    
    \declaretheorem
    [
        style = myTheoremStyle,
        name = Lemma,
        sharenumber = conjecture,
    ]
    {lemma}
    
    \declaretheorem
    [
        style = myTheoremStyle,
        name = Corollary,
        sharenumber = conjecture,
    ]
    {corollary}
    
    \declaretheorem
    [
        style = myTheoremStyle,
        name = Theorem,
        sharenumber = conjecture,
    ]
    {theorem}
    
    \declaretheorem
    [
        style = myTheoremStyle,
        name = Definition,
        sharenumber = conjecture,
    ]
    {definition}
    
    \declaretheorem
    [
        style = myTheoremStyle,
        name = Example,
        sharenumber = conjecture,
    ]
    {example}
    
    \declaretheorem
    [
        style = myTheoremStyle,
        name = Remark,
        sharenumber = conjecture,
    ]
    {remark}
    
\else
    \theoremstyle{plain}

\fi

\NewDocumentCommand{\functionTemplate}{m m m m o}%
{%
    \IfNoValueTF{#5}%
    {%
        \mathOrText{#1\left#2{#4}\right#3}%
    }%
    {%
        \mathOrText{#1#5#2{#4}#5#3}%
    }%
}

\newcommand*{\leftBracketType}{(}
\newcommand*{\rightBracketType}{)}

\NewDocumentCommand{\createFunction}{m m o o}%
{%
    \renewcommand*{\leftBracketType}{\IfNoValueTF{#3}{(}{#3}}%
    \renewcommand*{\rightBracketType}{\IfNoValueTF{#4}{)}{#4}}%
    \NewDocumentCommand{#1}{o o}%
    {%
        \IfNoValueTF{##1}%
        {%
            \mathOrText{#2}%
        }%
        {%
            \functionTemplate{#2}{\leftBracketType}{\rightBracketType}{##1}[##2]%
        }%
    }%
}

\DeclareDocumentCommand{\probabilisticFunctionTemplate}{m m O{} o}
{%
    \functionTemplate{#1}%
    {\lbrack}%
    {\rbrack}%
    {#2\IfEmptyTF{#3}{}{\ \IfNoValueTF{#4}{\left}{#4}\vert\ \vphantom{#2}#3\IfNoValueTF{#4}{\right.}{}}}%
    [#4]%
}

\ifboldNumberSets

    \newcommand*{\indicatorFunctionSymbol}{\mathbf{1}}
\else

    \newcommand*{\indicatorFunctionSymbol}{\mathds{1}}
\fi

\RenewDocumentCommand{\Pr}{m O{} o}%
{%
    \probabilisticFunctionTemplate{\mathrm{Pr}}{#1}[#2][#3]%
}

\NewDocumentCommand{\E}{m O{} o}%
{%
    \probabilisticFunctionTemplate{\mathrm{E}}{#1}[#2][#3]%
}

\NewDocumentCommand{\Var}{m O{} o}%
{%
    \probabilisticFunctionTemplate{\mathrm{Var}}{#1}[#2][#3]%
}

\DeclareDocumentCommand{\bigO}{m o}%
{%
    \functionTemplate{\mathrm{O}}{(}{)}{#1}[#2]%
}

\DeclareDocumentCommand{\smallO}{m o}%
{%
    \functionTemplate{\mathrm{o}}{(}{)}{#1}[#2]%
}

\DeclareDocumentCommand{\bigTheta}{m o}%
{%
    \functionTemplate{\upTheta}{(}{)}{#1}[#2]%
}

\DeclareDocumentCommand{\bigOmega}{m o}%
{%
    \functionTemplate{\upOmega}{(}{)}{#1}[#2]%
}

\DeclareDocumentCommand{\smallOmega}{m o}%
{%
    \functionTemplate{\upomega}{(}{)}{#1}[#2]%
}

\DeclareDocumentCommand{\eulerE}{o}%
{%
    \mathOrText{\mathrm{e}\IfNoValueTF{#1}{}{^{#1}}}%
}

\DeclareDocumentCommand{\poly}{m o}%
{%
    \functionTemplate{\mathrm{poly}}{(}{)}{#1}[#2]%
}

\createFunction{\id}{\mathrm{id}}

\NewDocumentCommand{\ind}{m o o}%
{%
    \IfNoValueTF{#2}%
    {%
        \mathOrText{\indicatorFunctionSymbol_{#1}}%
    }%
    {%
        \functionTemplate{\indicatorFunctionSymbol_{#1}}{(}{)}{#2}[#3]%
    }%
}

\DeclareDocumentCommand{\dom}{m o}%
{%
    \functionTemplate{\mathrm{dom}}{(}{)}{#1}[#2]%
}

\DeclareDocumentCommand{\rng}{m o}%
{%
    \functionTemplate{\mathrm{rng}}{(}{)}{#1}[#2]%
}

\DeclareDocumentCommand{\d}{o}%
{%
    \mathrm{d}\IfNoValueTF{#1}{}{^{#1}}%
}

\DeclareDocumentCommand{\set}{m m o}%
{
    \mathOrText{\IfNoValueTF{#3}{\left}{#3}\{#1\ \IfNoValueTF{#3}{\left}{#3}\vert\
        \vphantom{#1}#2\IfNoValueTF{#3}{\right.}{}\IfNoValueTF{#3}{\right}{#3}\}}
}

%% file: core/title_page/title_page.tex
\newgeometry
{
    textwidth = 134 mm,
    textheight = 220 mm,
    top = 38 mm,
    inner = 38 mm,
}

\begin{titlepage}
    \sffamily
    \begin{center}
        \vfil
        {\LARGE
            \rule[1 ex]{\textwidth}{1.5 pt}
            \onehalfspacing\printTitleBold\\
            \rule[-1 ex]{\textwidth}{1.5 pt}
        }
        \vfil
        {\Large\textbf{\printAuthor}}
        \vfil
        {\large \colloquialDegreeName's thesis\\[0.25 ex]
        submitted for the degree of}\\[0.25 ex]
        \bigskip
        {\Large \colloquialDegreeName{} of Science}\\[0.5 ex]
        {\large\emph{(\degreeAbbreviation\,Sc.)}}\\
        \bigskip
        {\large in\\[0.25 ex]
        \printProgram}
        \vfil
        {\large submitted on \printDateReceived{} to the\\[0.25 ex]
        Chair of Artificial Intelligence and Intelligent Systems\\[0.25 ex]
        Digital Engineering Faculty\\[0.25 ex]
        University of Potsdam}
    \end{center}

    \vfil
    \begin{table}[h]
        \centering
        \large
        \sffamily
        {\def\arraystretch{1.2}
            \begin{tabular}{>{\bfseries}p{3.8 cm}p{5.3 cm}}
                Reviewer    & \printNameOfSupervisor\\
                Supervisors & \printAdditionalExaminers
            \end{tabular}
        }
    \end{table}
\end{titlepage}

\restoregeometry